\documentclass[twoside]{article}

\usepackage[accepted]{aistats2022}
\usepackage[utf8]{inputenc}
\usepackage[english]{babel}
\usepackage{csquotes}
\usepackage{amsmath}
\usepackage{amsthm}
\usepackage{amsfonts}
\usepackage{algorithm} 
\usepackage{algpseudocode}
\usepackage{multirow}
\usepackage{multicol}
\usepackage{xcolor}
\usepackage{color}
\usepackage{graphicx}
\usepackage{hyperref}
\usepackage{array}
\usepackage{textcomp}
\usepackage[round]{natbib}

\newtheorem{propo}{Proposition}

\newcommand{\Li}{\mathbf{L}}
\newcommand{\R}{\mathbb{R}}
\newcommand{\cc}{\mathbf{c}}
\newcommand{\x}{\mathbf{x}}
\newcommand{\y}{\mathbf{y}}

\newcommand{\A}{\mathbf{A}}
\newcommand{\App}{\mathbf{\tilde{A}}}

\newcommand{\C}{\mathbf{C}}
\newcommand{\I}{\mathbf{I}}
\newcommand{\Ro}{\mathbf{R}}
\newcommand{\Kpp}{\mathbf{\tilde{K}}}
\newcommand{\Ksu}{\mathbf{\bar{K}}}
\newcommand{\K}{\mathbf{K}}
\newcommand{\U}{\mathbf{U}}

\newcommand{\W}{\mathbf{W}}
\newcommand{\Q}{\mathbf{Q}}

\newcommand{\Krein}{Kre\u{\i}n}
\newcommand{\Nys}{Nystr\"om}

\newcommand{\comm}[1]{}

\newcommand{\argmin}[1]{\underset{#1}{\operatorname{arg}\,\operatorname{min}}\;}

\renewcommand{\footskip}{35pt}

\newcolumntype{?}{!{\vrule width 1pt}}
\newcolumntype{H}{>{\setbox0=\hbox\bgroup}c<{\egroup}@{}}
\begin{document}
\twocolumn[
\runningtitle{Revisiting Memory Efficient Kernel Approximation}
\aistatstitle{Revisiting Memory Efficient Kernel Approximation: \\
An Indefinite Learning Perspective}
\aistatsauthor{ Simon Heilig \And Maximilian M\"unch \And  Frank-Michael Schleif }

\aistatsaddress{University of Bamberg \And UAS W\"urzburg-Schweinfurt,\\University of Groningen \And UAS W\"urzburg-Schweinfurt} ]

\begin{abstract}
  Matrix approximations are a key element in large-scale algebraic machine learning approaches. The recently proposed method MEKA \citep{MEKA} effectively employs two
common assumptions in Hilbert spaces:  the low-rank property of an inner product matrix obtained from a shift-invariant kernel function and a data compactness hypothesis by means of an inherent block-cluster structure.
In this work, we extend MEKA to be applicable not only for shift-invariant kernels but also for non-stationary kernels like polynomial kernels and an extreme learning kernel. We also address in detail how to handle non-positive semi-definite kernel functions within MEKA, either caused by the approximation itself or by the intentional use of general kernel functions.
We present a Lanczos-based estimation of a spectrum shift to develop a stable positive semi-definite MEKA approximation, also usable in classical convex optimization frameworks. Furthermore, we support our findings with theoretical considerations and a variety of experiments on synthetic and real-world data. 
\end{abstract}

\section{INTRODUCTION}
Typical machine learning algorithms contain matrix operations that do not scale to scenarios with multiple thousand samples. 
Especially in the field of kernel machines, algorithms like the well-known support vector machine (SVM), developed by \citet{cortes1995support}, contain a square kernel matrix $\K \in \R^{n\times n}$ as the critical part, where $n$ is the number of data points. Similar issues also exist in Gaussian Process Models \citep{DBLP:books/lib/RasmussenW06}.
Those matrices cause both runtime and space complexity issues for practical applications. For this purpose, matrix approximation has been widely studied to speed up these algorithms \citep{halkoApproxRandom,MEKA,oglic2019scalable,DBLP:journals/ijon/GisbrechtS15}.

An alternative to matrix approximation is to approximate a particular kernel function using a mapping approach as in \citet{le2013fastfood} or by an 
explicit feature expansion with, e.g., random fourier features as proposed by \citet{rksRahimi}. This, however, performed inferior in comparison to the original MEKA.
Additionally, one would like to be able to use a wide variety of kernel functions,
preferable arbitrary similarity measures, to obtain an optimal and memory storage efficient prediction model.\\
As recently shown, symmetric matrices of arbitrary similarity measures can be effectively approximated by the \Nys{}-method 
\citep{oglic2019scalable,DBLP:journals/ijon/GisbrechtS15}.\\
In order to approximate a matrix using the \Nys{} approximation, a low-rank structure is required to guarantee small error. The matrix obtained by the classical \Nys{} method can be stored with $O(n\hat{k})$ space, where $n$ is the number of data points and $\hat{k}$ is the total rank of the approximation \citep{MEKA}. The objective of the original MEKA approach proposed by \citet{MEKA} is to obtain a much better space complexity compared to existing methods. 
After observing that clustering in the input space reveals a block-wise structure of the matrix, \citet{MEKA} developed a novel technique to exploit this structure without reconstructing the original kernel matrix at any step. By this, a memory storage of $O(nk + (ck)^2)$ for the approximated matrix is achieved, where $k$ is the rank per clustering-block and $c$ the number of clusters and the overall rank is $ck$ (note: $O(n\hat{k}) = O(nck)$). 

The kernel theory \citep{hofmann2008kernel} is based on positive semi-definite (psd) and symmetric matrices, but there is recent research that argues for also considering  indefinite kernels \citep{tl1Suykens,mehrkanoon2018indefinite,schleif2015indefinite,SchleifMEB}. Such kernels emerge from domain-specific object comparisons like protein sequence alignments or like in case of the truncated Manhattan kernel (TL1) \citet{tl1Suykens} are motivated by local learning properties. Since these kernels do not fulfill the psd assumption of the kernel theory, new techniques were proposed to handle such non-psd matrices \citep{loosli2015learning,liu2018indefinite,oglic2019scalable,DBLP:journals/ijon/GisbrechtS15} or to convert the matrices to be psd \citep{schleif2015indefinite}.\\
As briefly noticed in \citet{MEKA}, MEKA can lead to non-psd matrix approximations and hence this paper aims to discuss different aspects of this problem and to show how these limitations can be overcome. 
\begin{itemize}
    \item To what extend does indefiniteness occur in MEKA and what is the effect?
    \item How to implement a correction strategy while preserving memory efficiency?
    \item How can MEKA be extended to handle a more extensive range of kernel functions, especially indefinite ones?
\end{itemize}
To answer these questions, the paper is organized as follows: Section \ref{sec:preliminaries} introduces the relevant notation and background. Section \ref{sec:meka} describes MEKA with its core steps, studies the occurrence of invalid approximations, and extends MEKA to non-stationary and indefinite kernels. Subsequently in Section \ref{sec:revision}, the shift correction is realized by computing the shifting parameter using the Lanczos iteration in a highly efficient manner. Our proposal is evaluated with different classification tasks solved by an SVM using different kernel functions. 

\section{PRELIMINARIES}\label{sec:preliminaries}
This section introduces basic notation and concepts of matrix approximation and kernel methods.

\subsection{Matrix Approximation}
Given a matrix $\A \in \R^{n \times n}$, the objective of matrix approximation algorithms is to achieve a low approximation error at a given memory constraint,  specified by the rank of the resulting matrix $\App$:
\begin{equation}\label{eq:approximation}
    \min { \| \A -  \App\|_F}
\end{equation}
where $\App$ is the approximated matrix with rank $k$ and $\|\cdot\|_F$ the Frobenius norm \citep{kishore2017literature}. This concept can be seen as a data compression technique, since an intrinsic low-rank matrix contains many linear dependent rows or columns, reflecting redundant information. Note that the optimal solution for Equation \eqref{eq:approximation} is the truncated singular value decomposition (SVD), which is prohibitive in the large-scale setting. In order to maintain efficiency with huge amounts of data, probability-based decomposition methods such as randomised CUR decomposition have been developed \citep{drineas2008relative}. The idea is to select certain rows of $\A$ in matrix $\Ro$ and columns in matrix $\C$ to capture the relevant subspace, with $\U$ as a link matrix and eventually solve an optimization problem:
\begin{equation}\label{eq:CUR}
		\min \; ||\A - \C\U\Ro||^2_F.
\end{equation}
In particular for kernel matrices $\K$ (square and symmetric), according to \citet{Nystroem} and \citet{sun2015reviewNys}, it is possible to approximate the eigenspectrum with the \Nys{} expansion, where only $l$ landmarks (selected columns or rows) and a pseudo-inverse of a rank-$k$ approximation of the $l \times l$ intersection matrix $\W$ are required to build a rank-$k$ approximation of the original matrix: 
\begin{equation}\label{eq:Nys}
    \Kpp = \K_{(\cdot,\mathcal{I})}\W_k^\dagger\K_{(\cdot,\mathcal{I})}^T,
\end{equation}
where $\mathcal{I}$ is a set of landmark indicators consisting of $l$ indices and $\W_k^\dagger$ as the pseudo-inverse of the rank-$k$ approximation obtained by the truncated SVD. 
As it is later on required in the MEKA formulation to decompose a matrix not in three but in two matrices, one can split the $\W_k^\dagger$ matrix based on the eigenvalue decomposition (EVD): $\W_k = \U_{W,k}\mathbf{\Lambda}_{W,k}\U_{W,k}^T$.
Assuming only non-negative eigenvalues,
Equation \eqref{eq:Nys} is written as follows:
\begin{align}\label{eq:Nys3}
\begin{split}
	\Kpp &= \underbrace{\K_{(\cdot,\mathcal{I})}\U_{W,k}\left(\mathbf{\Lambda}_{W,k}^\dagger\right)^{\frac{1}{2}}}_{\mathbf{H}}  \underbrace{\left(\mathbf{\Lambda}_{W,k}^\dagger\right)^{\frac{1}{2}}  \U_{W,k}^T\K_{(\cdot,\mathcal{I})}^T}_{\mathbf{H}^T},\\
	&= \mathbf{H}\mathbf{H}^T.
\end{split}
\end{align}
\subsection{Kernel Theory}\label{subsec:kernel}
Given a collection of $n$ data points $X = \{\x_1, \x_2, \cdots, \x_n \},\text{ where } \x_i \in \mathcal{X} $, the kernel matrix $\K_{i,j} = k(\x_i,\x_j) \in \R^{n\times n}$ is created with a kernel function $k: \mathcal{X} \times \mathcal{X} \to \R$, where $k(\x,\y) = \langle \phi(\x), \phi(\y)  \rangle_\mathcal{H}$ with $\phi: \mathcal{X} \to \mathcal{H}$ as a mapping into an arbitrarily high dimensional Hilbert space $\mathcal{H}$. Following the line of \cite{MEKA}, we assume $\mathcal{X} = \R^d$ as $d$-dimensional input space.

As a matter of fact, any function of the form of $k$ which gives rise to a psd kernel matrix reflects such an implicit inner product in a possibly infinite-dimensional Hilbert space \citep{hofmann2008kernel}. This technique allows a flexible way of defining a data representation leading to a powerful and more important linear classifier in such a Hilbert space. By employing the inner product of this space implicitly, without the need of the explicit mapping $\phi$ to be present, a non-linear classification in the input space $\mathcal{X}$ is achieved, also known as the kernel trick \citep{hofmann2008kernel}.

If a kernel function leads to a non-psd kernel matrix, a \Krein{} space is induced instead \citep{duin2005dissimilarity,schleif2015indefinite}. A \Krein{} space $\mathcal{K}$ is spanned by two orthogonal Hilbert spaces $\mathcal{H}_+$ and $\mathcal{H}_-$ so that $\langle \x, \mathbf{y} \rangle_{\mathcal{K}} = \langle \x_+, \mathbf{y}_+ \rangle_{\mathcal{H}_+} - \langle \x_-, \mathbf{y}_- \rangle_{\mathcal{H}_-}$ for $\x, \mathbf{y} \in \mathcal{K}$ and $\x_+, \mathbf{y}_+$ as the projection onto $\mathcal{H}_+$ and $\x_-, \mathbf{y}_-$ as the projection onto $\mathcal{H}_-$.\\Consequently, the norm induced by the indefinite inner product lacks of metric properties, which is a problem for branch-and-bound methods \citep{MORRISON201679} and the objective function of a SVM becomes non-convex \citep{schleif2015indefinite}. For kernel approaches, this problem can be tackled by various strategies, e.g., learning a proxy matrix which is constrained to be psd \citep{luss2009support}, by correcting the eigenspectrum of the matrix to remove the negative eigenvalues \citep{schleif2015indefinite,munch2020structure}
or by developing methods in the \Krein{} space, e.g., the indefinite SVM from \citet{loosli2015learning,SchleifMEB,DBLP:journals/pr/SchleifT17} or \citet{oglic2019scalable,DBLP:journals/ijon/GisbrechtS15} and by a shift correction based SVM from \citet{DBLP:conf/esann/Loosli19}. These approaches are in general costly and do not scale to large data as considered by MEKA, but only to a few thousand samples at most.

A kernel function is shift-invariant if and only if $k(\x,\y) = f(\eta(\x - \y))$ for a function $f: \mathcal{X} \to \R$ and a scaling parameter $\eta$.
As a consequence, kernel functions based on distances, such as the Gaussian rbf kernel:
\begin{equation*}
    k_{rbf}(\x,\y) = \exp\left(-\gamma\|\x-\y\|_2^2\right),
\end{equation*} 
are shift-invariant. However, for inner product based (non-stationary) functions, a restriction to the unit sphere has to be made \citep{pennington2015spherical}.
\begin{propo}[\citet{pennington2015spherical}]\label{prop:invariant}
\textit{If the input data is restricted to a unit sphere via the norm $\|\cdot\|_2$, then any function $f: \mathcal{X} \times \mathcal{X} \to \R$ that is based on a computation of an inner product of the input data is shift-invariant.}
\begin{proof}
Given $\x,\y \in \mathcal{X}$ in $d$ dimensions, so that $\x = (x_1,x_2,\dots,x_{d-1},x_d)$ and $\y = (y_1,y_2,\dots,y_{d-1},y_d)$.
\allowdisplaybreaks
\begin{align*}
   & \|\x-\y\|_2^2 = \langle \x-\y,\x-\y\rangle_2\\
                  &= \sum_{i=1}^d{(x_i - y_i)^2} = \sum_{i=1}^d{(x_i^2 - 2x_iy_i + y_i^2)}\\
                  &= \underbrace{\sum_{i=1}^d{x_i^2}}_{= 1} + \sum_{i=1}^d{(-2x_iy_i)} + \underbrace{\sum_{i=1}^d{y_i^2}}_{= 1} \;\;, \|\x\|_2 = \|\y\|_2 = 1\\
                  &= 2 -2\sum_{i=1}^d{x_iy_i} 
                  = 2 -2\langle\x,\y\rangle_2\qedhere
\end{align*}
\end{proof}
\end{propo}
\paragraph{Remark 1.}
In case there is no real vector space given and only a kernel matrix is present, a matrix normalization is required. Since the diagonal elements represent the kernel function evaluation of $k(\x_i,\x_i),\; i = 1\dots n$, restricting the input space on an unit sphere is equivalent to a constant diagonal. This can be achieved by:
$\langle\x_i,\x_j \rangle = \frac{\langle\x_i,\x_j \rangle}{\sqrt{\langle\x_i,\x_i \rangle}\sqrt{\langle\x_j,\x_j \rangle}}$, where $\langle\x_i,\x_i \rangle > 0, \langle\x_j,\x_j \rangle > 0,\; i,j = 1 \dots n$.
Moreover, by this transformation, the self-similarities of any kernel become a constant. 
\paragraph{Remark 2.}
As noted in \citet{pennington2015spherical}, there are kernel functions that become non-psd in consequence of this normalization since it has been proven that specific kernel functions are not a Fourier transform of a finite non-negative Borel measure anymore. As an example, the polynomial kernel:
\begin{equation*}\label{eq:polykernel}
    k_{poly}(\x,\y) = \left(1-\frac{\|\x-\y\|^2_2}{a^2}\right)^p = \alpha(q +\langle \x,\y\rangle)^p,
\end{equation*}
with $q = ( a^2 / 2 )-1$ and $\alpha = (a^2/2)^p$ becomes non-psd for $a> 2$ and $p >= 1$. Hence, non-psd kernel functions occur more frequently than expected and need to be addressed \citep{pennington2015spherical,liu2021fast}.
\paragraph{Remark 3.}
The projection of the input space onto a sphere can lead to overlapping of clusters, which were on a similar projection ray but had a distinctive separation beforehand. Further, we will assume that this effect is negligible, since it is a data specific use case and multiple works follow the line of \cite{pennington2015spherical} such as \cite{liu2021fast}, which extends also to dot-product based kernel functions. A learned embedding on such a manifold while preserving the differences is able to overcome this problem, as mentioned in \citep{wilson2010spherical}, but it needs to fit in the large-scale setting of this work.

\section{ANALYSIS OF MEKA}\label{sec:meka}
\begin{figure*}[ht]
	\centering
	\begin{minipage}{0.242\textwidth}
		\centering
		\includegraphics[width=\textwidth]{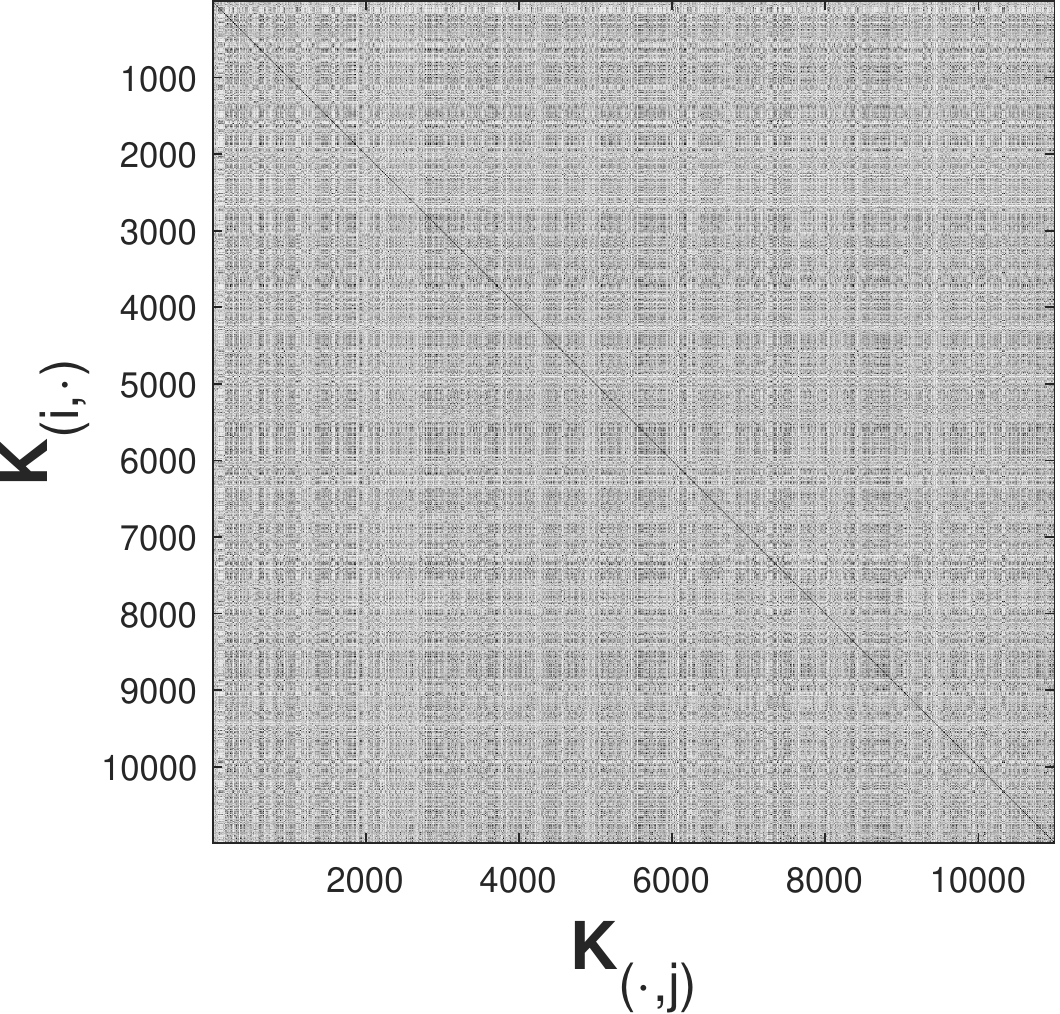}\\\hspace{0.7cm}(a)
	\end{minipage}
	\begin{minipage}{0.242\textwidth}
		\centering
		\includegraphics[width=\textwidth]{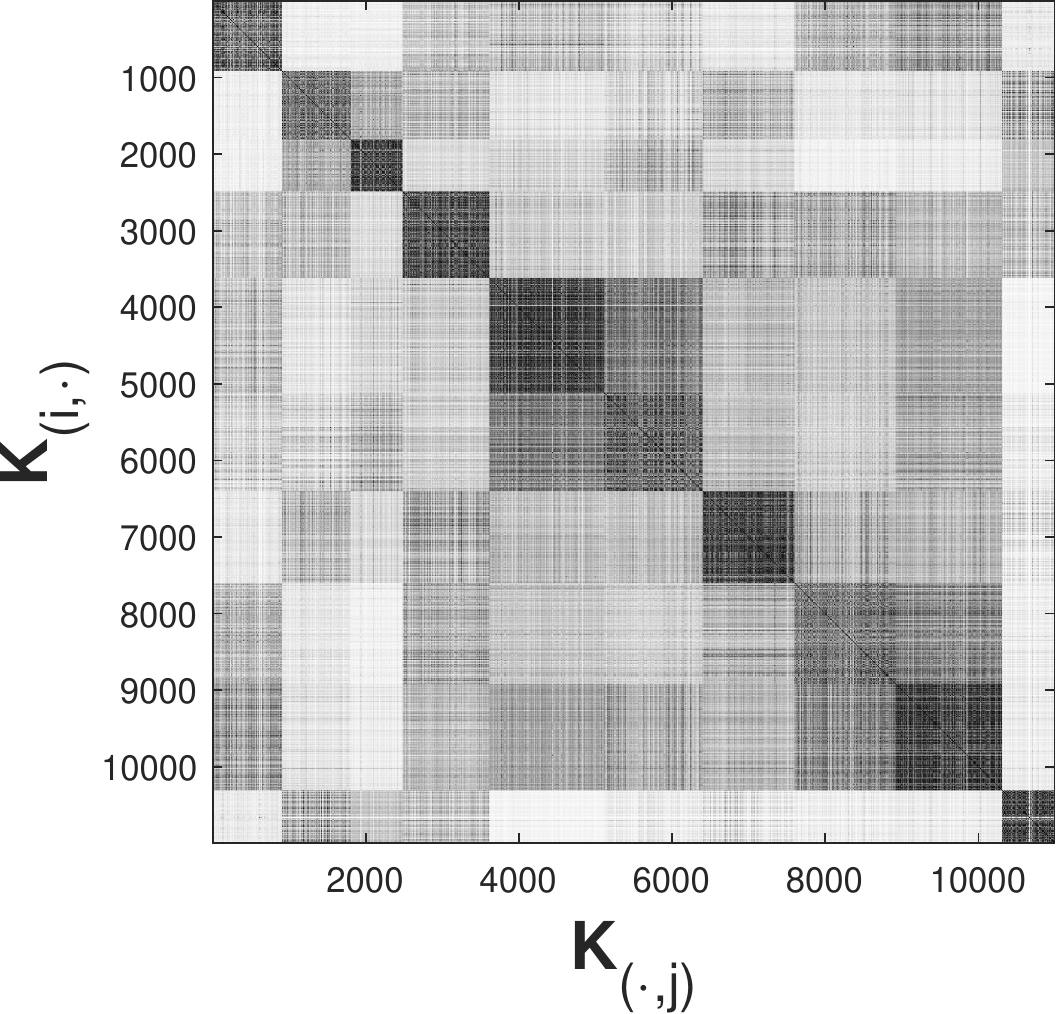}\\\hspace{0.7cm}(b)
	\end{minipage}
	\begin{minipage}{0.242\textwidth}
		\centering
		\includegraphics[width=\textwidth]{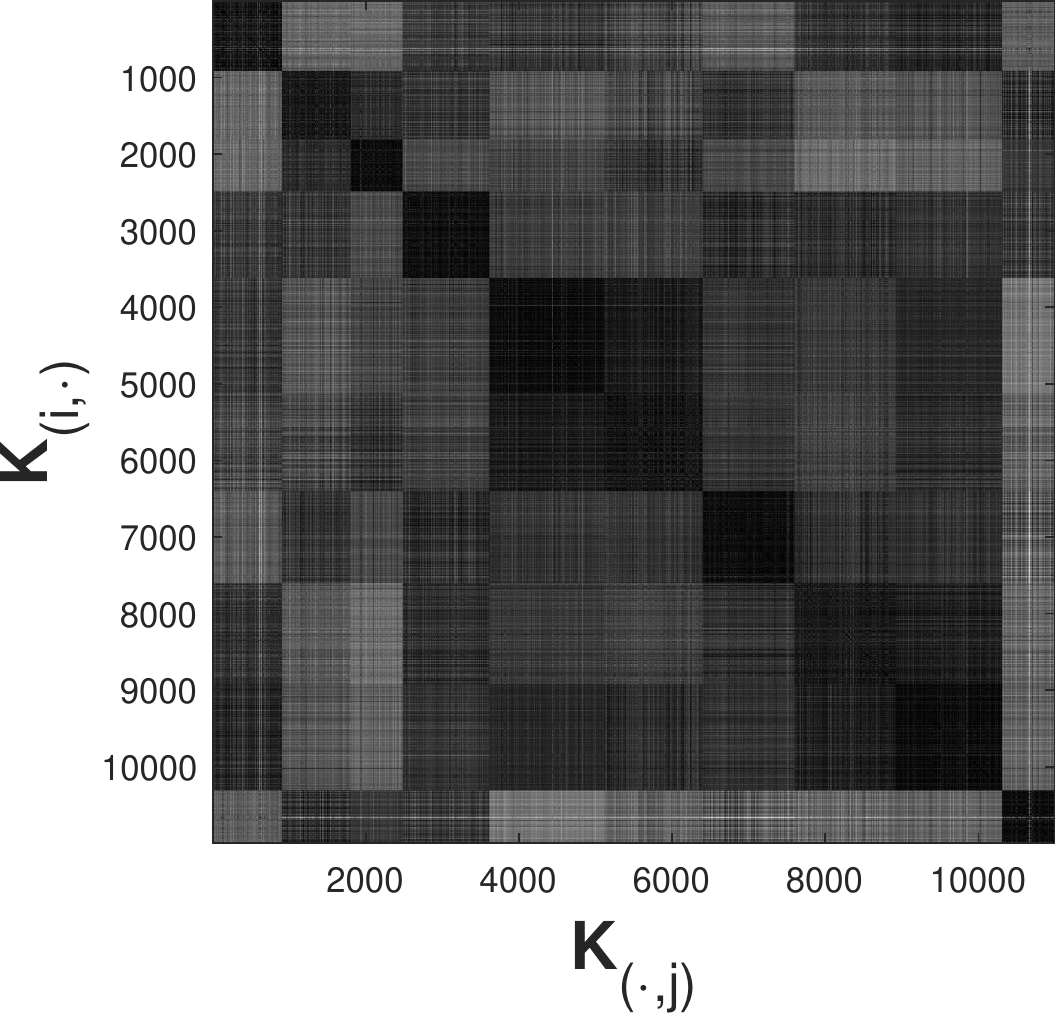}\\\hspace*{0.7cm}(c)
	\end{minipage}
	\begin{minipage}{0.242\textwidth}
		\centering
		\includegraphics[width=\textwidth]{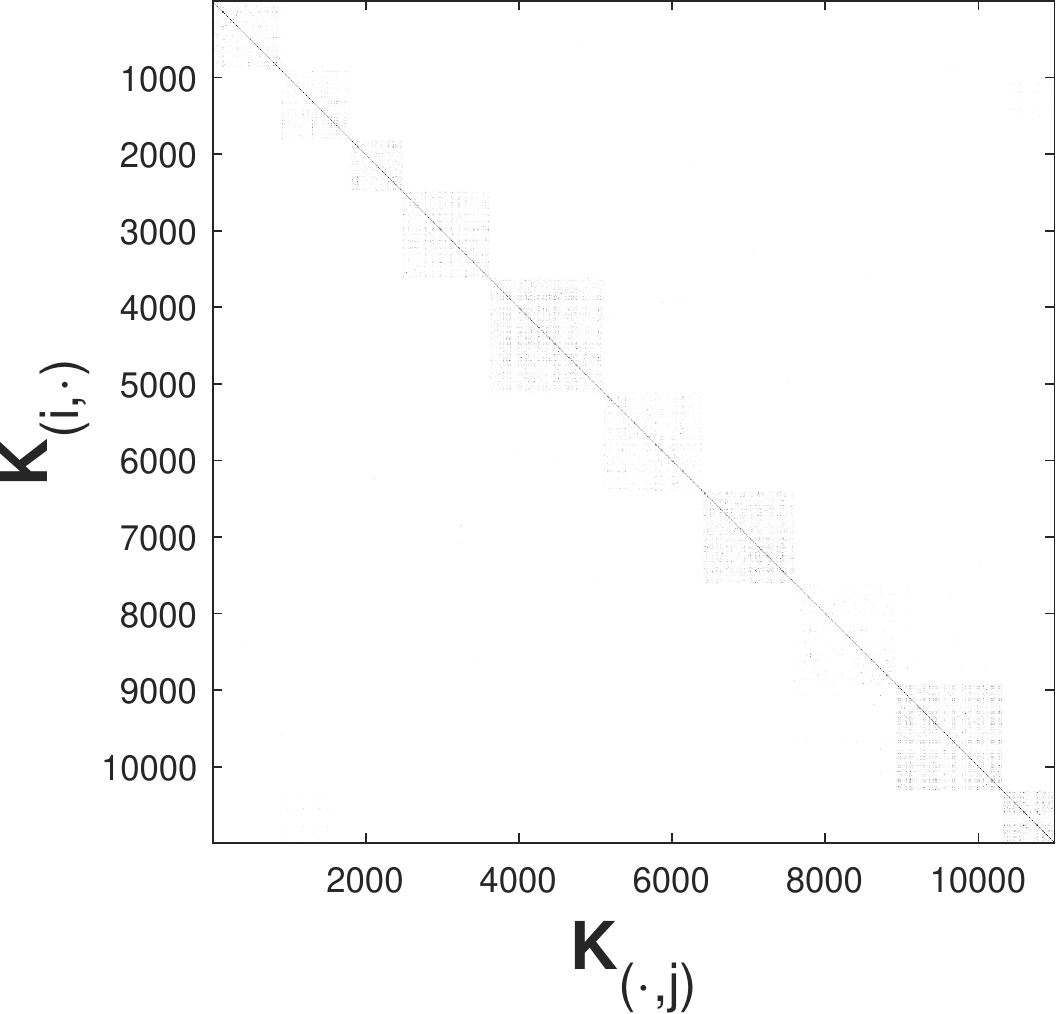}\\\hspace*{0.7cm}(d)
	\end{minipage}
	\caption{Illustration of the block-wise structure in the kernel matrix associated to a Gaussian rbf kernel, applied on the \textsf{pendigit} data set. The darker the higher the value of the kernel function. (a) the unordered kernel matrix for $\gamma = 0.05$. (b) rearranged matrix with respect to a c-means clustering. (c) $\gamma = 1\cdot 10^{-6}$ leads to the low-rank case. (d) $\gamma = 1.2$ leads to a block-diagonal kernel matrix and finally approaches an identity matrix.\label{pic:kmeansClustering}}
	
\end{figure*}
After introducing the basic foundations of kernels and matrix approximations thereof, a brief outline of the MEKA algorithm is presented below. Further, we enable MEKA to be applicable for non-stationary and indefinite kernels and the occurrence of both positive and negative eigenvalues is studied on real-world and synthetic data.
\subsection{Memory Efficient Kernel Approximation}
The key aspect of MEKA is the combination of a clustering in the input space exploiting the effect of the scaling parameter of a shift-invariant kernel function. 
As \citet{MEKA} observed, a kernel matrix arranged with respect to a clustering changes from low-rank to block-diagonal when the scaling parameter $\gamma$ increases.
Figure \mbox{\ref{pic:kmeansClustering} (a)} shows a kernel matrix created without any ordered structure. A rearranged version in Figure \ref{pic:kmeansClustering} (b) clearly shows a block-diagonal structure.
The influence of the scaling parameter can be seen in Figure \ref{pic:kmeansClustering} (c) - (d), where a small scaling parameter leads to a low-rank structure, whereas a large scaling parameter causes a block-diagonal matrix.
Exploiting this structure, a highly memory-efficient representation of the matrix is generated by MEKA, which is capable of handling both the low-rank and the block-wise scenario.
The algorithm is structured as follows:
\paragraph{Part 1.} Applying the c-means algorithm on a random subset of $X$ to compute $c$ cluster centers and assign the remaining data points to the nearest cluster. This results in $c$ cluster centroids $\cc_1, \cc_2,\dots,\cc_c$ with an associated set of indices per cluster $\mathcal{C}_1, \mathcal{C}_2, \dots, \mathcal{C}_c$ leading to $c$ disjoint sets $V_{1}, V_{2}, \dots ,V_{c}$, where $V_{i} = \{\x_j \in \R^d \;|\;  j \in \mathcal{C}_i\}$ and $V_{i} \cap V_{j} = \emptyset$ for $i \ne j$, hence $X = \bigcup_{i = 1}^c V_{i}$.

\paragraph{Part 2.} Each block on the diagonal $\K^{i} = k(\x_p,\x_q)\; \forall \x_p,\x_q \in V_i$ has a low-rank structure as shown by \citet{MEKA}, so a \Nys{} approximation is conducted to build a low-rank approximation $\K^{i} \approx \Q^{i}\left(\Q^{i}\right)^T$ according to Equation \eqref{eq:Nys3}.
The block-diagonal of $\K$ is represented as $\K_{blk\,diag} \approx \Q \Q^T, \; \text{with } \Q = \bigoplus_{i = 1}^c \Q^{i}$, using the notation of \citet{MEKA}, this provides a block-kernel approximation with zeros off-diagonal blocks. \citet{MEKA} proposed several strategies to determine a proper rank per block: 
1) Using the same rank $k$ in each \Nys{} approximation. This results in a fast algorithm but could lead to a high approximation error. In general, the overall target rank is $c k$ of the MEKA approximation.
2) Distribute the rank according to singular value contribution. Each block gets a rank $k_i$ such that $\sum_{i = 1}^{c} k_i = ck$ depending on the singular value contribution to the first $c k$ singular values of $\K_{blk\,diag}$. This strategy decreases MEKA's speed but results in a better approximation error. The overall target rank is $ck$.
3) Using a rank $k_i$ proportional to the ratio of the cluster size and the number of all samples $n$, namely $k_i = \frac{|V_i|}{n}k$ so that $\sum_{i = 1}^{c} k_i = k$. This results in an overall rank of $k$. In accordance with \citet{MEKA}, we follow this strategy.

\paragraph{Part 3.}\label{part3}
Since many relevant off-diagonal blocks are present in the low-rank cases, \citet{MEKA} suggested a highly efficient least squares approximation of such blocks $\K^{i,j} = k(\x_p,\x_q)\; \forall \x_p \in V_i, \forall \x_q \in V_j$. They proposed to subsample the original block $\K^{i,j}$ with random row and column indices sets $\mathcal{I}_i, \mathcal{I}_j$ to get $\Ksu^{i,j} = \K_{\mathcal{I}_i,\mathcal{I}_i}$ and to solve:
\begin{equation*}\label{eq:LeastSquareApp}
	\argmin{\Li^{i,j}} \left\|\K^{i,j} - \Q^{i}\Li^{i,j}\left(\Q^{j}\right)^T \right\|_F, \;\text{where } i \ne j,
\end{equation*}
with a closed form solution:
\begin{equation}\label{eq:SolutionLS}
	\Li^{i,j} = ((\Q^{i}_{\mathcal{I}_i})^T\Q^{i}_{\mathcal{I}_i})^{-1}  (\Q^{i}_{\mathcal{I}_i})^T \Ksu^{i,j}\Q^{j}_{\mathcal{I}_j}((\Q^{j}_{\mathcal{I}_j})^T\Q^{j}_{\mathcal{I}_j})^{-1}
\end{equation}
to finally approximate $\K$ by $\Kpp = \Q\Li\Q^T$. Algorithm \ref{alg:MEKA} provides a detailed procedure of MEKA.
\begin{algorithm}[ht]
	\caption{Memory Efficient Kernel Approximation (MEKA), adapted from \cite{MEKA}}
	\label{alg:MEKA}
	\begin{algorithmic}
		\Require {Data points in $X$, $\mathcal{X} = \R^{\times d}$, kernel parameter $\gamma$, target rank $k$, number of clusters $c$ and percentage of blocks to be truncated}
		\Ensure {Matrices $\Q \in \R^{n \times c k},\Li \in \R^{c k \times c k}$ }
		\State Run c-means to obtain $V_1, V_2, \dots ,V_c$\Comment{Part 1}
		\For{$i \leftarrow 1 \;\text{to}\; c$}\Comment{Part 2}
		\State Approximate $\K^{i}$ with rank $k_i$ by Equation \eqref{eq:Nys3}
		\State $\Q \leftarrow \Q \oplus \Q^{i}$
		\State $\Li \leftarrow \Li \oplus \I \in \R^{k \times k}$
		\EndFor
		\ForAll{$(i,j)(i \ne j)$} \Comment{Part 3}
		\If{Clusters too dissimilar}
		\State $\Li^{i,j} \leftarrow \mathbf{0}$
		\State Continue
		\EndIf
		\State Obtain $\Li^{i,j}$ by Equation \eqref{eq:SolutionLS}
		\EndFor
	\end{algorithmic} 
\end{algorithm}
\begin{figure*}[ht]
	\centering
	\begin{minipage}{0.32\textwidth}
		\centering
		\includegraphics[width=\textwidth]{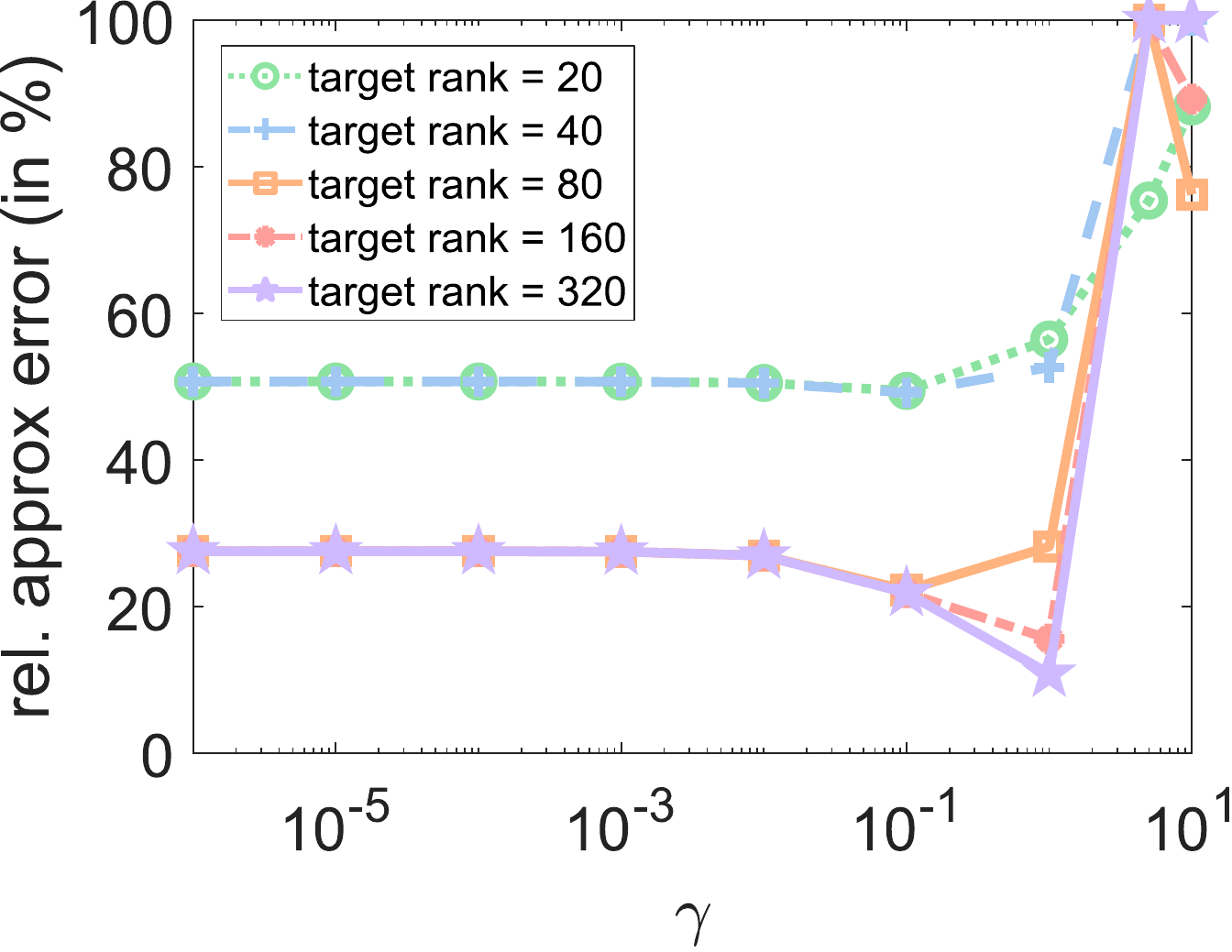}\\\hspace{0.7cm}(a)
	\end{minipage}
	\begin{minipage}{0.32\textwidth}
		\centering
		\includegraphics[width=\textwidth]{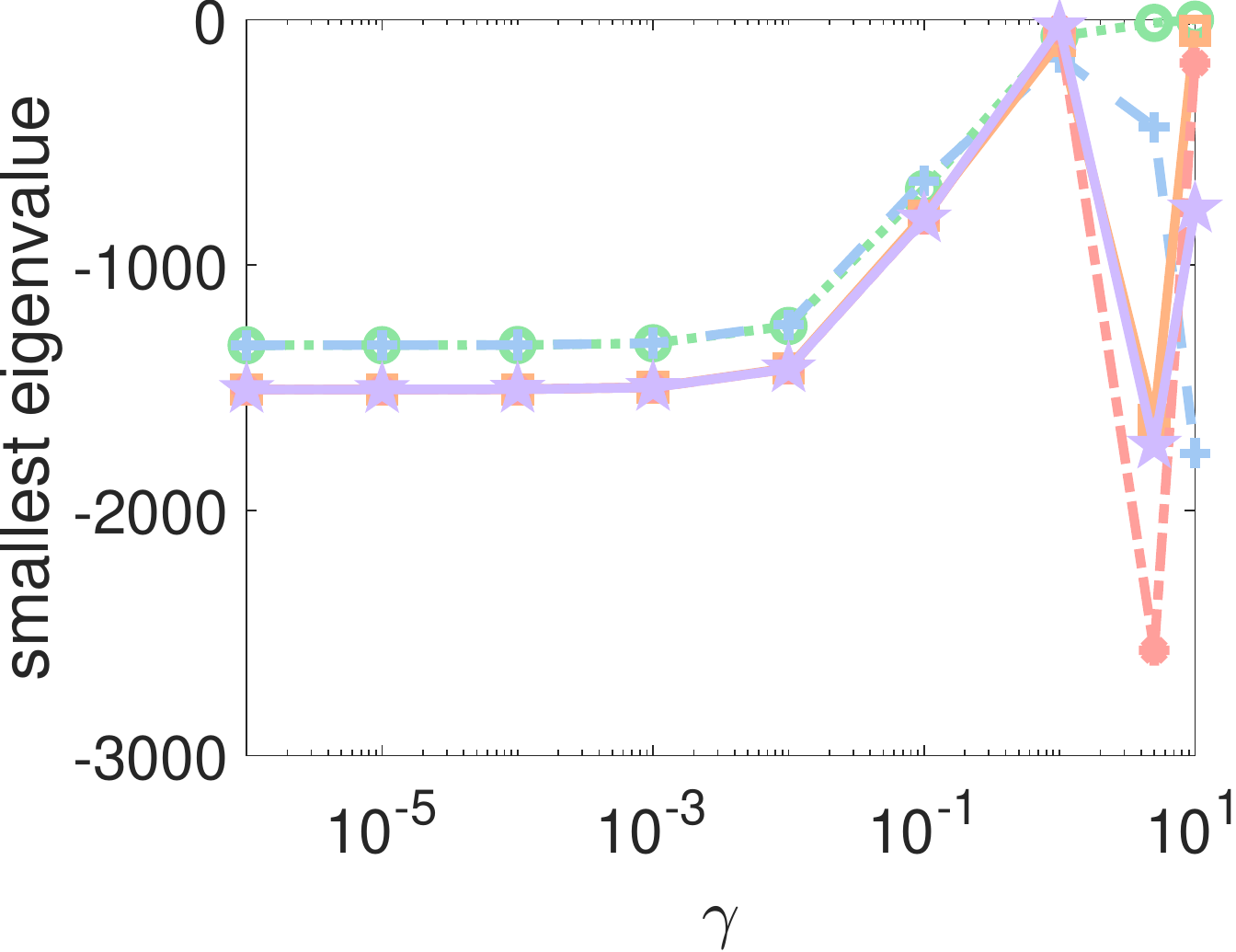}\\\hspace{0.7cm}(b)
	\end{minipage}
	\begin{minipage}{0.32\textwidth}
		\centering
		\includegraphics[width=\textwidth]{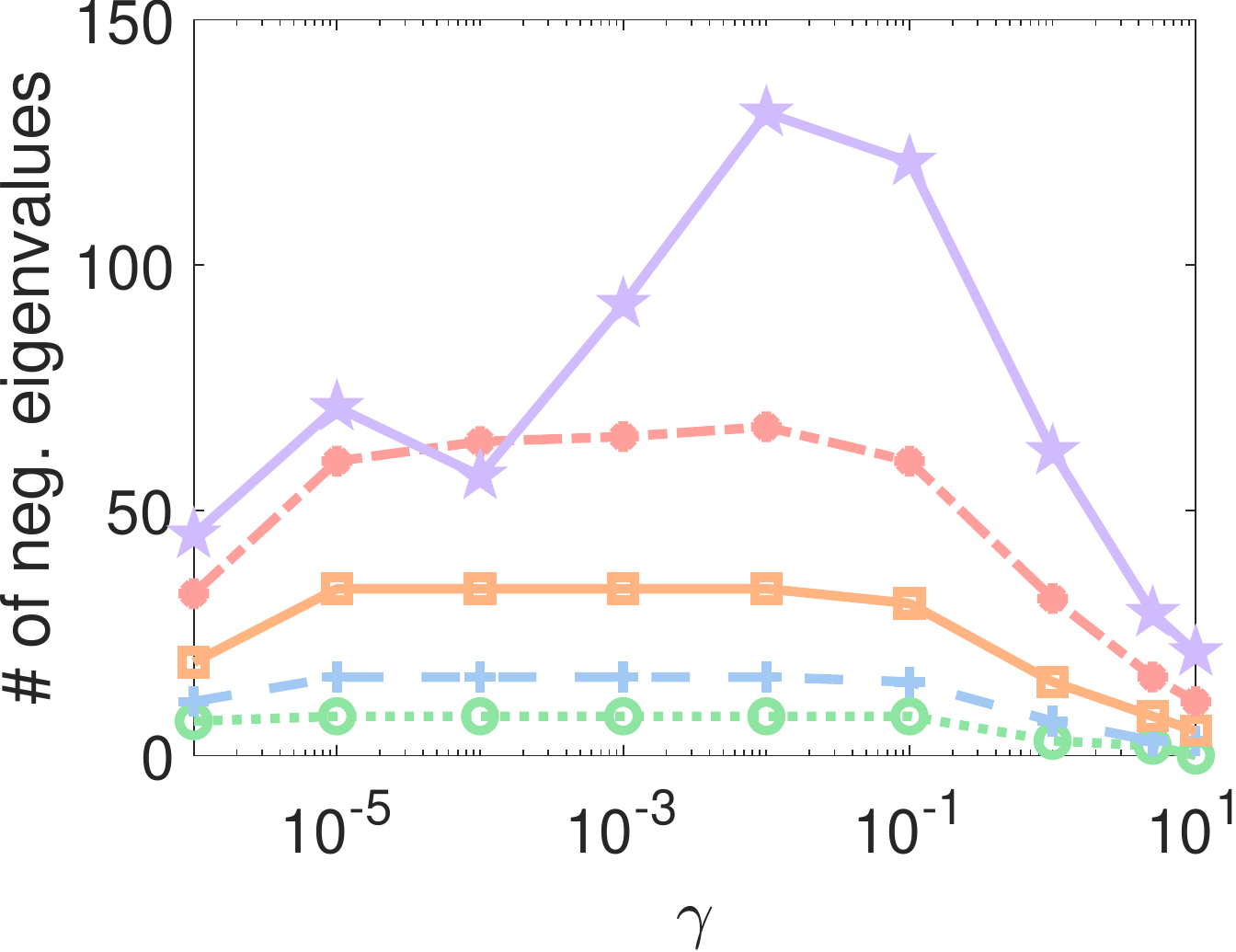}\\\hspace*{0.7cm}(c)
	\end{minipage}
	\caption{MEKA experiments with \textsf{pendigit} data set with $c=10$ and Gaussian rbf kernel. (a) illustrates the relative approximation error according to the target rank and the scaling parameter. Note that after $\gamma > 0.1$, the matrix tends approach the full rank, hence, the approximation error worsens and is limited to 100\%. (b) shows the value of the largest negative eigenvalue. (c) presents the total number of negative eigenvalues.\label{pic:rbf}}
\end{figure*}
\subsection{Non-Stationary and Indefinite Kernels}
\citet{MEKA} assumed the input data to originate from a real vector space and accessed the Hilbert space via a kernel function. Furthermore, their theorems and error-bounds are based on the concept of shift-invariant functions. 
In order to ensure a wider applicability of MEKA, an extension to non-rbf like kernels is necessary.
By applying Proposition \ref{prop:invariant} to the input space, non-stationary kernel functions become shift-invariant.
While MEKA considers the effect of the Gaussian rbf scaling parameter in the light of a shift-invariant kernel function, Proposition \ref{prop:invariant} allows a significant increase of freedom to select a kernel function
\footnote{Further visualisations of non-stationary matrices with a clear block-wise structure are given in the supplements.}. 
However, as mentioned above, kernels such as the polynomial, arc-cosine or neural tangent kernel can become indefinite by applying this normalization (Remark 1) to the input space or the kernel function \citep{pennington2015spherical,liu2021fast}. This leads to the question of how to approximate indefinite kernels with MEKA.\\
Therefore, the crucial step of MEKA is the approximation of each diagonal block computed by \Nys{}. This can be accomplished due to \citet{oglic2019scalable},
who showed that \Nys{} methods are able to handle both psd and indefinite kernel matrices within a bounded error.
\subsection{Occurrence of Negative Eigenvalues}\label{sec:experiments}
\begin{figure*}
\centering
	\begin{tabular}{c c c}
	   \includegraphics[width=0.3\textwidth]{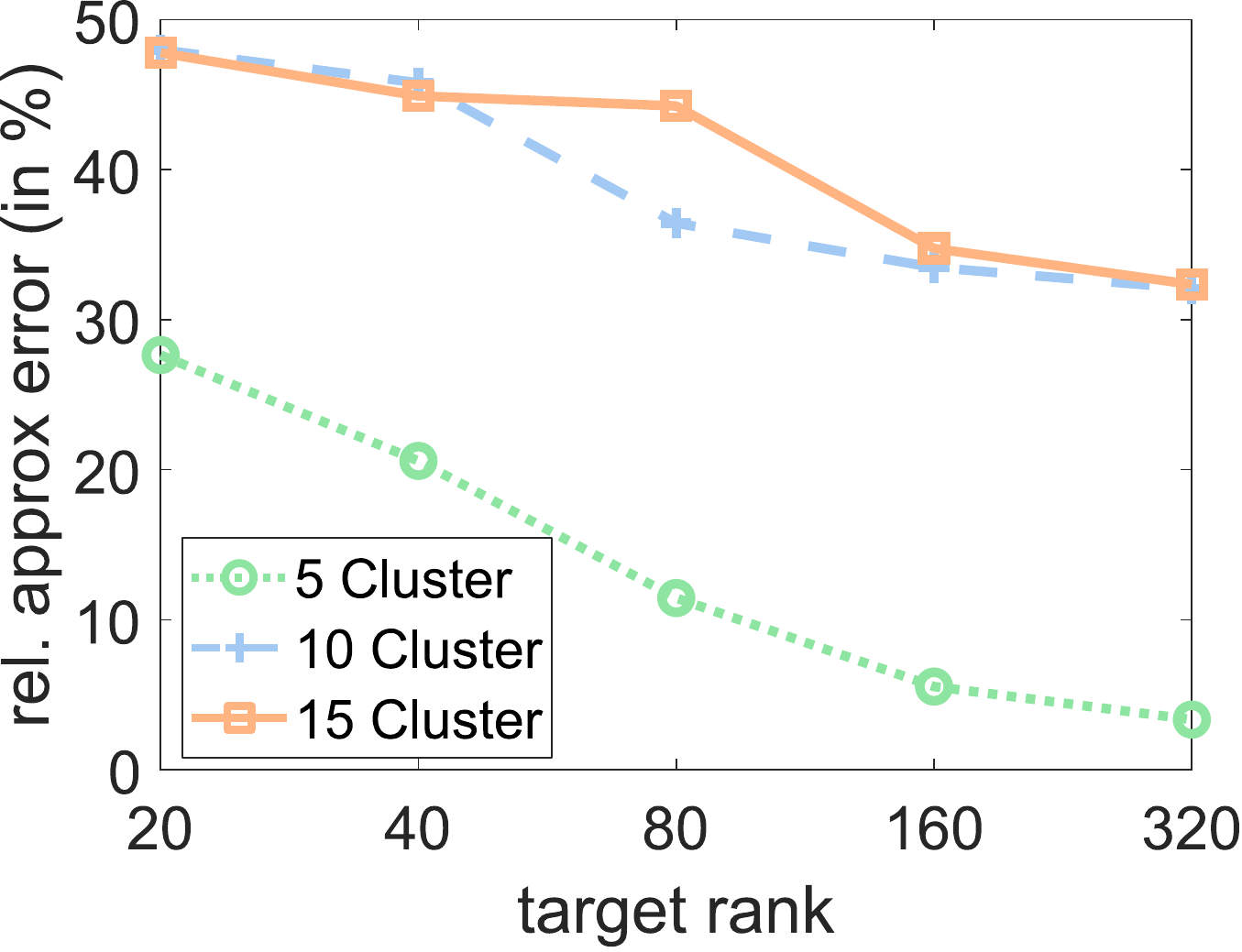}  &  \includegraphics[width=0.3\textwidth]{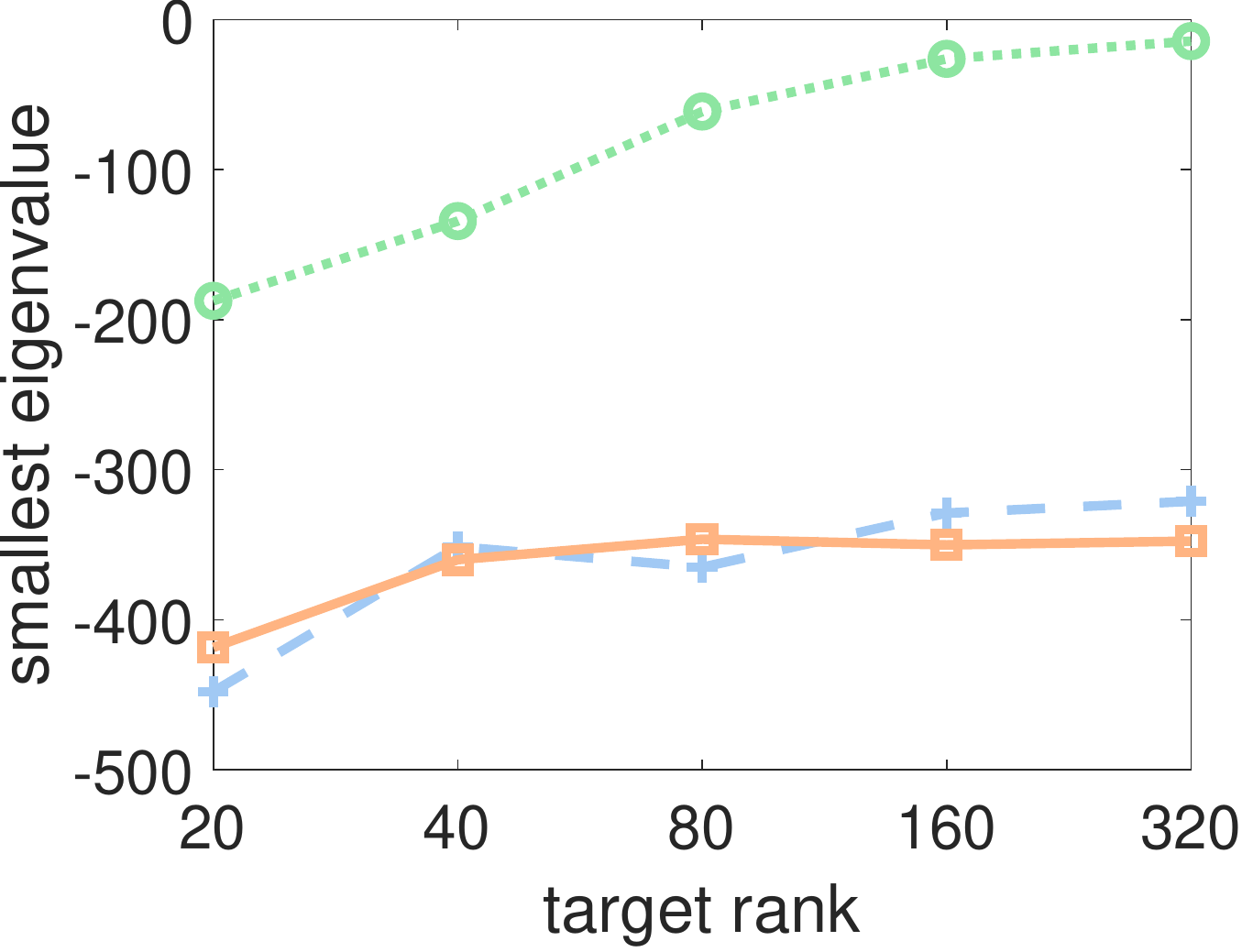}
	     & \includegraphics[width=0.3\textwidth]{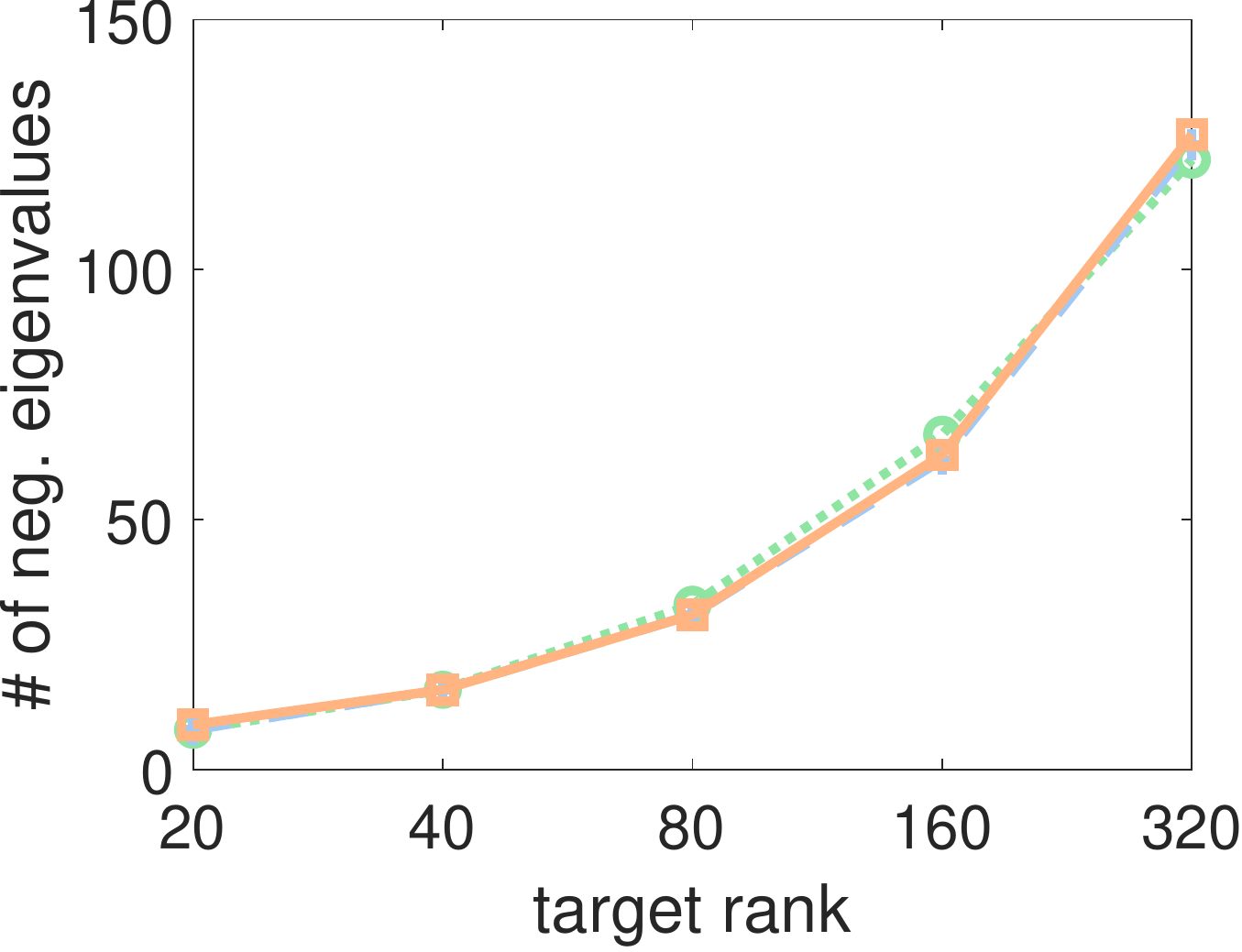}\\
	     \hspace{0.5cm}(a)&\hspace{0.7cm}(b)&\hspace{0.5cm}(c)\\
	\end{tabular}
	\caption{MEKA experiments with \textsf{gesture} data set and extreme learning kernel. (a) shows the relative approximation error along the different target ranks and numbers of clusters. (b) presents the value of the smallest eigenvalue and (c) shows the total number of negative eigenvalues.\label{pic:elm}}
\end{figure*}
As \citet{MEKA} mentioned, the subsampling strategy of the off-diagonal block approximation (Part 3 of MEKA) potentially results in a non-psd approximation. In the following, we critically discuss the importance of  MEKA's hyperparameters with respect to indefiniteness.
Following \citet{MEKA}, we first focus on the Gaussian rbf kernel, and then we apply Proposition \ref{prop:invariant} to show the approximation capabilities on inner product based kernels, in particular, the extreme learning kernel (ELM):
\begin{equation}\label{eq:elmkernel}
\resizebox{0.48\textwidth}{!}{$k_{elm}(\x,\y) = \frac{2}{\pi}\arcsin\left(\frac{1 + \langle \x,\y \rangle}{\sqrt{(\frac{1}{2\sigma^2_w}+1+\langle\x,\x\rangle)(\frac{1}{2\sigma^2_w}+1+\langle\y,\y\rangle)}}\right)$}
\end{equation} from \citet{frenay2011parameter}, as a parameter-insensitive Gaussian rbf kernel alternative independent of $\sigma_w$.

\begin{table}
	\caption{Overview of data set characteristics.\label{tab:datasets}}
	\begin{center}
	\begin{tabular}{c|c|c|c}
		\textbf{Data set}& \textbf{ID} & \textbf{Samples} $n$ & \textbf{Attributes} $d$  \\
		\hline&&&\\
		spambase&1& 4,601 & 57 \\
		artificial 1&2& 7,500 & 2 \\
		cpusmall &3& 8,192 & 12\\
		gesture &4& 9,873 & 32\\
		artificial 2&5& 10,000 & 15 \\
		pendigit&6 & 10,992 & 16\\
	\end{tabular}
	\end{center}
\end{table}
In Table \ref{tab:datasets}, the utilized data sets are listed. We retrieved the \textsf{gesture} data set from \cite{madeo2013gesture} and \textsf{spambase} from the UCI repository\footnote{https://archive.ics.uci.edu/ml/index.php} and \textsf{cpusmall} and \textsf{pendigit} are obtained from the LIBSVM repository\footnote{https://www.csie.ntu.edu.tw/\~{}cjlin/libsvmtools/datasets/}. We also included two artificial data sets created with scikit-learn to demonstrate performance with overlapping clusters. Within the experiments, we collect the following statistics: relative approximation error $\|\K - \Kpp\|_F / \|\K\|_F$, the smallest eigenvalue and the number of negative eigenvalues. Note that for reproducibility, a random generator seed has been set and the calculations were executed with Matlab R2020a on a machine with Intel$^\text{\textregistered}$ Xeon W-2175 CPU @ 2.50GHz $\times$ 28.
The code of the extended MEKA implementation and the experimental setups of all experiments in this work can be found at: \textit{https://github.com/simonheilig/indefinite-meka}.

Figure \ref{pic:rbf} presents an analysis of MEKA with respect to the scaling parameter $\gamma$ of the Gaussian rbf kernel and the target rank of the approximation. To follow the line of \citet{MEKA}, the rank distribution strategy relative to the cluster size is utilized in this experiment. It is apparent that a substantial negative eigenspace is present for $10^{-6} \le \gamma \le 1 $. For $\gamma > 1$, the matrix is close to full rank, resulting in an increasing approximation error. Figure \ref{pic:rbf} (c) shows a direct correlection between the target rank of the approximation and the number of negative eigenvalues.\\
Note that the results for the remaining data sets are given in the supplements, because they show similar behavior except for small changes.

Figure \ref{pic:elm} shows the results for the \textsf{gesture} data set and the extreme learning kernel. Since the kernel performance is invariant to its parameter (see Equation \eqref{eq:elmkernel}), the development along with the target rank and the number of clusters is presented. Figure \ref{pic:elm} (a) shows the connection between the relative approximation error and the target rank of the approximation.\\Obviously, an increasing target rank results in a lower approximation error. Furthermore, an increasing number of clusters leads to higher approximation errors, since more off-diagonal blocks are present in those cases. Figure \ref{pic:elm} (b) reflects a similar behavior, as more off-diagonal blocks require many subsampling steps in part 3 of MEKA. Finally, Figure \ref{pic:elm} (c) shows a positive correlation between the target rank and the number of negative eigenvalues.

\section{REVISION}\label{sec:revision}
From the analyses above, it becomes evident that a psd-correction is required to apply MEKA to kernel methods in a variety of optimization frameworks. This is necessary due to psd violations in the original MEKA approximation, but also if more general kernel functions are used, which are non-psd.
\citet{MEKA} mentioned briefly that psd violations may occur but claimed it as a rare event. They proposed to clip the link matrix $\Li$, which leads to the closest psd version of the approximation on the cone of all psd matrices under the Frobenius norm \citep{schleif2015indefinite}.
Since this matrix is $ck\times ck$, it is prohibitive to obtain an eigenvalue decomposition if there is a large number of clusters, but if the negative eigenvalues are due to noise, clip is the method of choice.
As shown previously, indefinite contributions may occur regularly, but we cannot yet ensure whether this is exclusively due to noise or not.
The second approach, suggested by \citet{MEKA}, is to apply an appropriate shift on the diagonal of the entire approximation without specifying how to determine a shift parameter while maintaining the time and memory efficiency.
In particular, if the indefinite part contains relevant information, a shift correction is a better choice \citep{DBLP:conf/sspr/DuinP10,10.3389/fams.2020.553000}.
\\Given a computed shift parameter $\lambda_{shift}$ and the smallest eigenvalue $\lambda_{min}$ of $\Kpp$, the factorized MEKA approximation can be efficiently corrected by $\Kpp = \Q\Li\Q^T + \lambda_{shift}\I$, where $\lambda_{shift} \ge |\lambda_{min}|$ and $\I$ as identity matrix, since the shift operation on the diagonal of the similarity matrix causes an analogous shift of the eigenspectrum of the respective matrix \citep{DBLP:journals/pr/LaubRBM06,schleif2015indefinite,DBLP:conf/esann/Loosli19,munch2020structure}.
Note that it is prohibitive to determine this shift parameter by an eigenvalue decomposition, as it violates the runtime and memory complexity constraint.
The out-of-sample extension is discussed in the supplements.

A well-known method to calculate only a few of the eigenvalues, in particular the extreme values, is given by the \textit{Lanczos iteration} in a very efficient manner \citep{cullum2002lanczos}. The essential characteristic of this approach is that it works with a matrix-times-vector multiplication, since it would be prohibitive to build up kernel matrices entirely in the large-scale setting. The factorization obtained with MEKA enables a matrix-times-vector multiplication $\y = \Kpp \x$ within a time complexity proportional to the memory requirement $O(n k + (c k)^2)$ \citep{MEKA}.\\ See Table \ref{tab:clipshiftCosts} for an overview of the complexities of the correction strategies, where $n$ is the original number of rows or columns, $m$ the remaining number of eigenvalues and eigenvectors after truncation, $j$ the number of Lanczos iterations (usually $j \approx 2\sqrt{n}$ is enough \citep{parlett1982estimating}) and $T$ as the complexity of matrix-times-vector multiplication (for MEKA $ T = O(j(nk + (ck)^2) + j\ln{j})$) \citep{mahadevan2008fast,coakley2013fast}.\\The advantage of a Lanczos-based shift is to allow any matrix representation, as long as the matrix-times-vector multiplication can be guaranteed. By utilizing the MEKA factorization of the kernel matrix, it is unnecessary to keep the entire matrix in memory, thus maintaining memory efficiency.\\
\begin{table}
    \caption{Overview of cost complexities of different correction methods and their storage requirements (n.d. = not defined). \label{tab:clipshiftCosts}}
    \begin{center}
    \begin{tabular}{c|c|c|c}
        \textbf{Type} & \textbf{Clip} & \textbf{Shift} & \textbf{Lanczos-Shift}\\
        \hline&&&\\
        Runtime & $O(n^3)$ & $O(n^3)$ & $O(jT + j\ln{j})$\\
        Storage & $O(nm + m^2)$ & $O(n^2)$ & n.d.\\
    \end{tabular}
    \end{center}
\end{table}
\subsection{Lanczos Iteration}\label{sec:lanczos}
The Lanczos iteration aims to minimize the following optimization problem (also known as the Rayleigh quotient):
$
	\lambda_{min} = \min_{\x \ne 0}\frac{\langle\x,\,\A\x\rangle}{\langle\x,\,\x\rangle},
$	
with a symmetric matrix $\A \in \R^{n \times n}$ and its smallest eigenvalue $\lambda_{min}$ \citep{cullum2002lanczos}.
Note that situations occur where the algorithm needs to be restarted or a reorthogonalization is required.
However, there exist approaches without the need for such reorthogonalization, by identifying the eigenvalues which lead to the linear dependencies in the Lanczos vectors and ignoring them \citep{cullum2002lanczos}. This is particularly important in our scenario to keep a good runtime performance. The convergence rate is heavily related to the overall stiffness of the matrix, but usually, the extreme Ritz values are obtained after $i \approx 2\sqrt{n}$ iterations, leading to good approximations of the corresponding eigenvalues of $\A$ \citep{parlett1982estimating}.

\subsection{Error Bound Analysis}
\begin{table*}
	\caption{Classification results using a SVM with L1QP solver on the original and unapproximated matrix (M0), approximated matrix before (M1) and after correction (M2) by a 10-fold CV accuracy and standard deviation in \% (n.c. $=$ not converged). Typical memory costs of the approximation in (Mem). Experiments for M0$^{*}$  are conducted on a shift corrected matrix.
		\label{tab:revision_experiments}}
	\begin{center}
		\small{
			\begin{tabular}{c|c|c|c|c|c|c|c|c|c|c|c|Hc}
				\multicolumn{2}{c|}{}&\multicolumn{3}{c|}{\textbf{Shift Invariant}}&\multicolumn{6}{c|}{\textbf{Non-Stationary}}&\multicolumn{3}{c}{\textbf{Indefinite}}\\
				\multicolumn{2}{c|}{}&\multicolumn{3}{c|}{\textbf{RBF}}&\multicolumn{3}{c|}{\textbf{ELM}}&\multicolumn{3}{c|}{\textbf{Poly}}&\multicolumn{3}{c}{\textbf{TL1}}\\
				\hline
				\textbf{ID}& \textbf{Mem}  & \textbf{M0} & \textbf{M1} & \textbf{M2}&\textbf{M0} & \textbf{M1} & \textbf{M2}&\textbf{M0} & \textbf{M1} & \textbf{M2}&\textbf{M0}* & \textbf{M1}* & \textbf{M2}\\
				\hline
				&&&&&&&&&&&&\\
				\multirow{2}{*}{1} & \multirow{2}{*}{4.14 \%}& 93.63 & \multirow{2}{*}{\textbf{n.c.}}    & 88.85&95.09&\multirow{2}{*}{\textbf{n.c.}}&81.81& 91.48 & \multirow{2}{*}{\textbf{n.c.}} & 73.66 & 94.41&&85.57\\
				&&($1.27$)&&($4.32$)&($0.81$)&&($5.17$)&($1.92$)&&($7.40$)&($1.29$)&&($3.74$)\\[0.1cm]
				
				\multirow{2}{*}{2} &\multirow{2}{*}{1.11 \%}& 89.19 & 85.48 & 89.23&88.60&82.20&88.63& 82.43 & \multirow{2}{*}{\textbf{n.c.}} & 82.49 & 78.96&&89.31\\
				&&($1.34$)&($11.93$)&($1.42$)&($1.32$)&($17.63$)&($1.37$)&($1.15$)&&($1.03$)&($13.32$)&&($0.50$)\\[0.1cm]
				
				\multirow{2}{*}{3} &\multirow{2}{*}{0.30 \%}&  89.42 & \multirow{2}{*}{\textbf{n.c.}}   & 86.47&89.01&\multirow{2}{*}{\textbf{n.c.}}&86.93& 82.46 & \multirow{2}{*}{\textbf{n.c.}} & 77.27& 89.22&&88.53\\
				&&($0.91$)&&($1.32$)&($1.15$)&&($1.29$)&($1.81$)&&($1.76$)&($1.26$)&&($1.48$)\\[0.1cm]
				
				\multirow{2}{*}{4} & \multirow{2}{*}{0.18 \%}&   58.99    & \multirow{2}{*}{\textbf{n.c.}}    &44.13&68.02&\multirow{2}{*}{\textbf{n.c.}}&45.50& 64.18 & \multirow{2}{*}{\textbf{n.c.}} & 38.57 & 54.78&\multirow{2}{*}{\textbf{n.c.}}&45.56\\
				&&($0.95$)&&($1.13$)&($2.45$)&&($1.23$)&($1.75$)&&($1.53$)&($1.51$)&&($0.95$)\\[0.1cm]
				
				\multirow{2}{*}{5} & \multirow{2}{*}{1.75 \%}&  88.79     & 72.77 &88.76&88.74&\multirow{2}{*}{\textbf{n.c.}}&85.71& 88.79 & 88.79 & 88.79 & 88.66&&88.61\\
				&&($1.08$)&($24.32$)&($0.98$)&($1.35$)&&($1.87$)&($0.95$)&($0.95$)&($0.95$)&($0.83$)&&($1.00$)\\[0.1cm]
				
				\multirow{2}{*}{6} & \multirow{2}{*}{2.18 \%}&  99.58& 21.04& 87.79 & 99.52& 10.37& 83.26& 99.52 & 39.23 & 98.01 & 99.18&&96.53\\
				&&($0.10$)&($12.88$)&($2.54$)&($0.18$)&($0.00$)&($1.96$)&($0.29$)&($32.40$)&($0.56$)&($0.26$)&&($0.71$)\\
			\end{tabular}
		}
	\end{center}
\end{table*}
Given the error bound of Theorem 3 of \cite{MEKA}:
\begin{align*}
    &\|\K - \Kpp\|_F \leq \|\K - \K_k\|_F  + \left(\frac{64k}{l}\right)^{\frac{1}{4}}n\K_{max}(1 +\theta)^{\frac{1}{2}}\\&\qquad\qquad\qquad+ 2\|\Delta\|_F
\end{align*}
where $\theta = \sqrt{\frac{n-l}{n-0.5}\frac{1}{\beta(l,n)}\log{\frac{1}{\delta}}}d_{max}^{\K}/\K^{\frac{1}{2}}_{max}$;  $\beta(l,n)=1-\frac{1}{2\max\{l,n-l\}}$; $\K_{max} = \max_i\K_{(i,i)}$; and $d_{max}^{\K}$ represents the distance $max_{ij}\sqrt{\K_{(i,i)} + \K_{(j,j)} - 2\K_{(i,j)}}$.
The error bound for the corrected MEKA (assuming that $\K,\Kpp$ are non-psd) is:
\begin{small}
\begin{align*}
    &\|(\K - \Kpp) + \lambda_{shift}\mathbf{I}\|_F \\&= \|(\K^+ - \Kpp^+) + (-\K^- + \Kpp^-)+ \lambda_{shift}\mathbf{I}\|_F\\
    &\leq \|\K^+ - \Kpp^+\|_F + \|-\K^- + \Kpp^-\|_F + \|\lambda_{shift}\mathbf{I}\|_F\\
    &= \|\K^+ - \Kpp^+\|_F + \|\K^- - \Kpp^-\|_F + \|\lambda_{shift}\mathbf{I}\|_F\\
    &\leq \|\K^+ - \K^+_k\|_F  + \left(\frac{64k}{l}\right)^{\frac{1}{4}}n\K^+_{max}(1 +\theta)^{\frac{1}{2}}+ 2\|\Delta_+\|_F\\&+\|\K^- - \K^-_k\|_F  + \left(\frac{64k}{l}\right)^{\frac{1}{4}}n\K^-_{max}(1 +\theta)^{\frac{1}{2}}+ 2\|\Delta_-\|_F\\&+\sqrt{n}|\lambda_{shift}|
\end{align*}
\end{small}
where $\K = \K^+ - \K^-$ and $\Kpp = \Kpp^+ - \Kpp^-$ are provided by the definition of \Krein{} spaces, which can be decomposed into two orthogonal Hilbert spaces. 
Apparently, the shift parameter linearly affects the error bound and it is important to estimate a reliable $\lambda_{shift}$ as proposed in Section \ref{sec:lanczos}. The value of the smallest eigenvalue is bounded by the row sums of the matrix, as given by Gerschgorin disks \citep{gerschgorinDisks}.
\subsection{Classification Experiments}
To study the effect of the Lanczos-shift, we compare our proposed correction on a kernel SVM with a classical QP solver. Four kernel functions are considered: the Gaussian rbf kernel, polynomial kernel, extreme learning kernel, and an instance of an indefinite kernel, the shift invariant truncated Manhattan kernel (TL1):
\begin{equation*}\label{eq:tl1}
    k_{tl1}(\x,\y)= \max\{\rho - \|\x-\y\|_1,0\}
\end{equation*}
The TL1 kernel is successfully used for local training and is less sensitive to its hyperparameter \citep{tl1Suykens,mehrkanoon2018indefinite}. According to \citet{tl1Suykens}, $\rho = 0.7d$ is superior to some hyperparameter optimized Gaussian rbf kernel results. The other kernel parameters are also optimized using a grid search $\gamma \in \{1,2,\hdots,10,15\}$ and $a \in \{2,3\}$, $p \in \{2,4,8,10\}$.
The $c, k$ parameters of MEKA are determined to have the lowest approximation error with minimal memory usage and the soft-margin parameter $C$ of the SVM is chosen to minimize the classification loss of the original kernel SVM using a 5-fold cross-validation (CV) in a grid of $C \in \{10^{-3},1,10,100,1000\}$. 
For comparison purposes, the rank distribution strategy with the same rank per cluster is used.
Further details on preprocessing, typical hyperparameters and a brief summary of utilized kernels is given in the supplements.
For the TL1 kernel, the original kernel matrix is non-psd and thus the parameter optimizations are executed on a shift corrected matrix fulfilling the psd constraint.\\
In Table \ref{tab:revision_experiments}, the classification accuracy of a 10-fold CV (with a nested 5-fold CV for parameter estimation) of our Lanczos-shift correction is compared to the original kernel SVM and the original MEKA approximation. Also, the typical memory consumption of the factorization provided by MEKA is given relative to the original matrix. The approximation quality can be found in the supplementary files.

The results show that the frequent negative eigenvalues (see Sec. \ref{sec:experiments}), result in common optimization problems, occurring with indefinite kernels. 
The shift correction effectively solved this issue. Comparing the psd results M0 (exact) for \{RBF,ELM,Poly\} with M2 (approximated), 
we see that the proposal has often a similar accuracy. In the rare cases where classical MEKA (M1) converged, the proposed solution (M2) is 
consistently better (one case on par). For the indefinite TL1 we always used a shift correction such that no M1 exists. Here we see competitive results of the exact and corrected M0$^*$ compared with the approximated and corrected M2. As expected the memory consumption using
M1 or M2 can be kept very small, with only a few percent of the original space and minimum information loss for M2, whereas M1 (original MEKA)
causes convergence problems due to the non-psd effect. 
In particular, the results show that MEKA is now also effectively applicable to more generic kernel functions using the proposal with the underlying
normalization procedure.

\section{CONCLUSION}\label{sec:conclusion}
We showed that MEKA can be effectively extended to a variety of common kernel functions. 
In particular, we showed that a spherical projection extends the applicability to non-stationary kernels.
Moreover, we analyzed the properties of the approximated matrices and showed that MEKA frequently 
generates non-psd matrices and a correction approach is inevitable. Accordingly, we suggested an efficient shift correction, based on the Lanczos iteration, to fulfill the psd assumption of kernel methods, while maintaining the memory efficiency. This allows us now to approximate indefinite kernel functions with MEKA. 
Our evaluation demonstrated the importance of the psd correction and that the revised MEKA 
approach is not only more flexible, but remains competitive in accuracy and memory consumption.
In particular, the technique overcomes limitations of the original MEKA and 
enables the valid usage of a wide range of kernel functions in large-scale data analysis
as demanded in recent work \citep{Lee2018,DBLP:conf/esann/Loosli19,Liu2020}.
As the revised MEKA technique is now applicable to approximate a variety of kernel functions and leads to consistent approximations, there are other properties of the original kernel function which are not yet preserved. For example, the approximated Gaussian rbf kernel may contain negative similarities, although being psd. This is not meaningful and can be handled by truncations followed by the suggested shift correction. Here it may be desirable to enforce additional constraints in the least squares approximation of the off-diagonal blocks. However, this can not be easily achieved in closed form and will be subject of further research.

\subsubsection*{Acknowledgements}
SH is supported by the ESF (WiT-HuB 4/2014-2020), project KI-trifft-KMU, StMBW-W-IX.4-6-190065. MM is supported by the Bavarian HighTech agenda and the W\"urzburg Center for Artificial Intelligence and Robotics (CAIRO). 

\bibliography{literature}

\end{document}


\onecolumn
\runningtitle{Revisiting Memory Efficient Kernel Approximation}
\aistatstitle{Revisiting Memory Efficient Kernel Approximation: \\
An Indefinite Learning Perspective - \\
Supplementary Material}
\aistatsauthor{ Simon Heilig \And Maximilian M\"unch \And  Frank-Michael Schleif }

\aistatsaddress{University of Bamberg \And UAS W\"urzburg-Schweinfurt,\\University of Groningen \And UAS W\"urzburg-Schweinfurt}
\section{INTRODUCTION}
As derived in the paper, MEKA can be used with a large variety of practical relevant kernel functions as long as some moderate constraints, such as the normalization of the input space and a data compactness hypothesis by means of an inherent block-cluster structure, are fulfilled. Subsequently, we provide additional results and details:
\begin{itemize}
    \item[-] Section \ref{sup:sec:1} provides a brief orientation of the implementation given in the supplement folder.
    \item[-] Section \ref{sup:sec:2} shows additional non-stationary kernel matrices with block-wise structure.
    \item[-] Section \ref{sup:sec:3} provides additional results of negative eigenvalue analysis.
    \item[-] Section \ref{sup:sec:4} presents an out-of-sample extension to the corrected MEKA approximation.
    \item[-] Section \ref{sup:sec:5} presents further experimental information, like hyperparameters and approximation error.
    \item[-] Section \ref{sup:sec:6} presents an overview of utilized kernel functions and their relevant properties.
\end{itemize}

\section{REFACTORED AND EXTENDED IMPLEMENTATION}\label{sup:sec:1}
The code of this paper is written in Matlab and can be found at: \textit{https://github.com/simonheilig/indefinite-meka}. It is structured as follows:

The folder \textit{MEKA} contains all code of the MEKA approximation, which is a refactored version of \cite{MEKA} extended for a generic use of kernel functions with the Lanczos iteration based shift correction (main file: meka.m). Folder \textit{kernels} contains all implemented kernel functions and the folder \textit{experiments} consists of two parts, first the analysis of negative eigenvalues and second, the classification experiments (main files: setupEigenvalues.m, setupClassification.m). The data sets are together with the generation script provided in folder \textit{data}.

\section{BLOCK-WISE STRUCTURE IN NON-STATIONARY KERNELS}\label{sup:sec:2}
\begin{figure}
    \begin{center}
    \begin{tabular}{c c c c}
        \includegraphics[width=3.9cm,height=3.9cm]{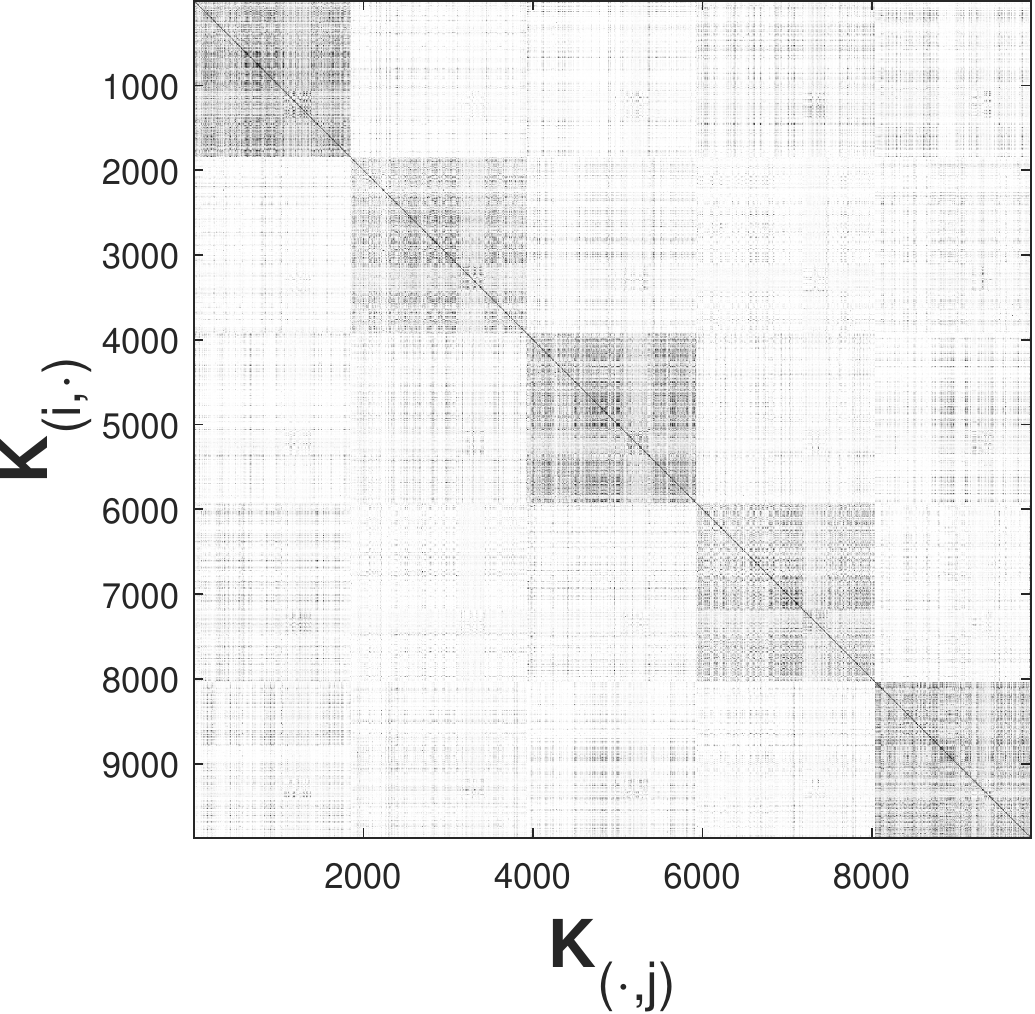}&\includegraphics[width=3.9cm,height=3.9cm]{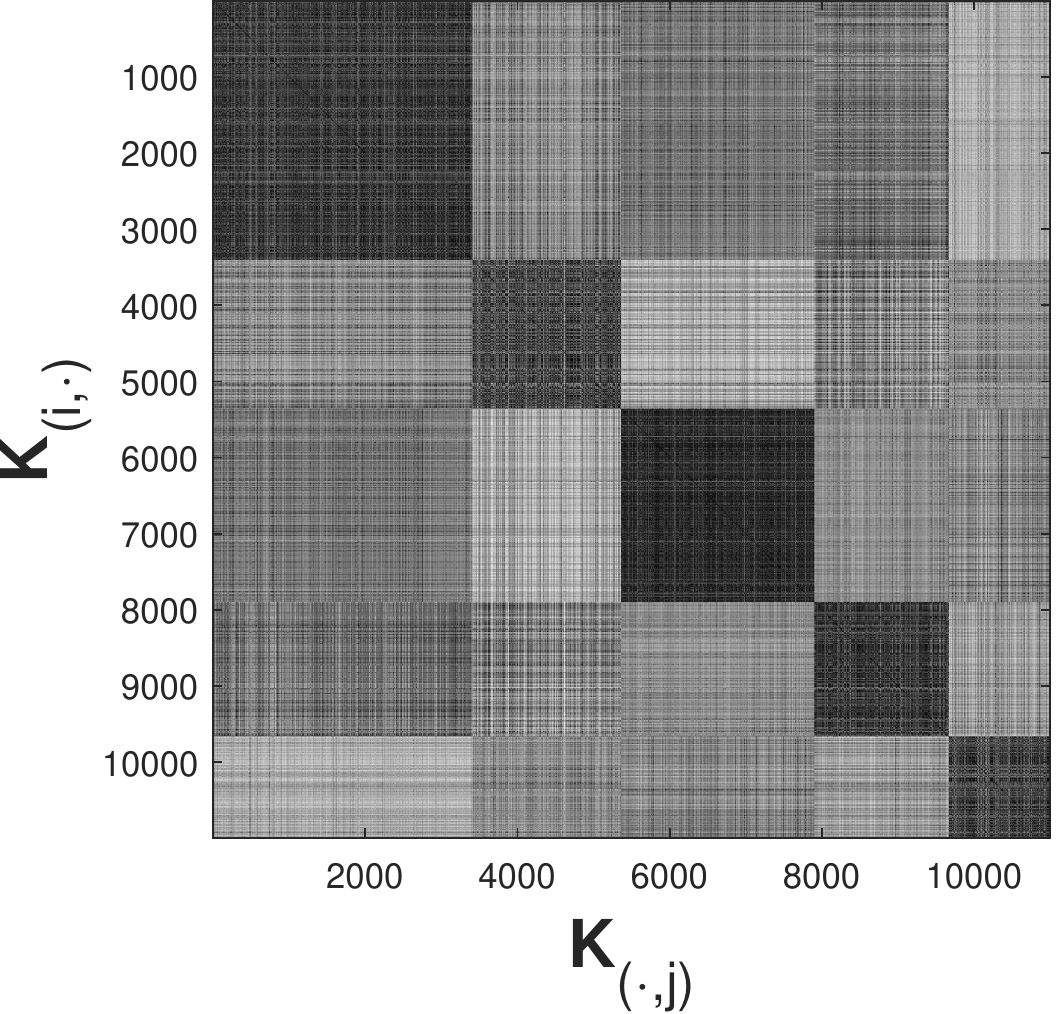} &\includegraphics[width=3.9cm,height=3.9cm]{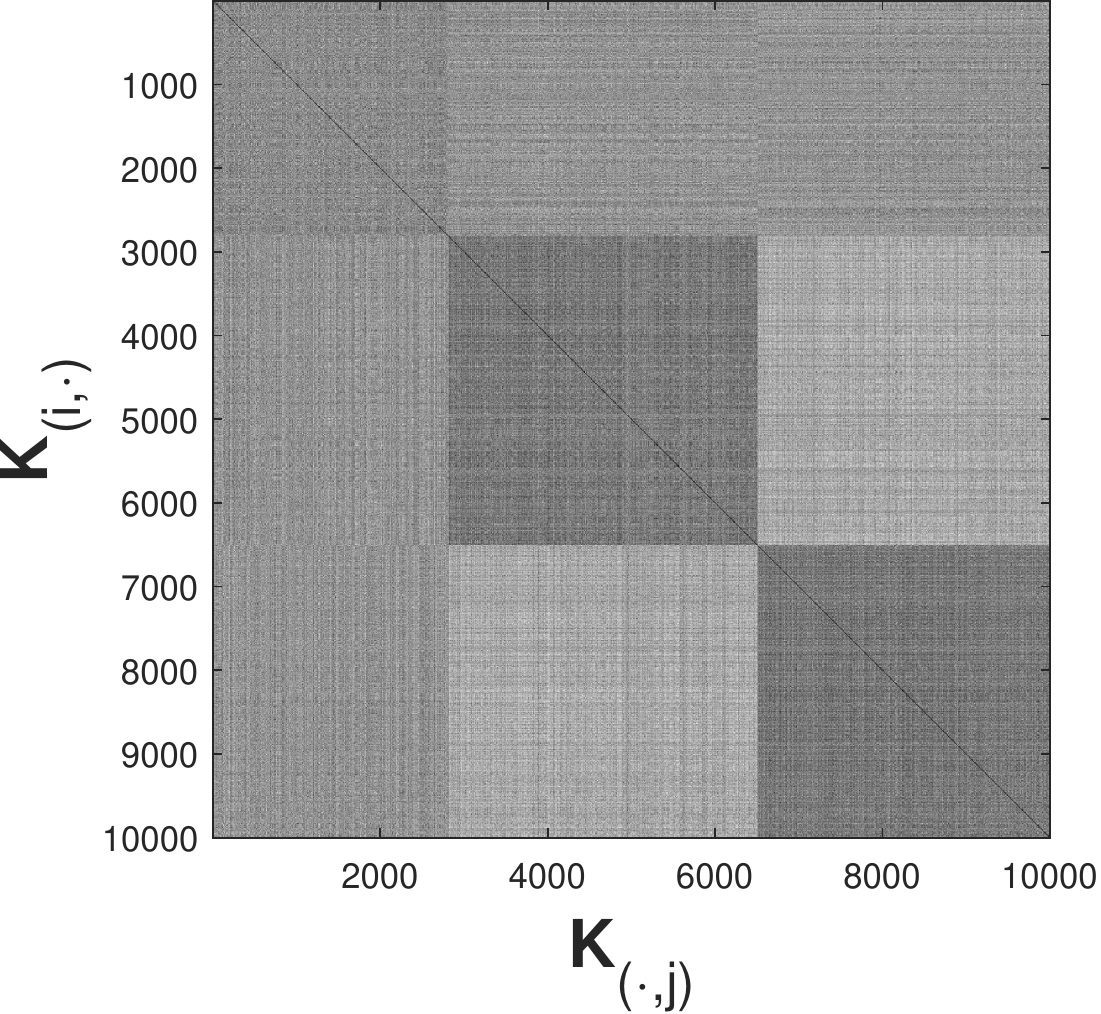}  & \includegraphics[width=3.9cm,height=3.9cm]{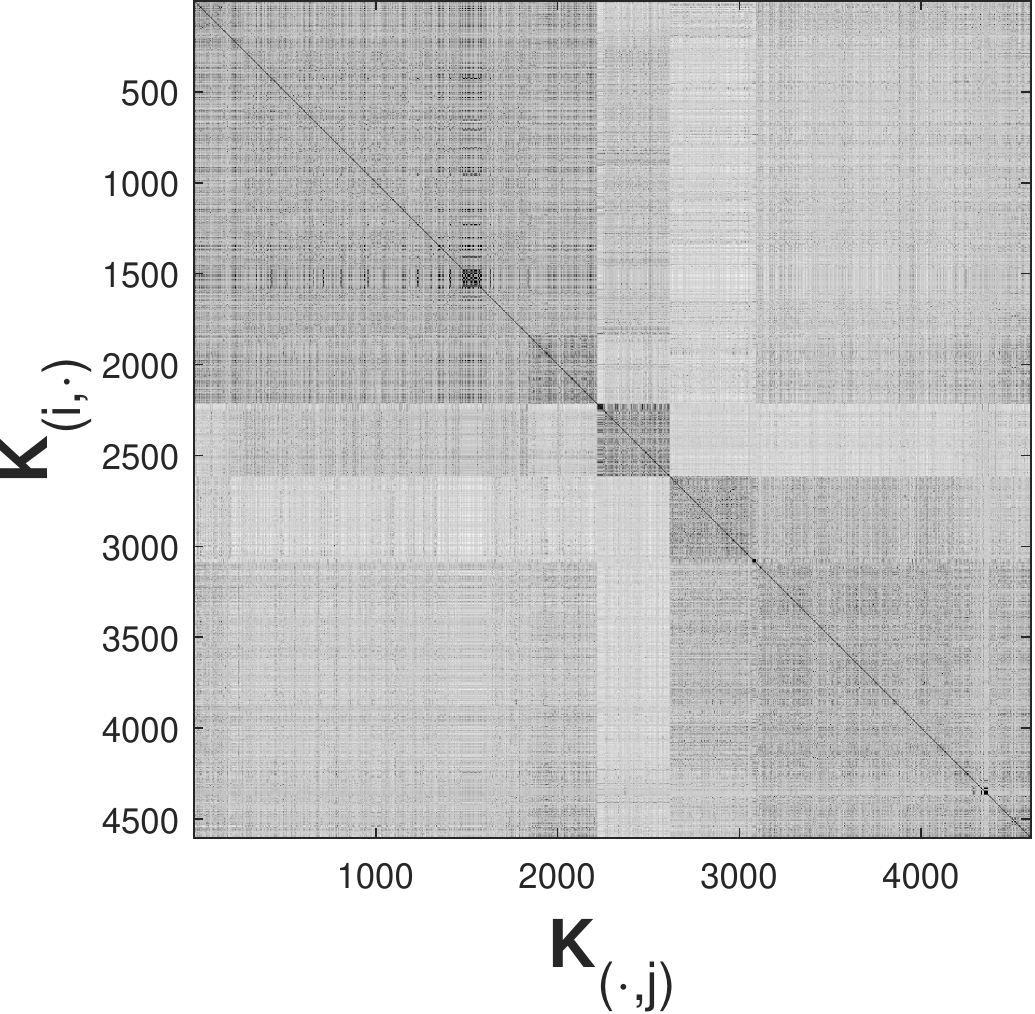} \\
        \hspace*{0.7cm}(a) & \hspace*{0.7cm}(b)&\hspace*{0.7cm}(c) & \hspace*{0.7cm}(d)\\
    \end{tabular}
    \end{center}
    \caption{(a) polynomial kernel of \textsf{gesture}, (b) polynomial kernel of \textsf{pendigit}, (c) extreme learning kernel of \textsf{artificial2}, (d) extreme learning kernel of \textsf{spambase}. Note that the darker the color, the higher the similarity value of the entry in the matrix.}\label{pic:nonstationary}
\end{figure}

In our work, we extended the use of MEKA to non-stationary kernels by applying a projection to the unit sphere in the input space (See Proposition 1 and Section 3.2). Figure \ref{pic:nonstationary} shows an excerpt of the kernel matrices obtained with non-stationary kernel functions. 

It is clearly visible, that a block-wise structure is created after clustering in the input space and thereafter rearranging the respective kernel matrix. Together with the results of the experiments conducted in the main paper, this emphasizes the approximation capabilities of MEKA in case of non-stationary kernel functions. 

\section{OCCURRENCE OF NEGATIVE EIGENVALUES}\label{sup:sec:3}
\begin{figure}[h!]
\centering
	\begin{tabular}{c c c}
	&\hspace{0.7cm}\includegraphics[width=0.15\textwidth]{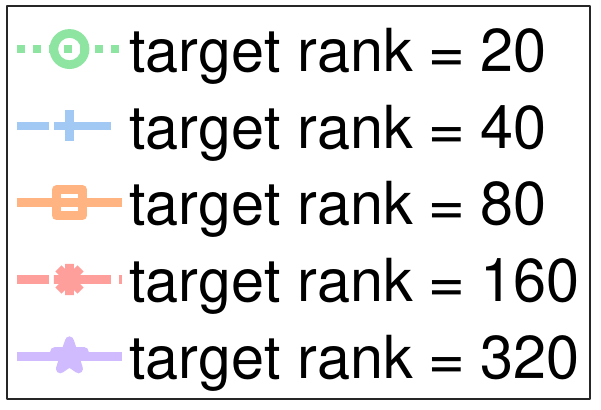}&\\
	   \includegraphics[width=0.3\textwidth]{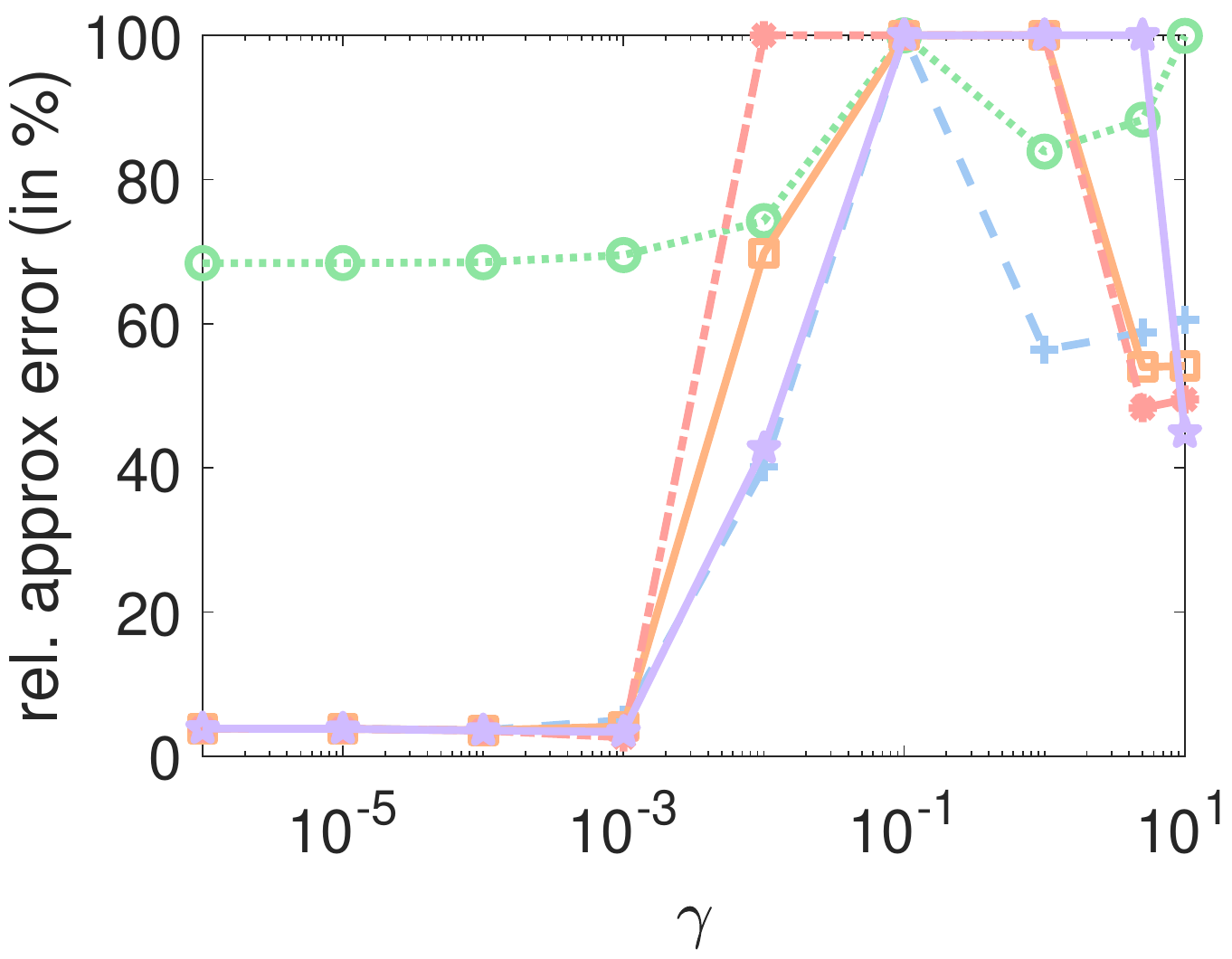}  &  \includegraphics[width=0.3\textwidth]{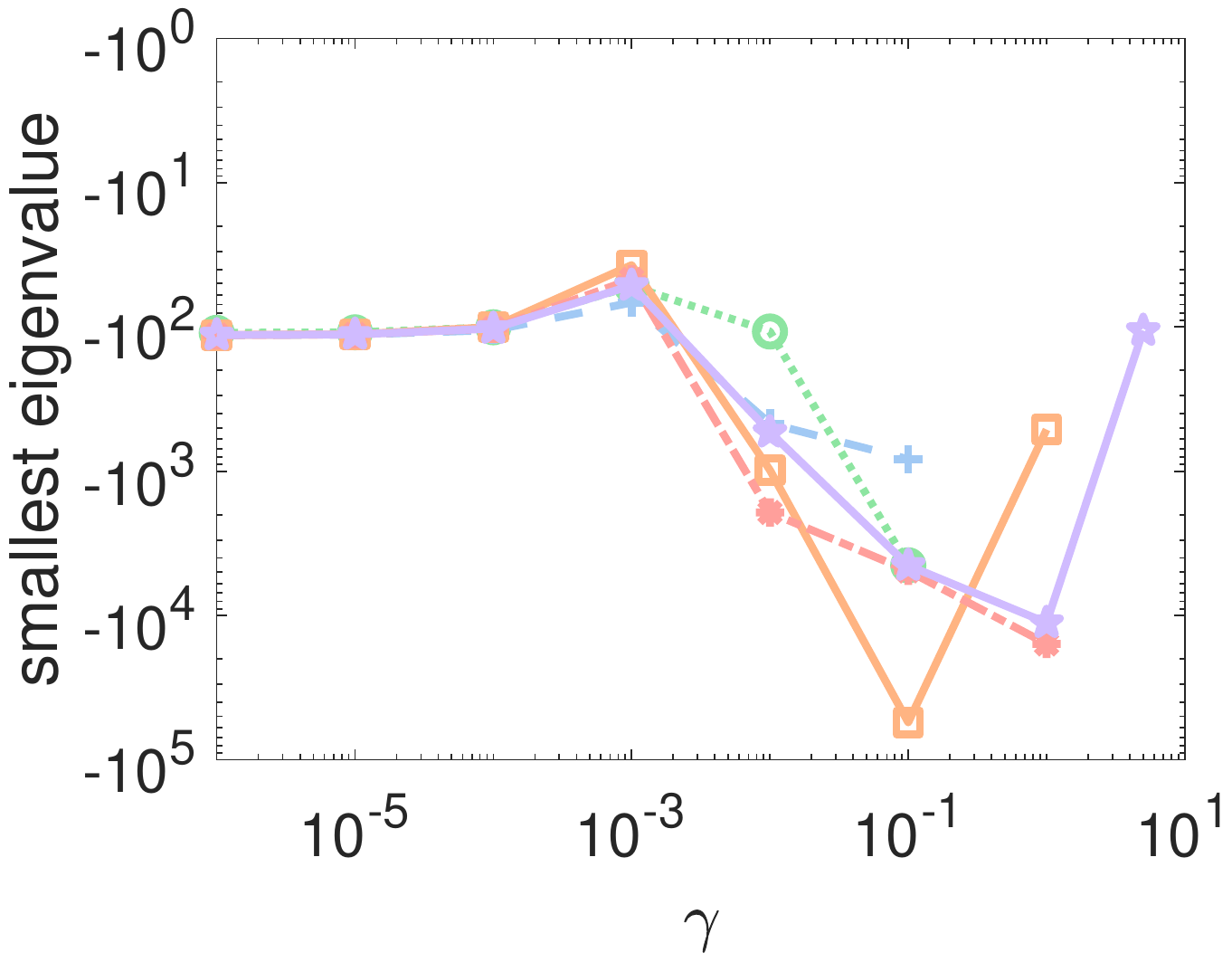}
	     & \includegraphics[width=0.3\textwidth]{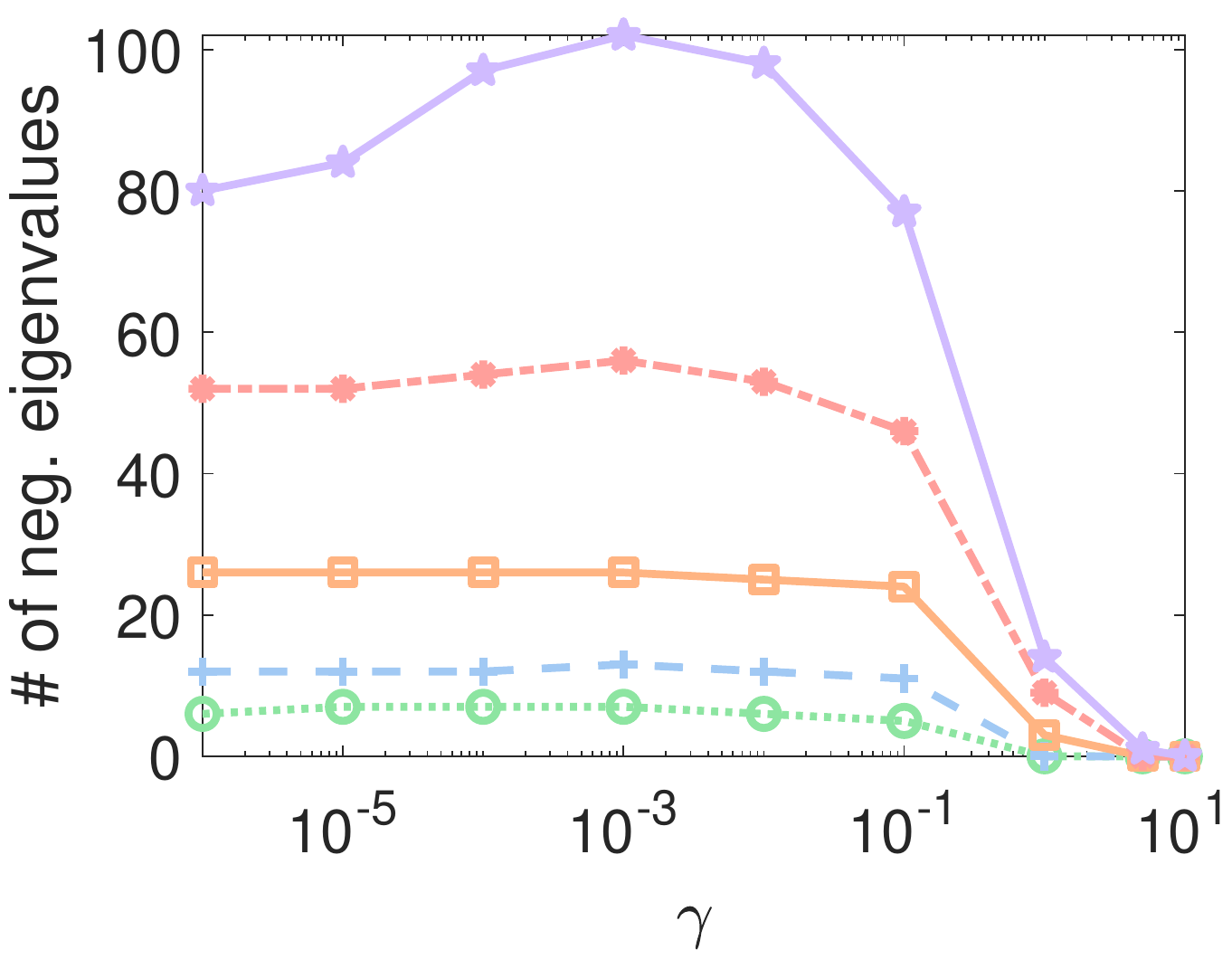}\\
	     \hspace{0.5cm}(a)&\hspace{0.7cm}(b)&\hspace{0.5cm}(c)\\
	\end{tabular}
	\caption{MEKA experiments with \textsf{spambase} data set and Gaussian rbf kernel.\label{pic:rbfResults1}}
	
\end{figure}

\begin{figure}[h!]
\centering
	\begin{tabular}{c c c}
	   \includegraphics[width=0.3\textwidth]{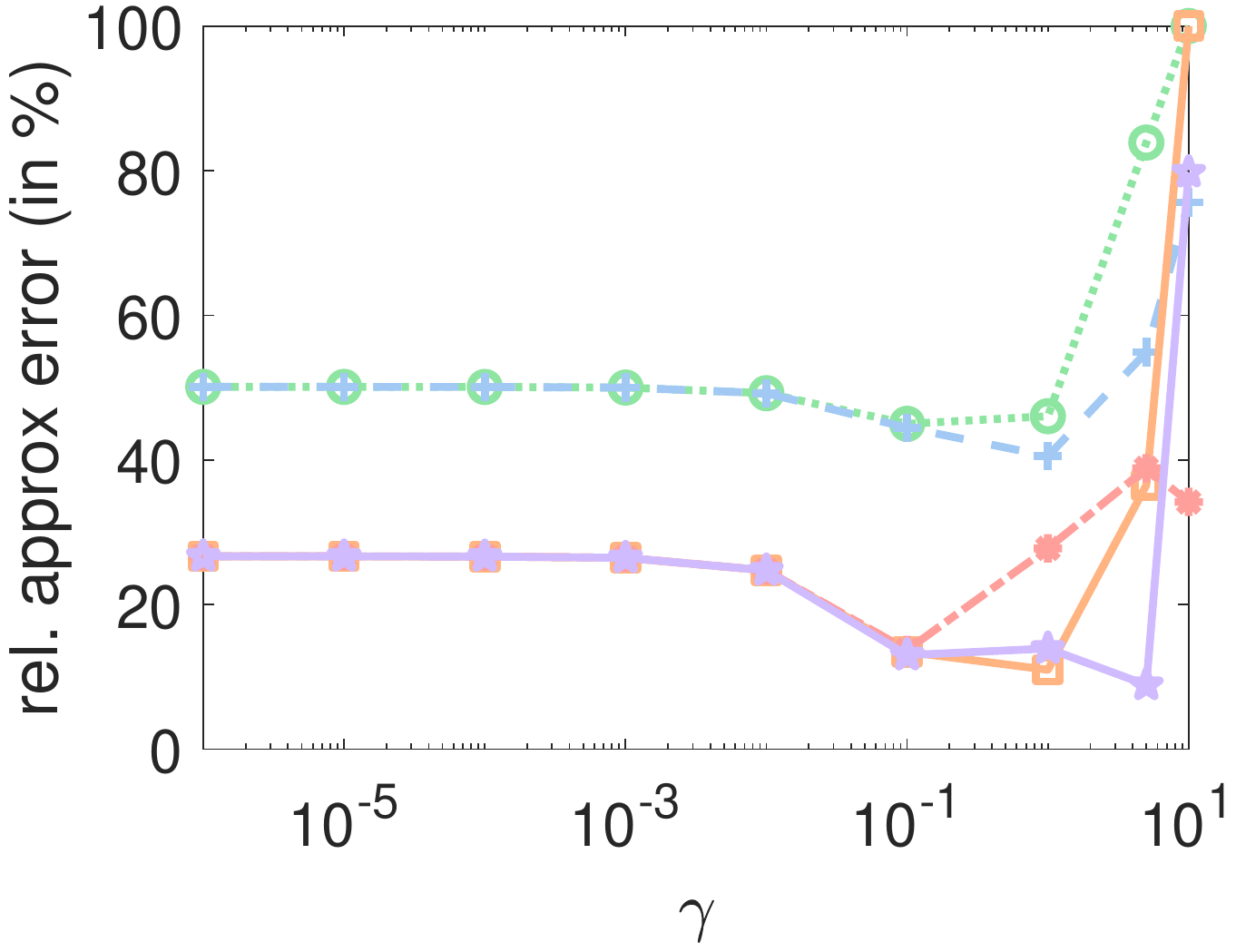}  &  \includegraphics[width=0.3\textwidth]{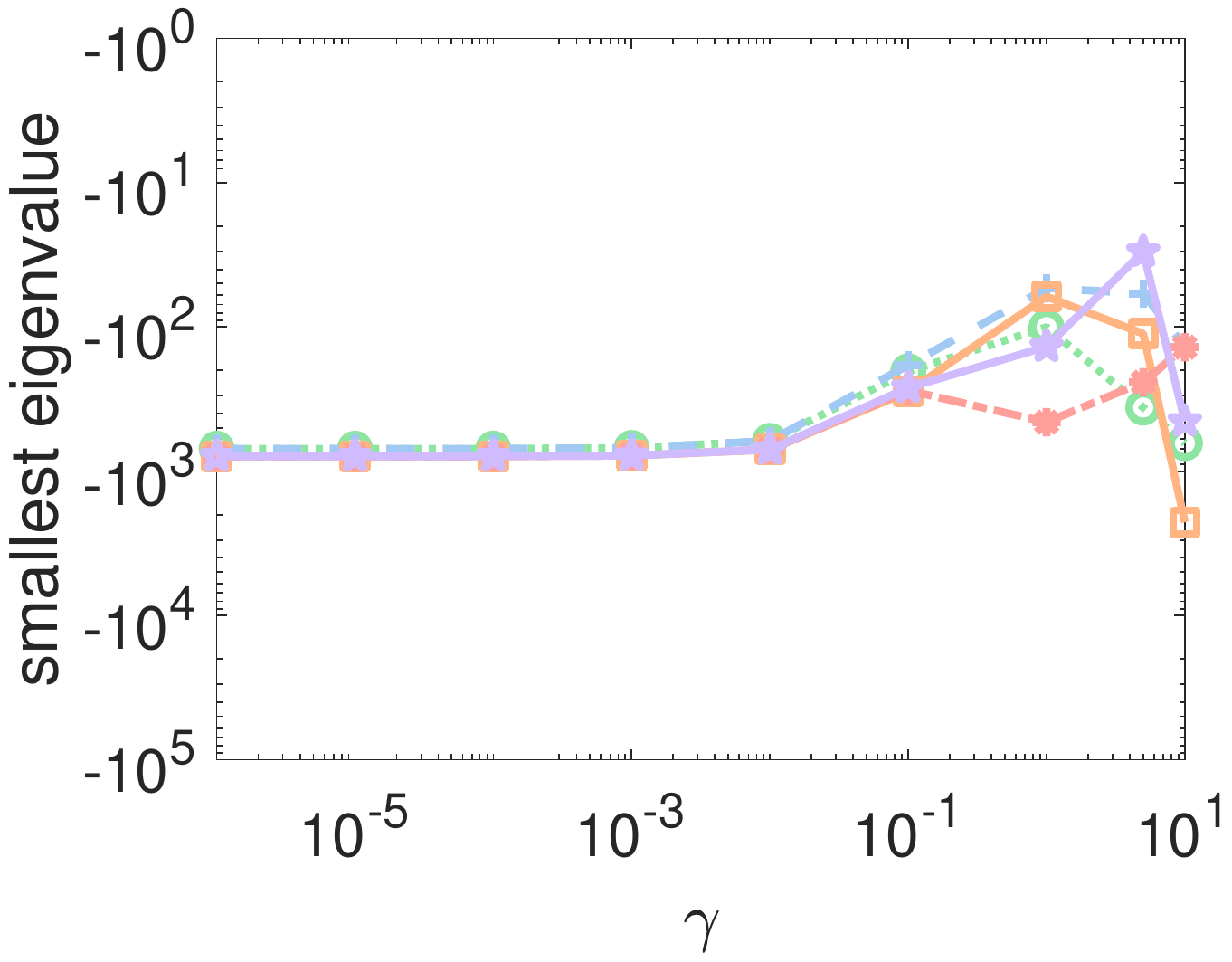}
	     & \includegraphics[width=0.3\textwidth]{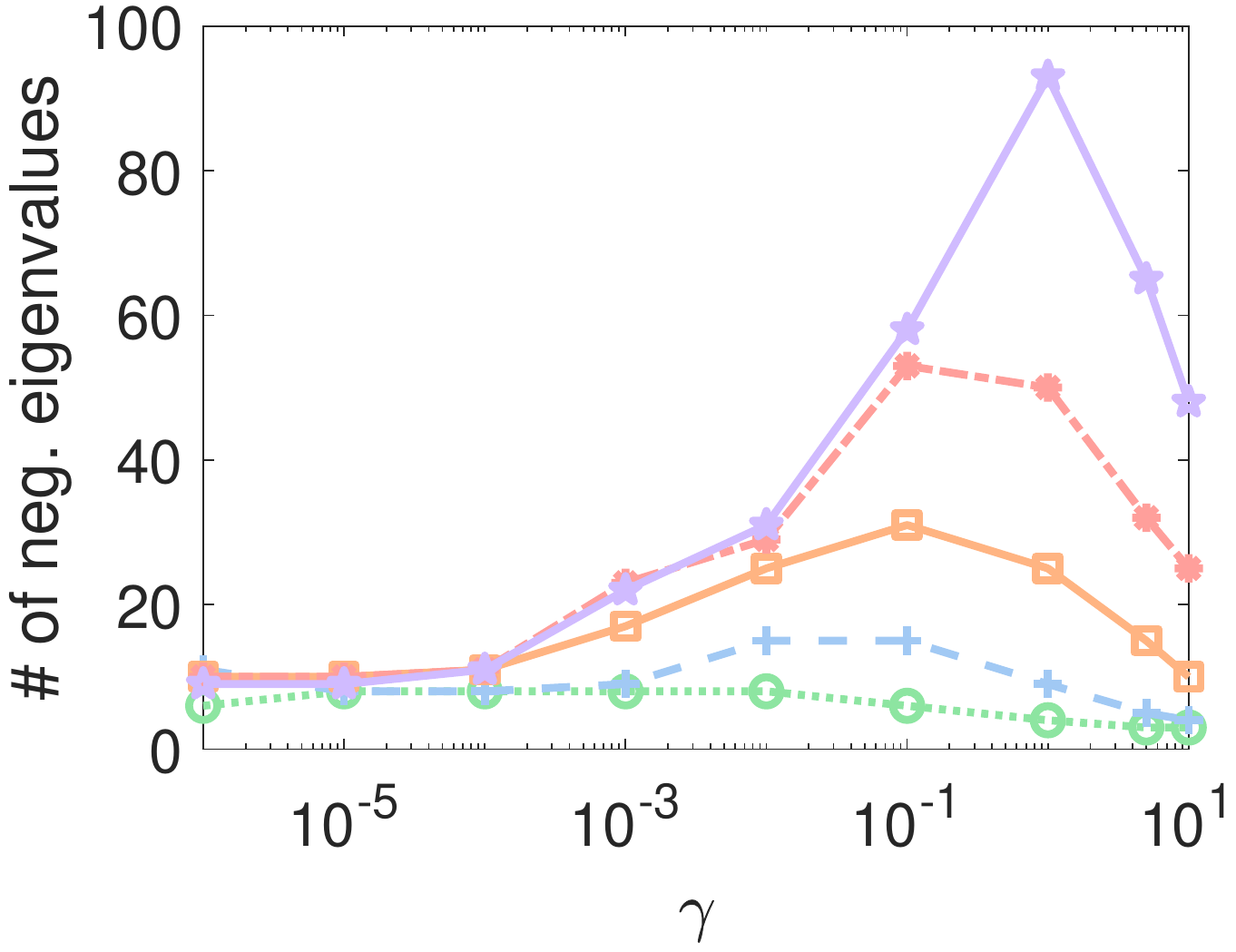}\\
	     \hspace{0.5cm}(a)&\hspace{0.7cm}(b)&\hspace{0.5cm}(c)\\
	\end{tabular}
	\caption{MEKA experiments with \textsf{artificial 1} data set and Gaussian rbf kernel. Same legend as in Figure \ref{pic:rbfResults1}.\label{pic:rbfResults2}}
	
\end{figure}

\begin{figure}[h!]
\centering
	\begin{tabular}{c c c}
	   \includegraphics[width=0.3\textwidth]{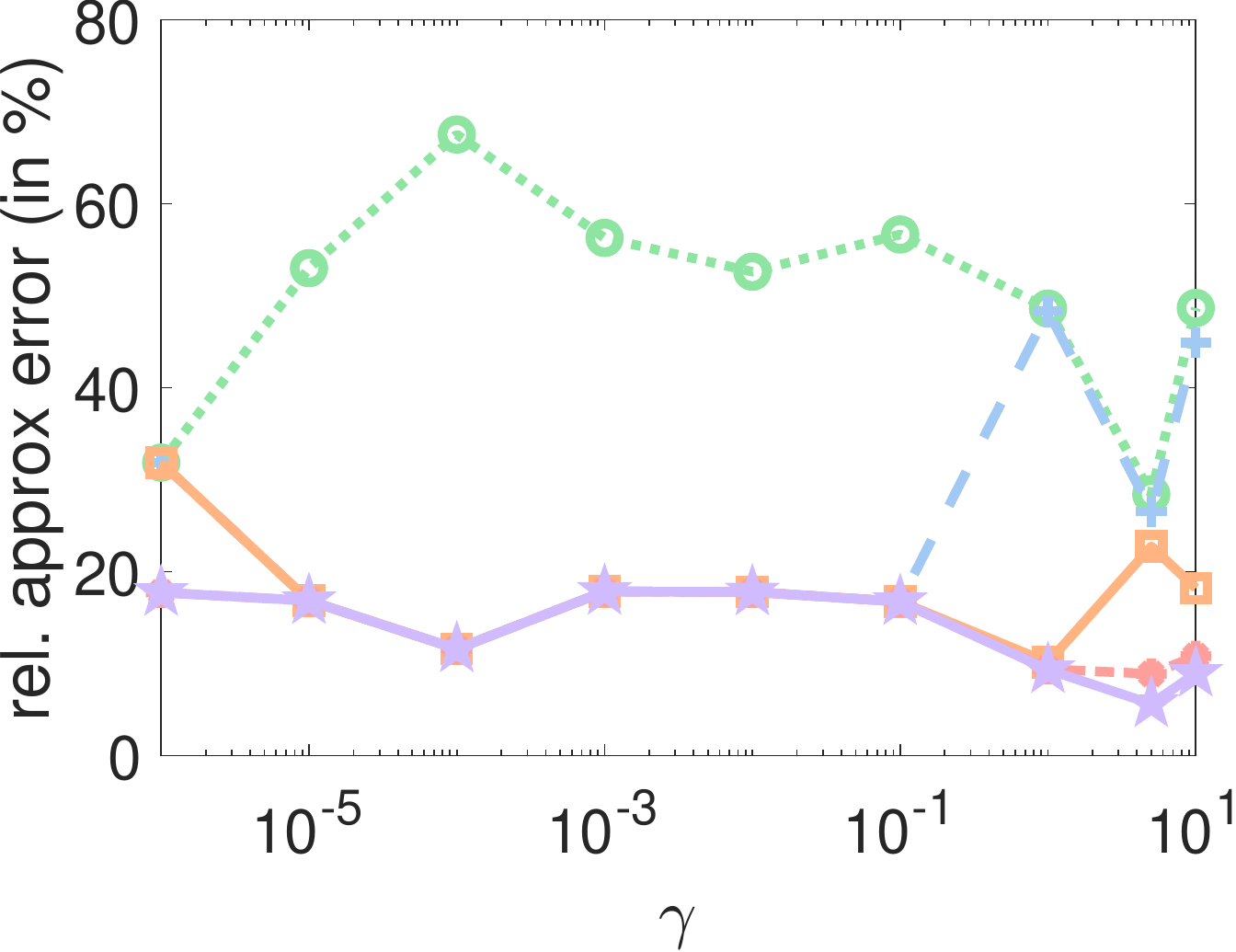}  &  \includegraphics[width=0.3\textwidth]{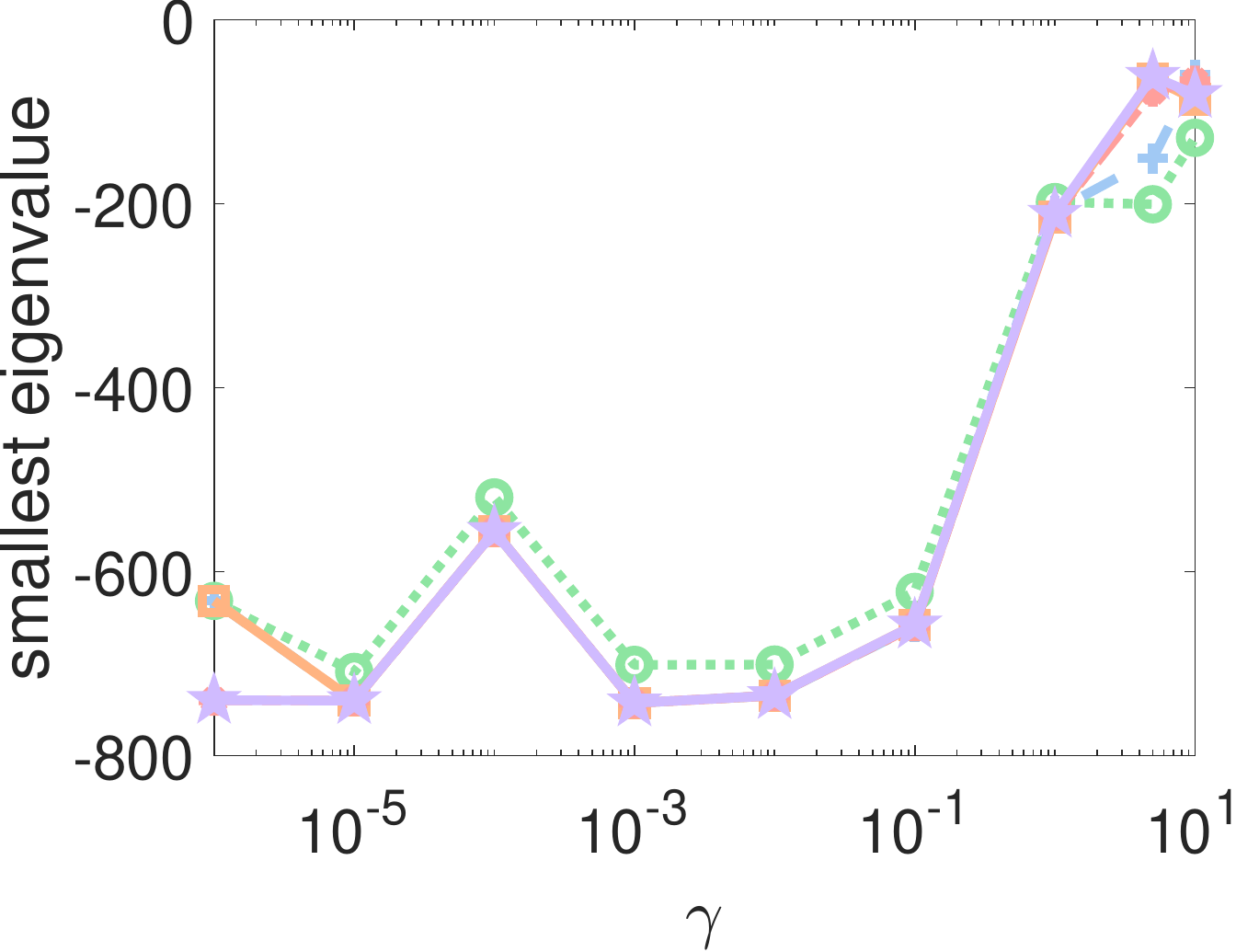}
	     & \includegraphics[width=0.3\textwidth]{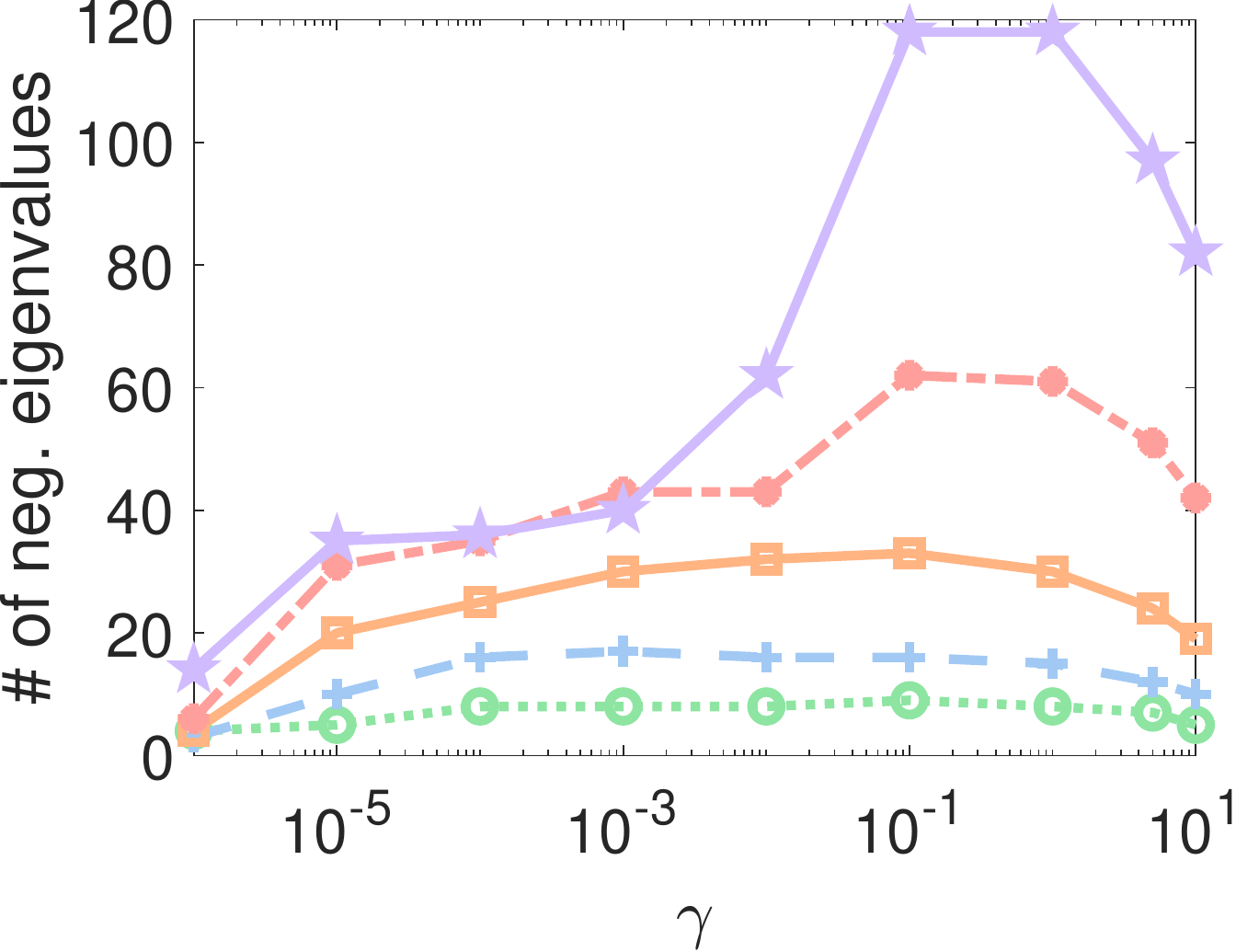}\\
	     \hspace{0.5cm}(a)&\hspace{0.7cm}(b)&\hspace{0.5cm}(c)\\
	\end{tabular}
	\caption{MEKA experiments with \textsf{cpusmall} data set and Gaussian rbf kernel. Same legend as in Figure \ref{pic:rbfResults1}.	\label{pic:rbfResults3}}

\end{figure}

\begin{figure}[h!]
\centering
	\begin{tabular}{c c c}
	   \includegraphics[width=0.3\textwidth]{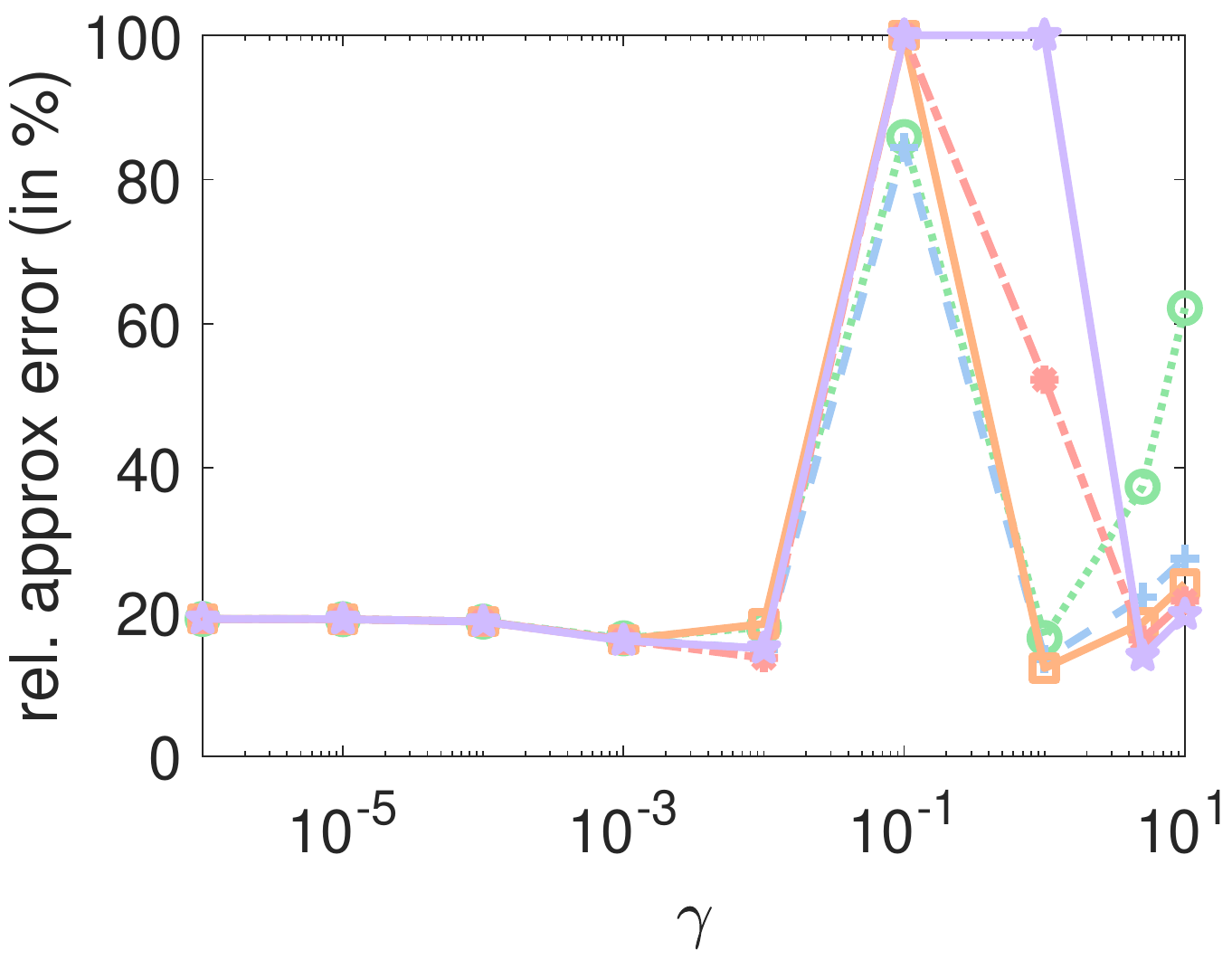}&  \includegraphics[width=0.3\textwidth]{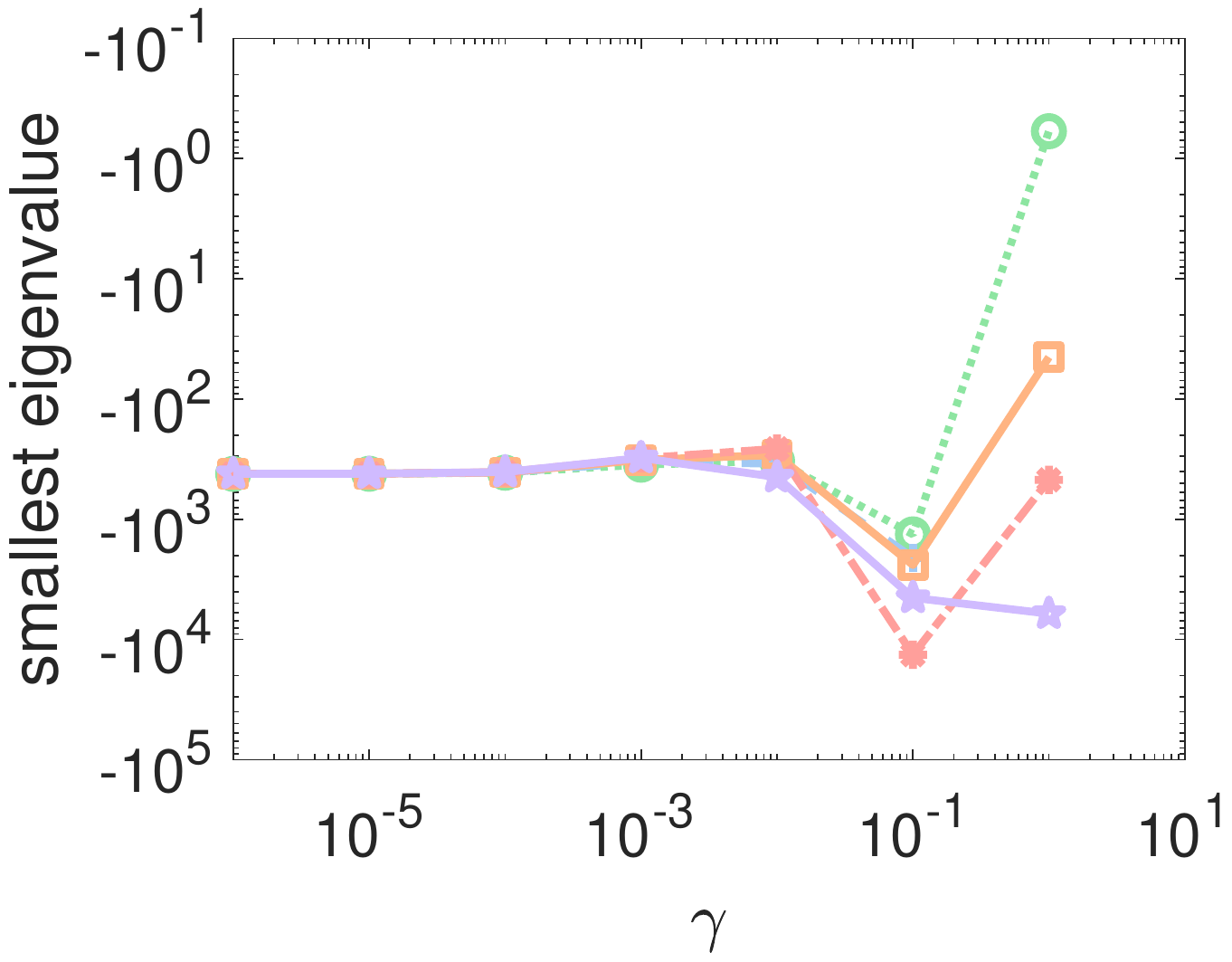}
	     & \includegraphics[width=0.3\textwidth]{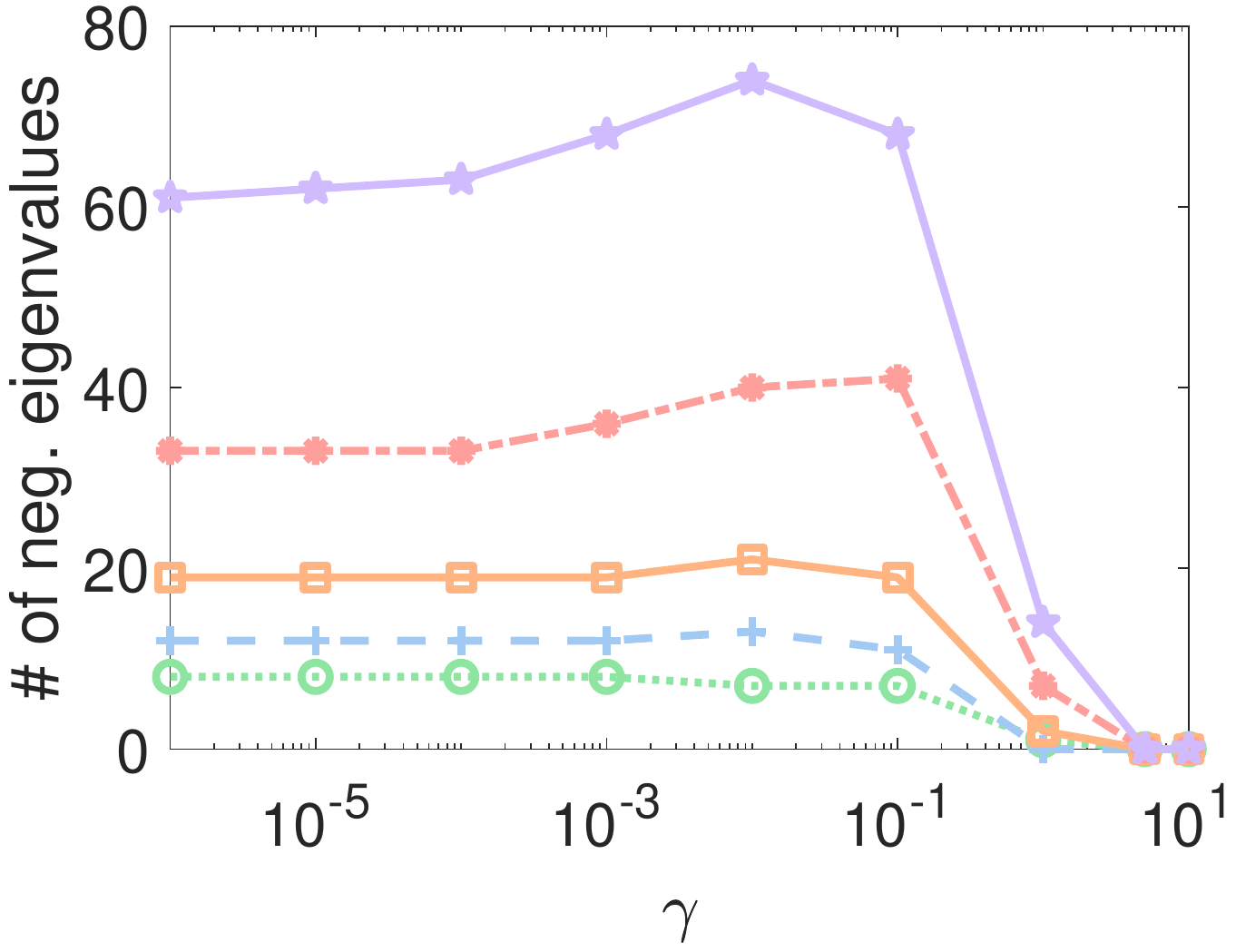}\\
	     \hspace{0.5cm}(a)&\hspace{0.7cm}(b)&\hspace{0.5cm}(c)\\
	\end{tabular}
	\caption{MEKA experiments with \textsf{gesture} data set and Gaussian rbf kernel. Same legend as in Figure \ref{pic:rbfResults1}.\label{pic:rbfResults4}}
	
\end{figure}

\begin{figure}[h!]
\centering
	\begin{tabular}{c c c}
	   \includegraphics[width=0.3\textwidth]{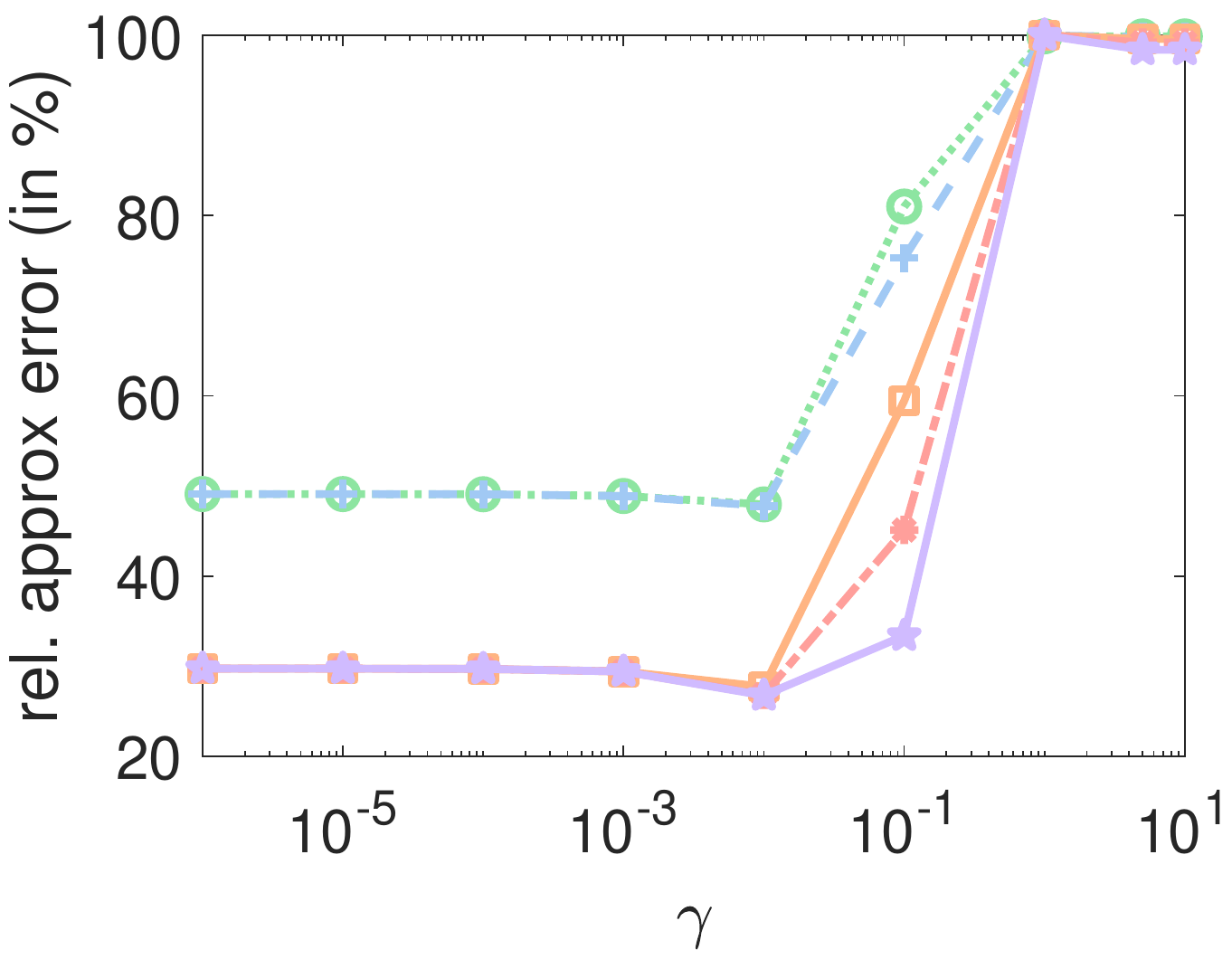}  &  \includegraphics[width=0.3\textwidth]{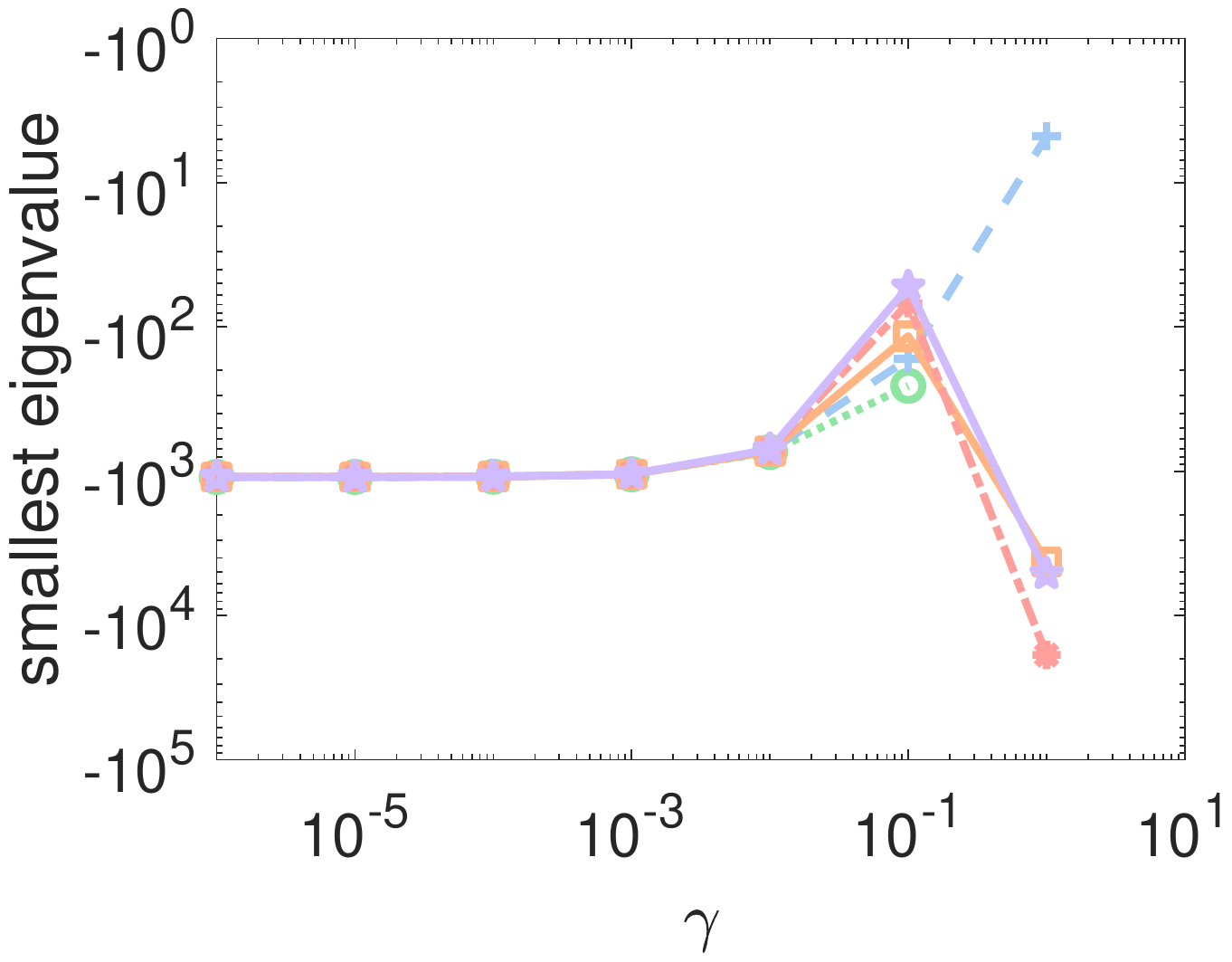}
	     & \includegraphics[width=0.3\textwidth]{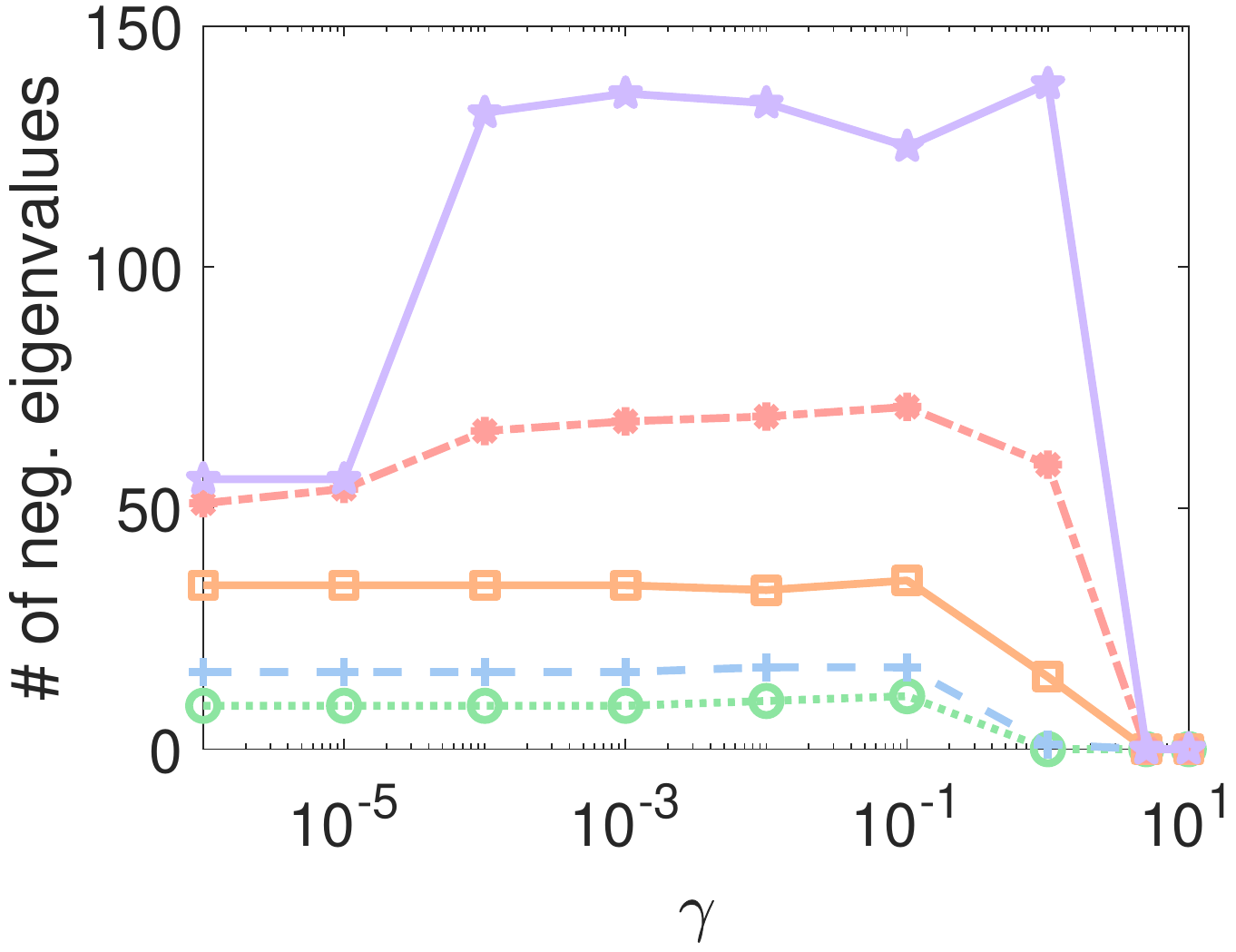}\\
	     \hspace{0.5cm}(a)&\hspace{0.7cm}(b)&\hspace{0.5cm}(c)\\
	\end{tabular}
	\caption{MEKA experiments with \textsf{artificial 2} data set and Gaussian rbf kernel. Same legend as in Figure \ref{pic:rbfResults1}.\label{pic:rbfResults5}}
	
\end{figure}

In the paper we studied the occurrence of invalid kernel approximations and in the following the remaining results are presented. Figures \ref{pic:rbfResults1}-\ref{pic:rbfResults5} illustrate the remaining Gaussian rbf kernel results and Figures \ref{pic:elmResults1}-\ref{pic:elmResults5} focus on the extreme learning kernel. (a) shows in the following figures the relative approximation error along the different target ranks and numbers of clusters, (b) presents the value of the smallest eigenvalue and (c) shows the total number of negative eigenvalues.

It is apparent that the evaluations show similar results, by means that MEKA achieves lower approximation error with higher rank and produces among all constellations a non-negligible amount of negative eigenvalues. It is also noteworthy that a higher target rank leads to a higher amount of negative eigenvalues, as it can be consistently seen in Figures \ref{pic:rbfResults1}-\ref{pic:rbfResults5} (c).

\begin{figure}[h!]
\centering
	\begin{tabular}{c c c}
		&\hspace{0.7cm}\includegraphics[width=0.1\textwidth]{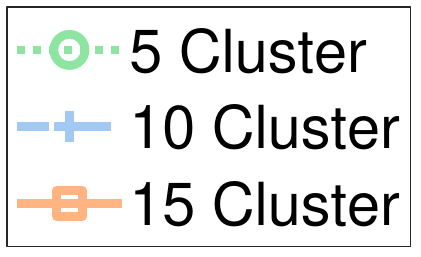}&\\
	   \includegraphics[width=0.3\textwidth]{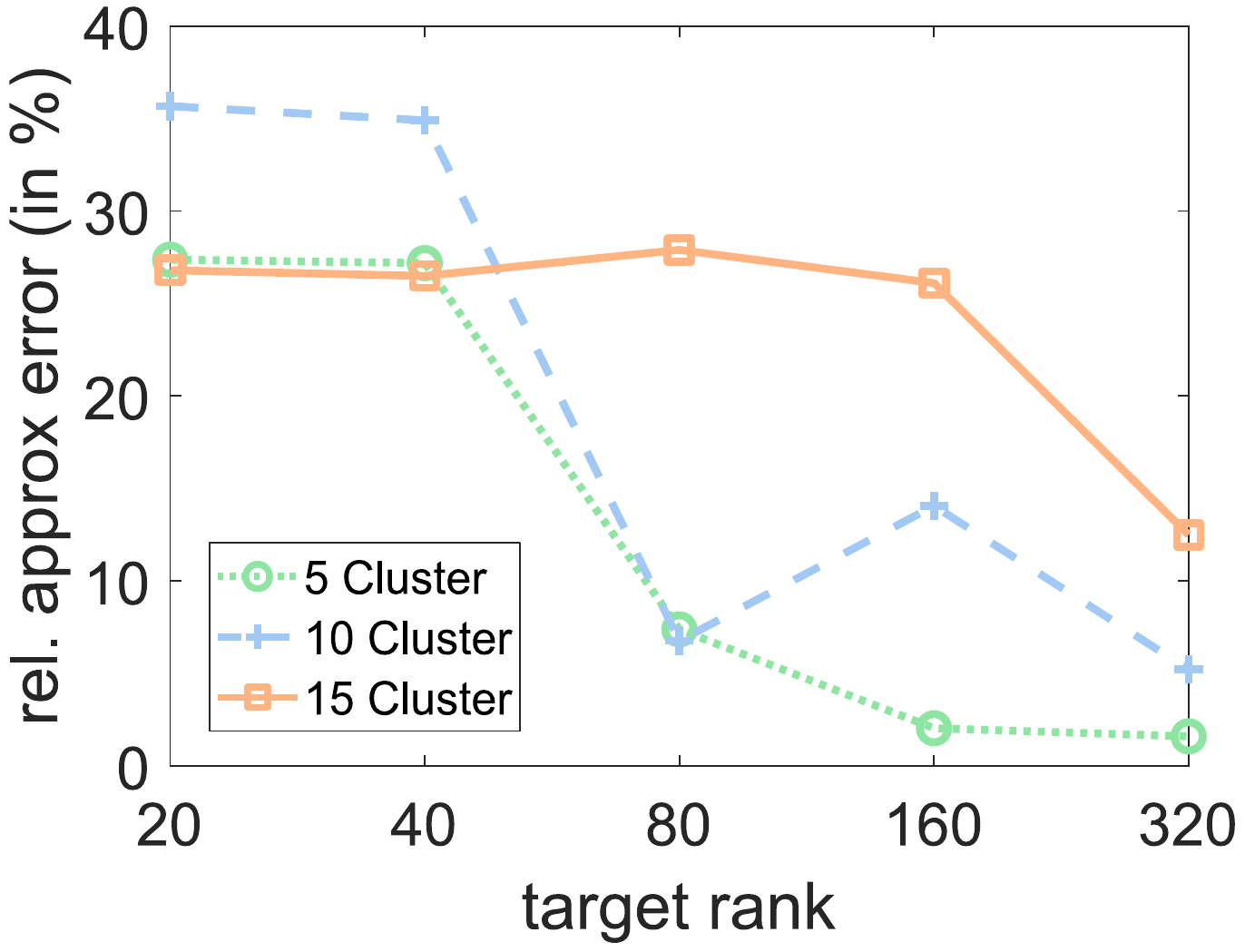}  &  \includegraphics[width=0.3\textwidth]{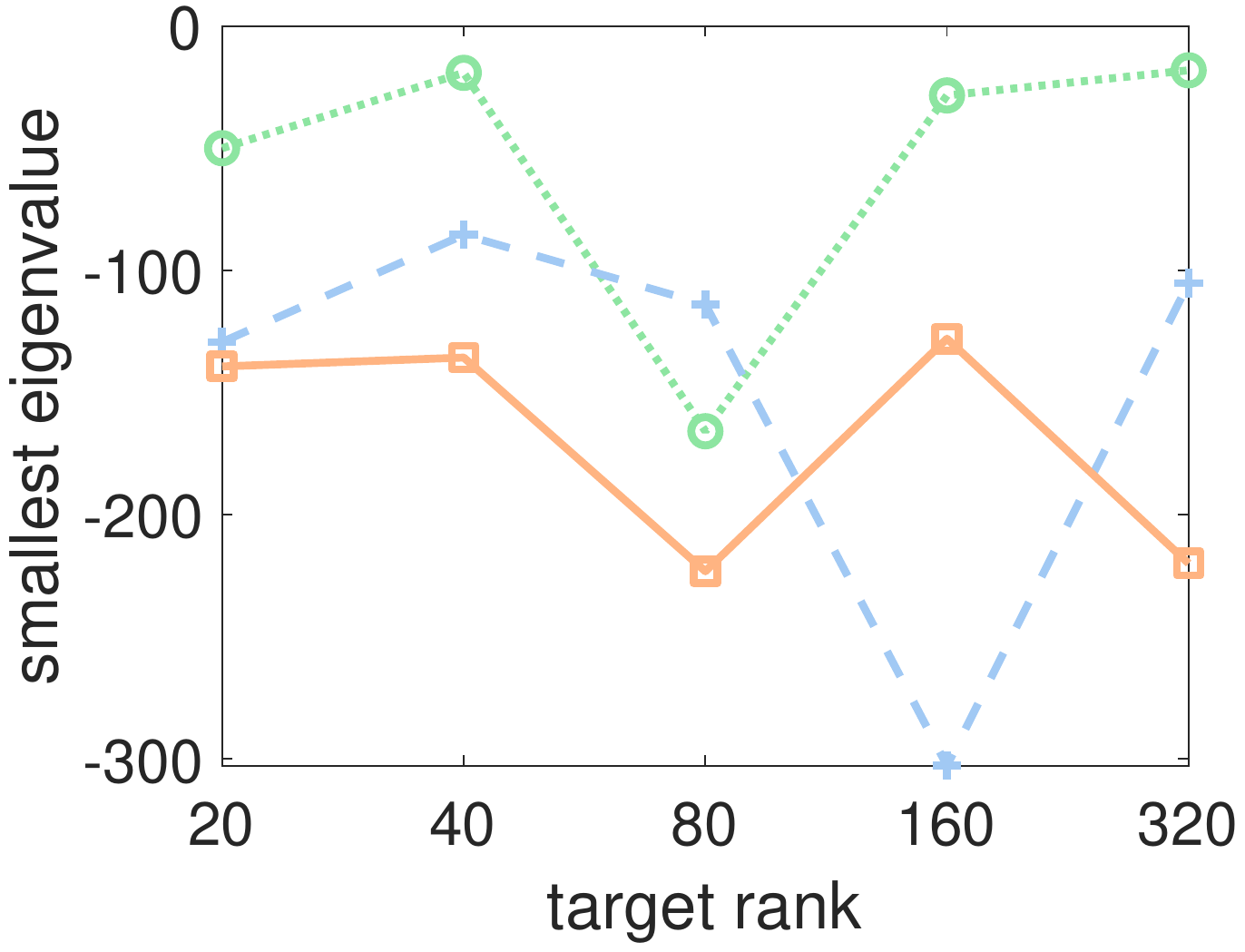}
	     & \includegraphics[width=0.3\textwidth]{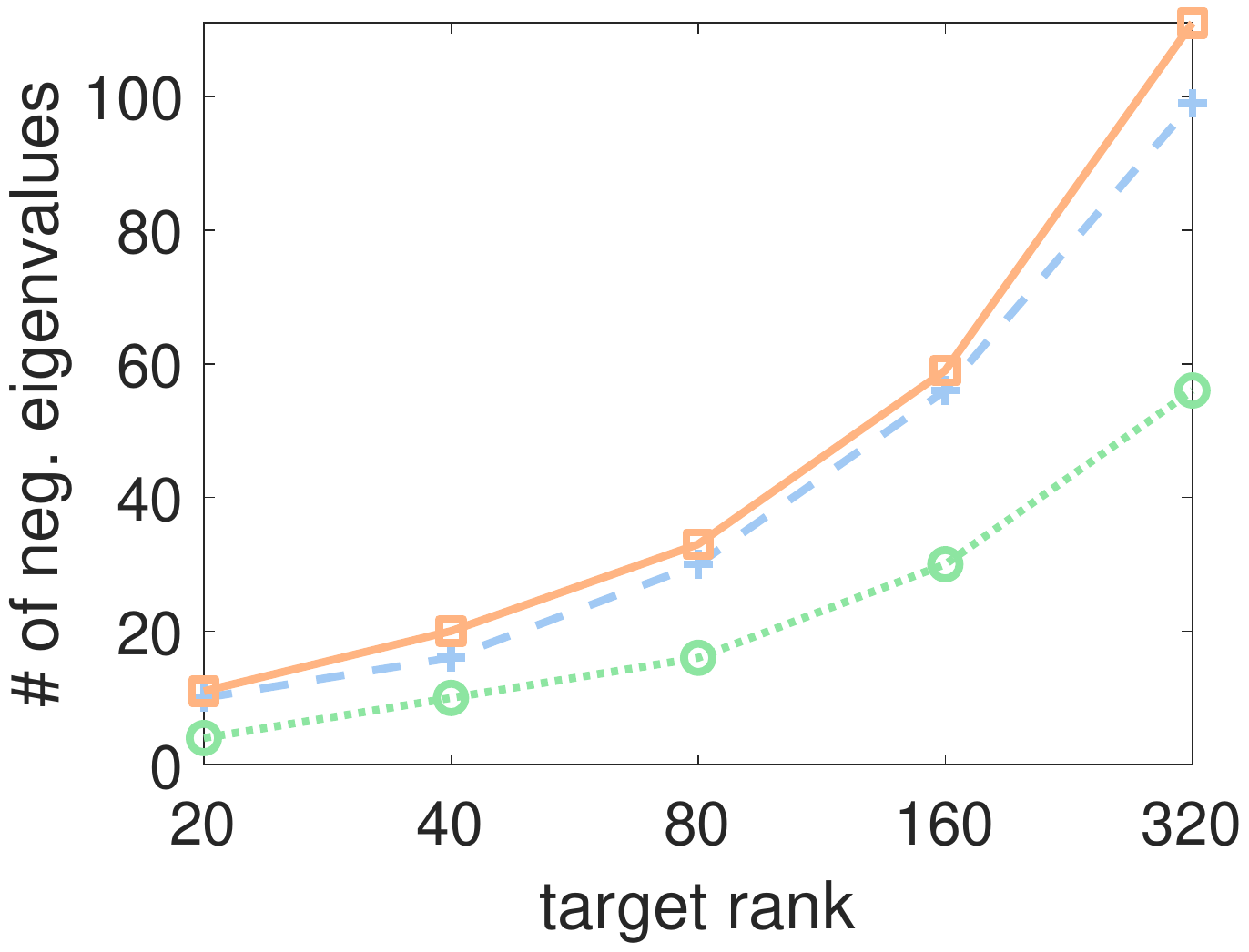}\\
	     \hspace{0.5cm}(a)&\hspace{0.7cm}(b)&\hspace{0.5cm}(c)\\
	\end{tabular}
	\caption{MEKA experiments with \textsf{spambase} data set and extreme learning kernel. \label{pic:elmResults1}}
	
\end{figure}

\begin{figure}[h!]
\centering
	\begin{tabular}{c c c}
	   \includegraphics[width=0.3\textwidth]{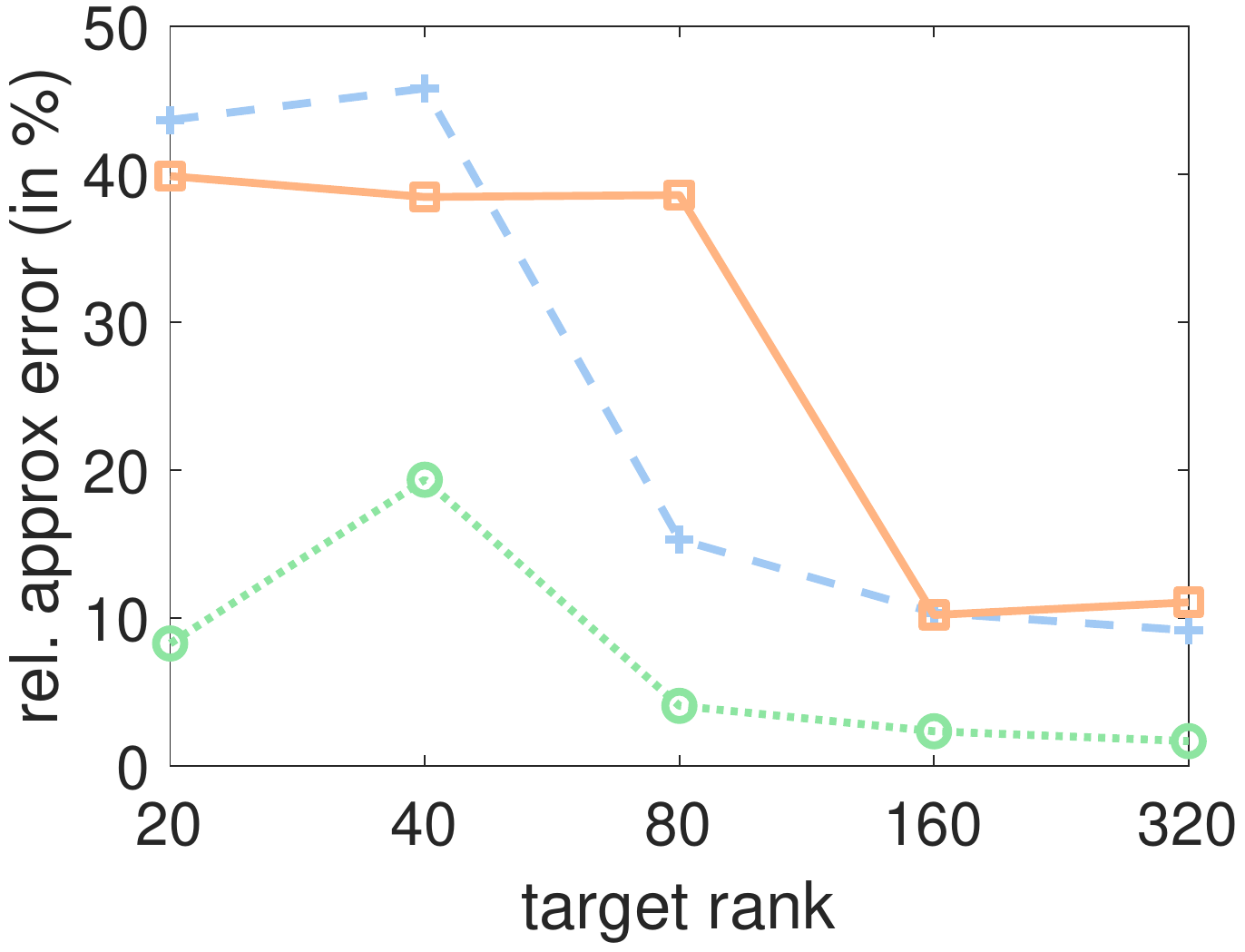}  &  \includegraphics[width=0.3\textwidth]{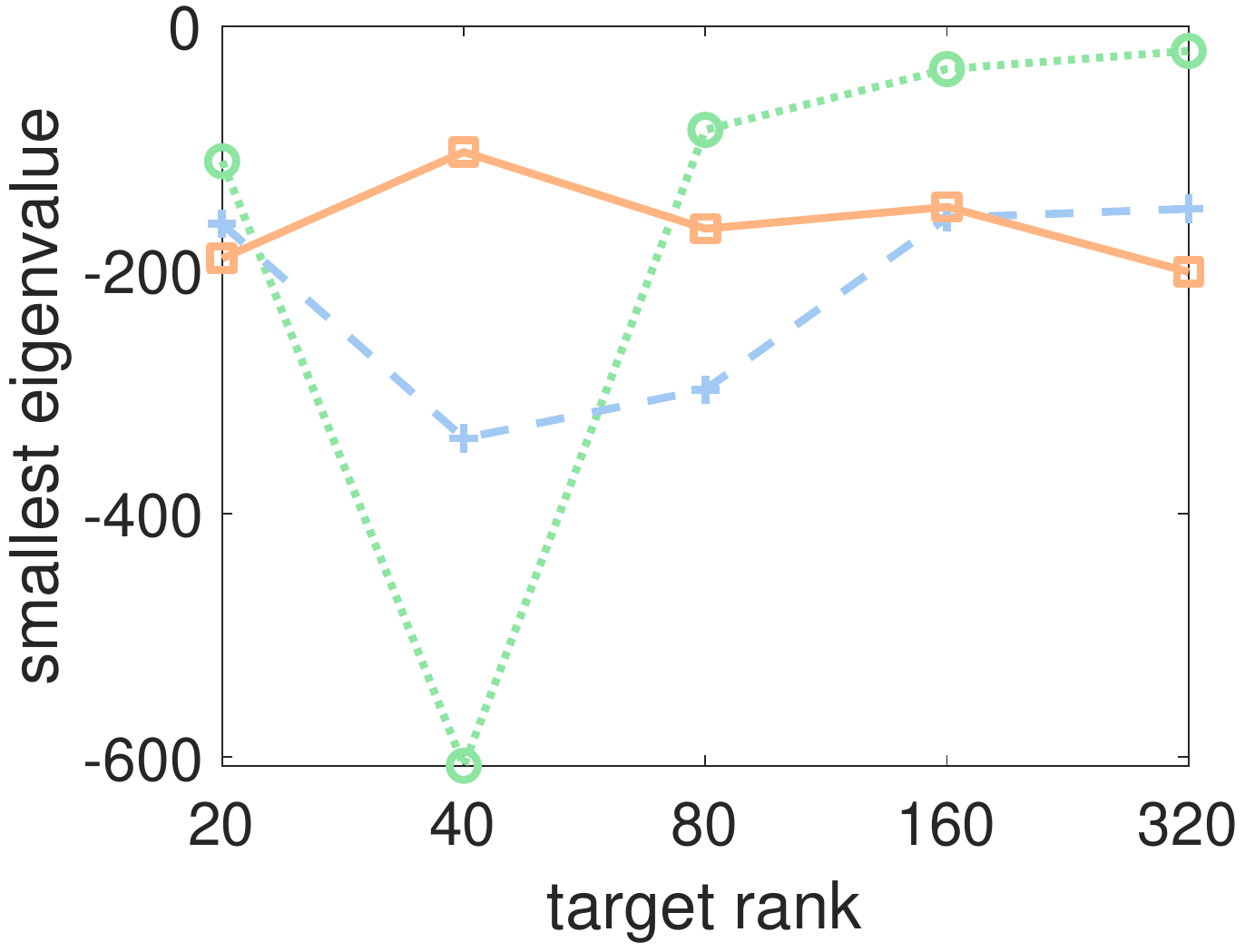}
	     & \includegraphics[width=0.3\textwidth]{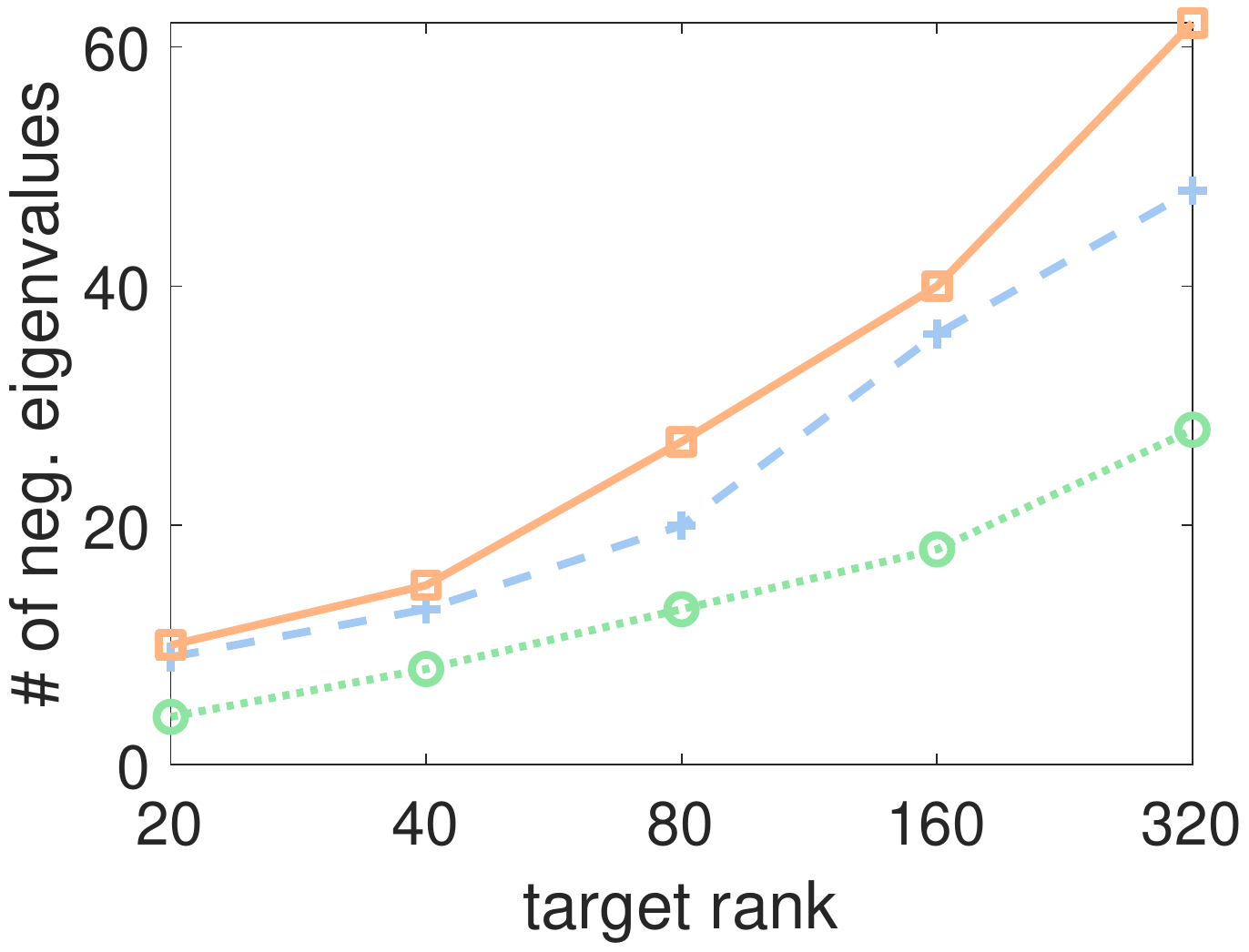}\\
	     \hspace{0.5cm}(a)&\hspace{0.7cm}(b)&\hspace{0.5cm}(c)\\
	\end{tabular}
	\caption{MEKA experiments with \textsf{artificial 1} data set and extreme learning kernel. Same legend as in Figure \ref{pic:elmResults1}. \label{pic:elmResults2}}
	
\end{figure}

\begin{figure}[h!]
\centering
	\begin{tabular}{c c c}
	   \includegraphics[width=0.3\textwidth]{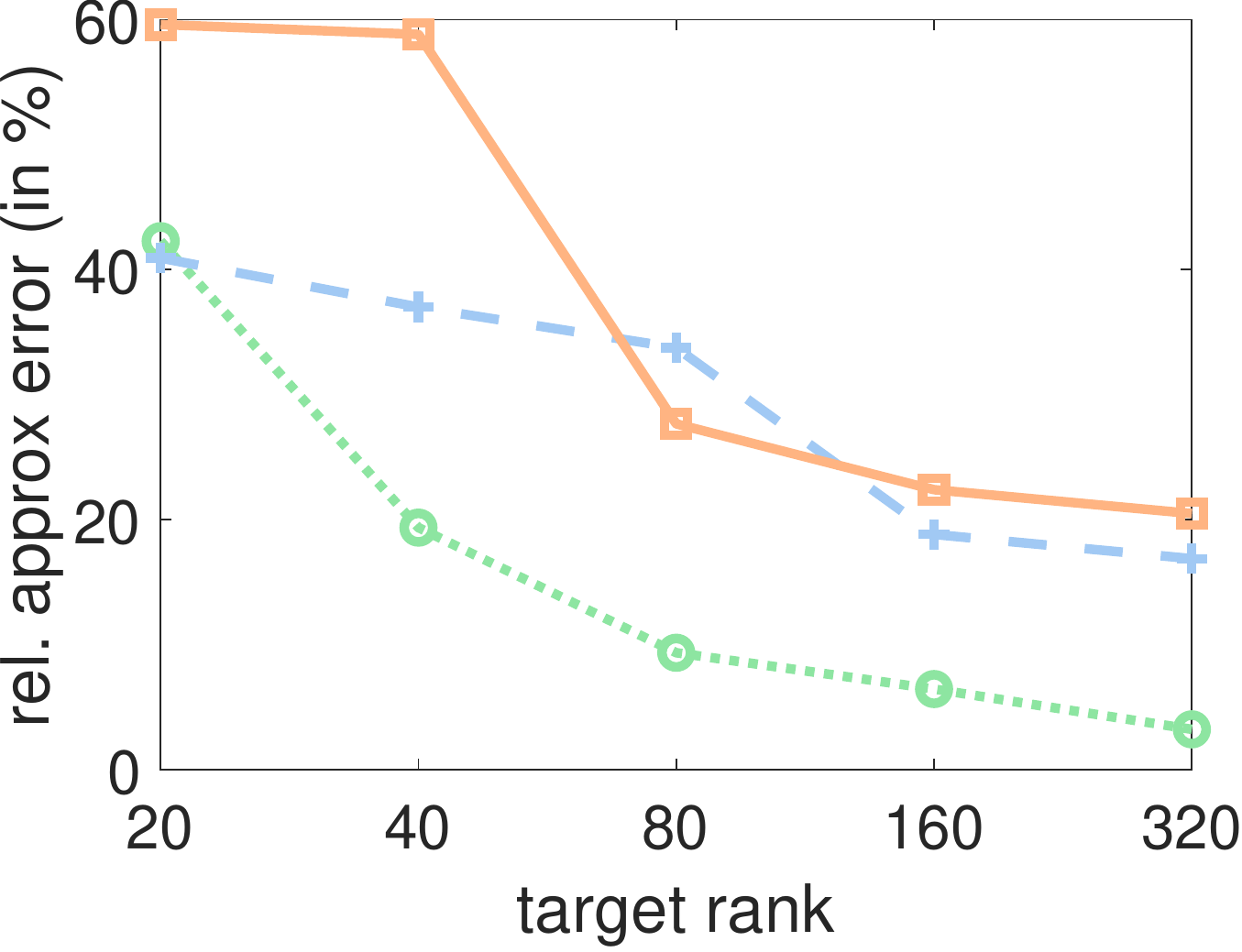}  &  \includegraphics[width=0.3\textwidth]{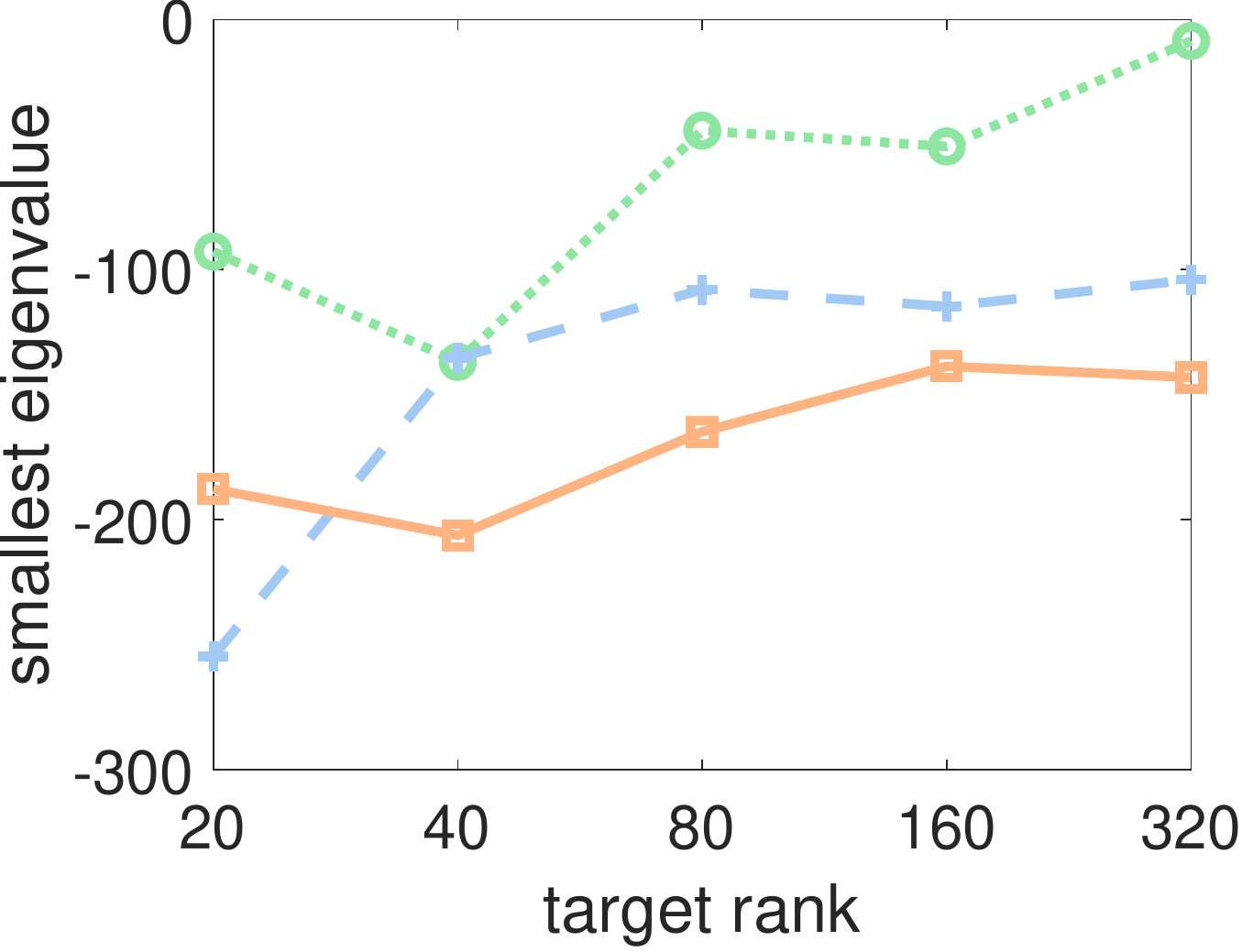}
	     & \includegraphics[width=0.3\textwidth]{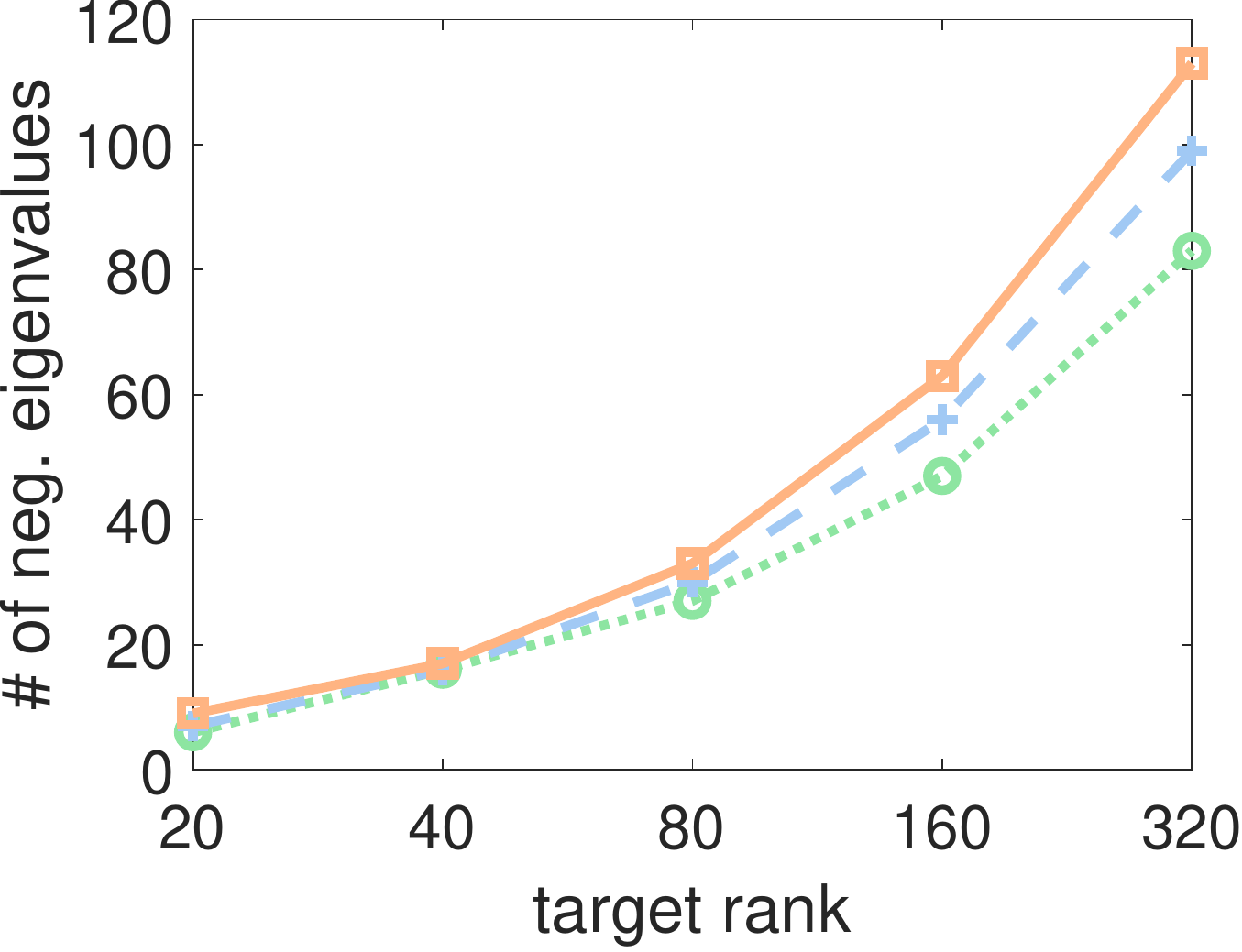}\\
	     \hspace{0.5cm}(a)&\hspace{0.7cm}(b)&\hspace{0.5cm}(c)\\
	\end{tabular}
	\caption{MEKA experiments with \textsf{cpusmall} data set and extreme learning kernel. Same legend as in Figure \ref{pic:elmResults1}.	\label{pic:elmResults3}}
	
\end{figure}

\begin{figure}[h!]
\centering
	\begin{tabular}{c c c}
	   \includegraphics[width=0.3\textwidth]{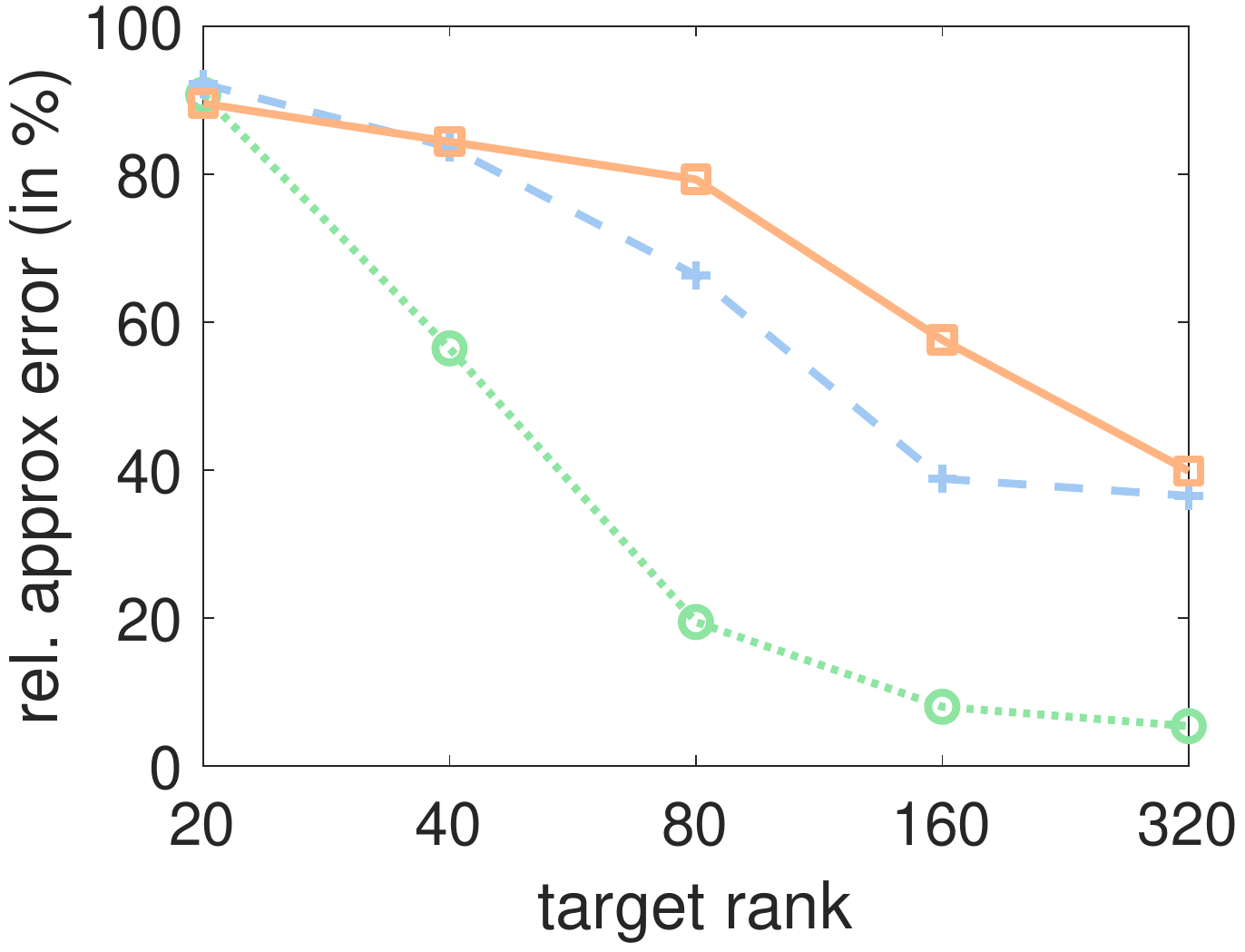}  &  \includegraphics[width=0.3\textwidth]{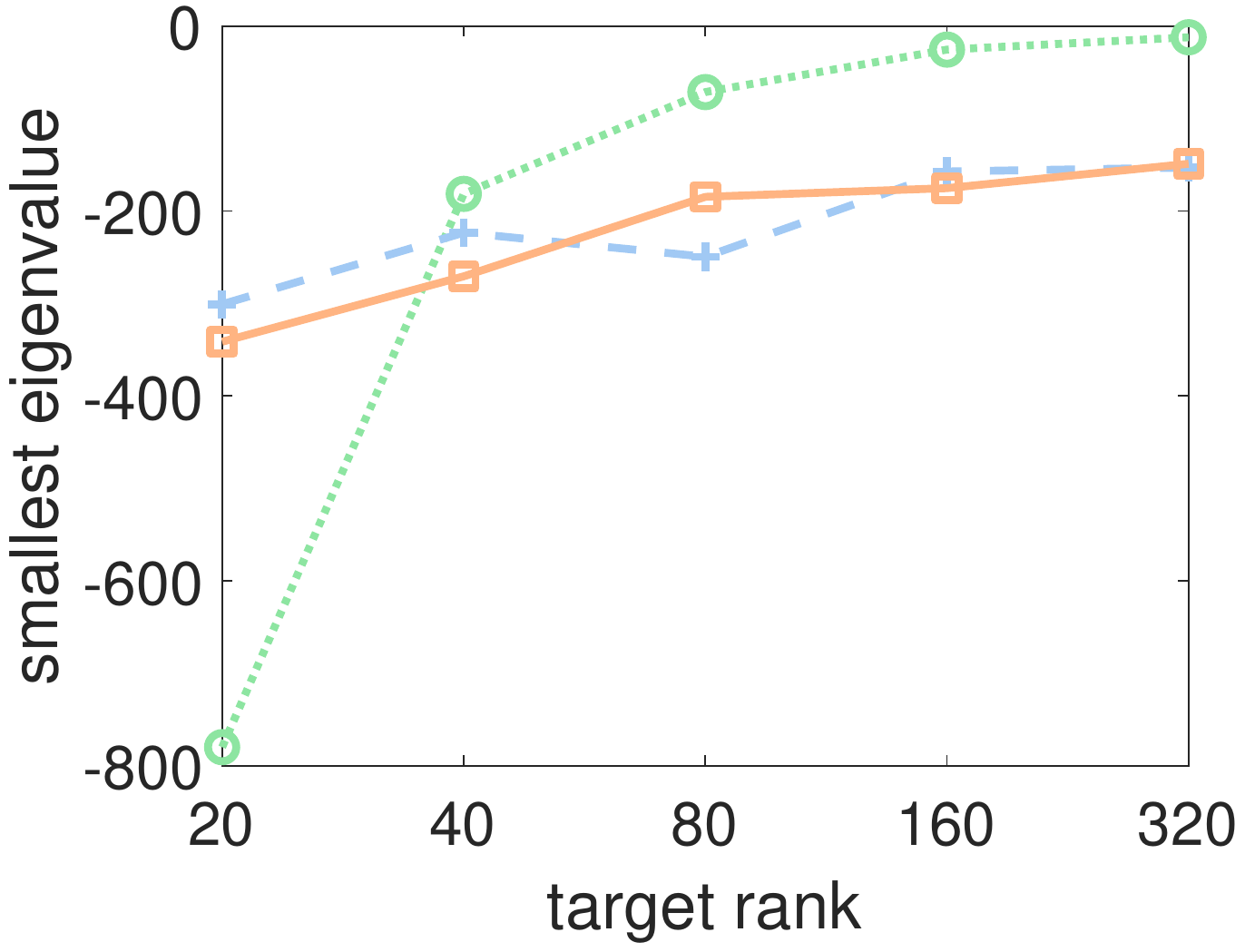}
	     & \includegraphics[width=0.3\textwidth]{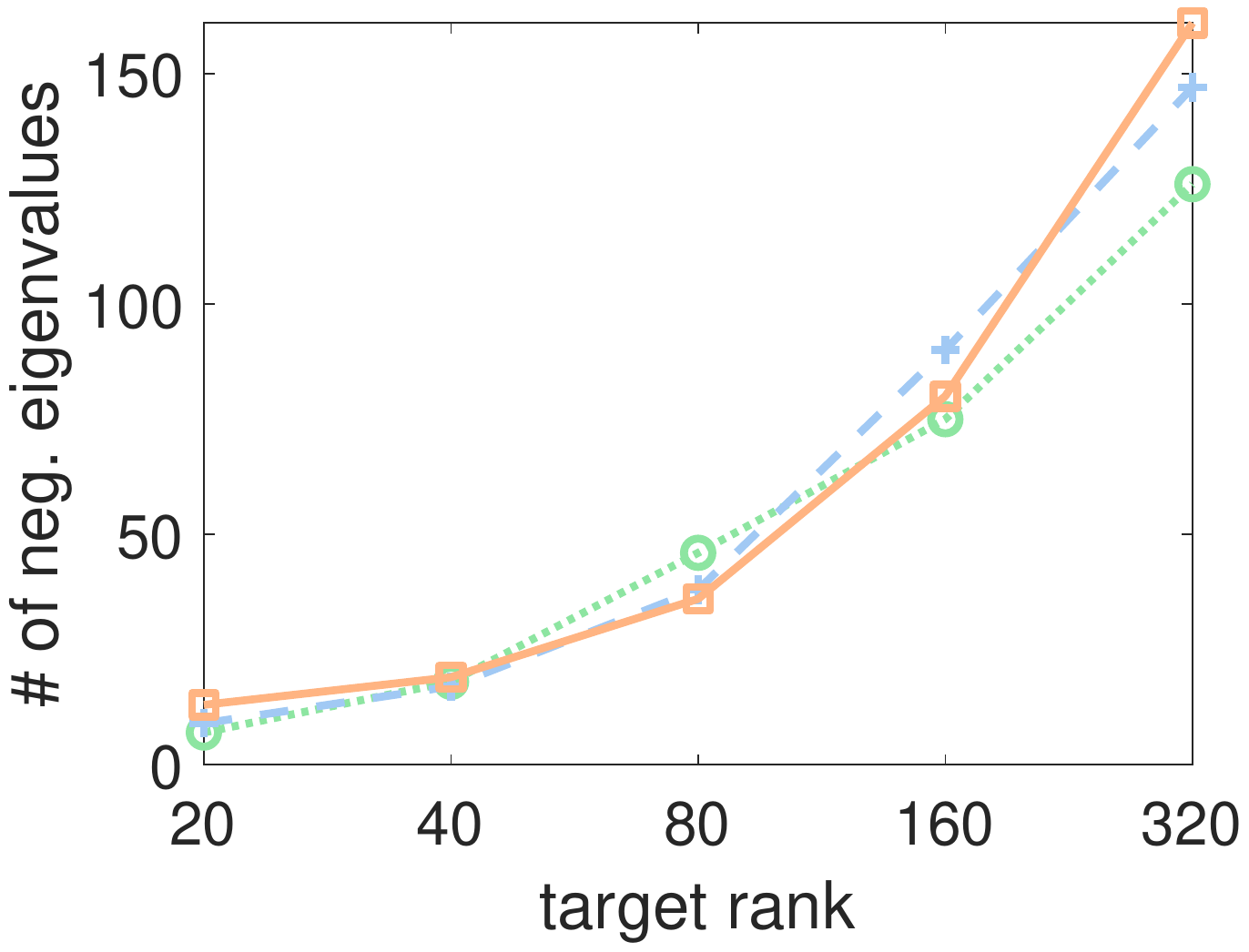}\\
	     \hspace{0.5cm}(a)&\hspace{0.7cm}(b)&\hspace{0.5cm}(c)\\
	\end{tabular}
	\caption{MEKA experiments with \textsf{artificial 2} data set and extreme learning kernel. Same legend as in Figure \ref{pic:elmResults1}.\label{pic:elmResults4}}
	
\end{figure}
\FloatBarrier
\begin{figure}[t!]
\centering
	\begin{tabular}{c c c}
	 \includegraphics[width=0.3\textwidth]{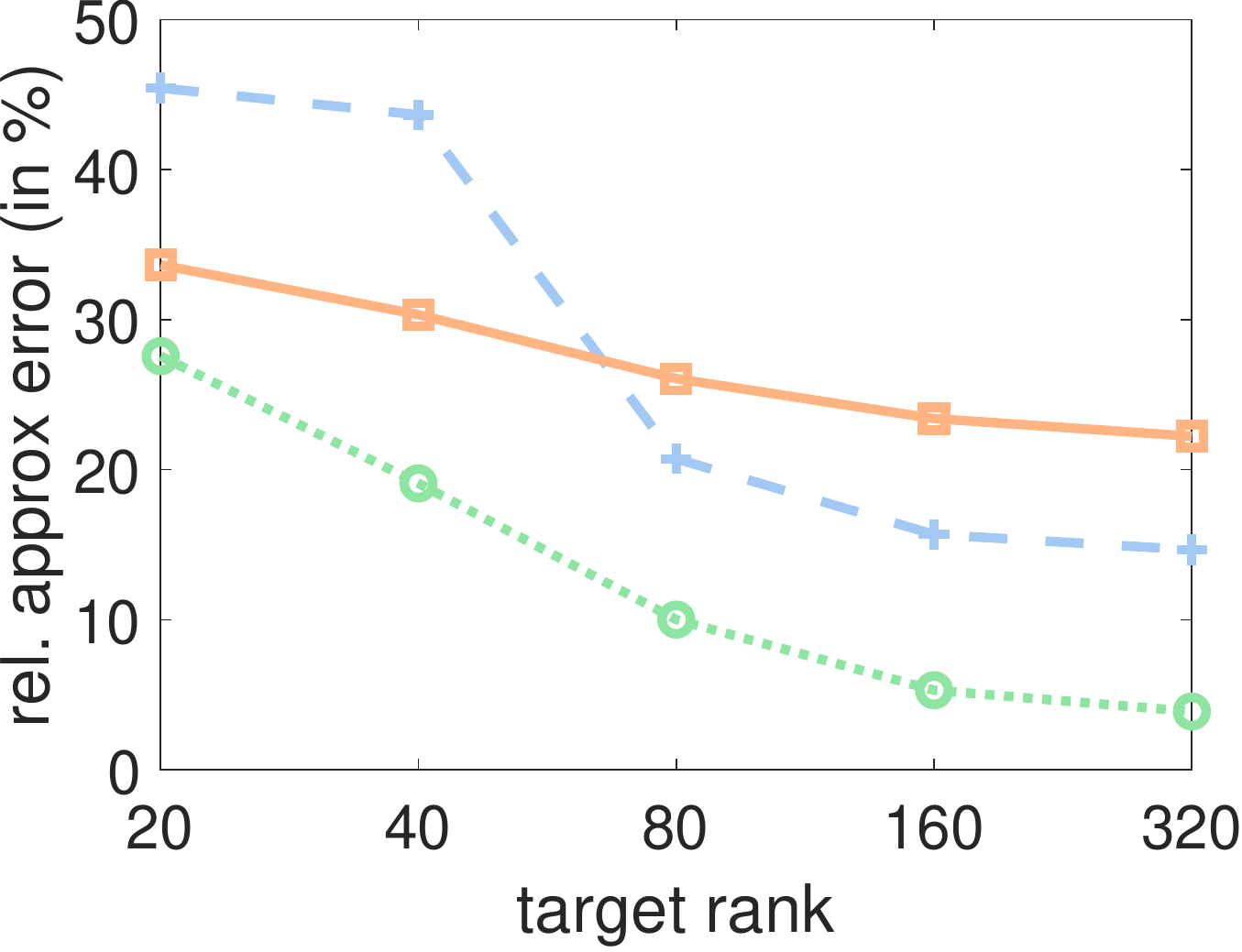}  &  \includegraphics[width=0.3\textwidth]{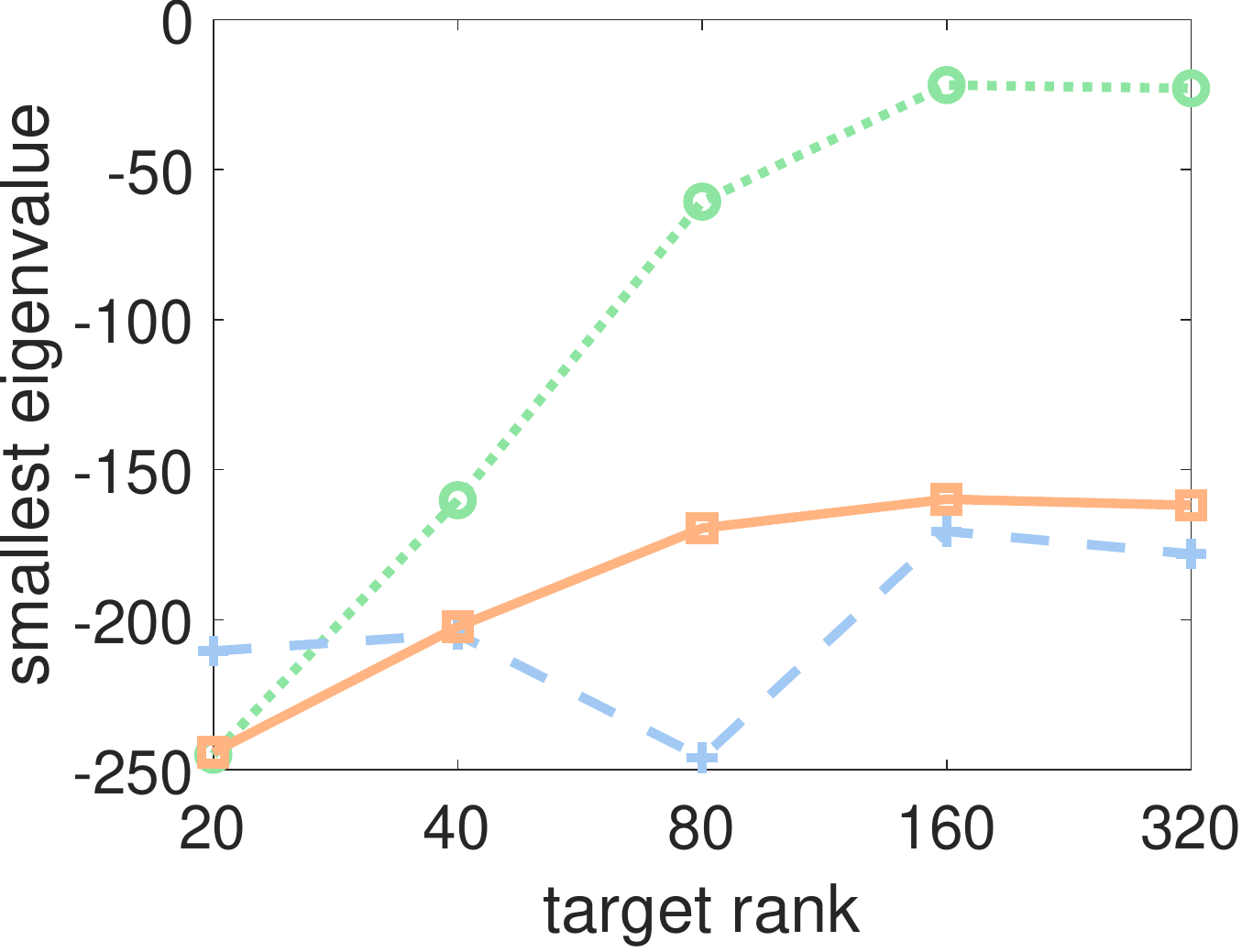}
	     & \includegraphics[width=0.3\textwidth]{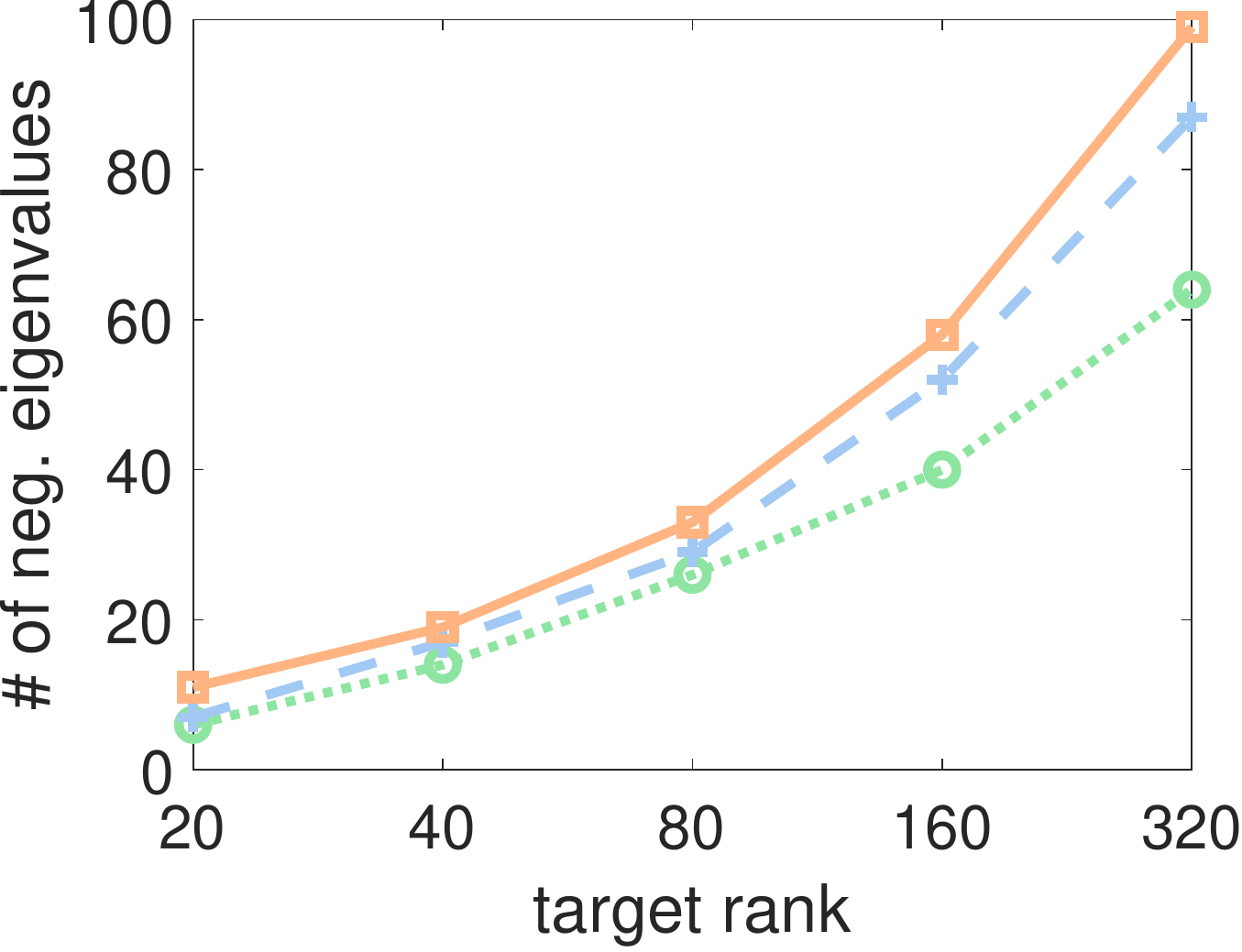}\\
	     \hspace{0.5cm}(a)&\hspace{0.7cm}(b)&\hspace{0.5cm}(c)\\
	\end{tabular}
	\caption{MEKA experiments with \textsf{pendigit} data set and extreme learning kernel. Same legend as in Figure \ref{pic:elmResults1}.	\label{pic:elmResults5}}
	
\end{figure}

\section{OUT-OF-SAMPLE EXTENSION}\label{sup:sec:4}
If the model has to be applied to new data points, one would like to modify them in a consistent way with respect to the training scenario. 
In our proposal we provide a strategy for a shift correction and a normalization, which plays a key role and needs to be taken into account, if a new test point $\x'$ is considered. 

The training model finally consists of the matrices $\Q^{i}$, the link matrix $\Li$ a shift parameter $\lambda_{shift}$ and an index set $\mathcal{I}$ (block-wise denoted as $\mathcal{I}_i$), referring to known reference points. Further, the self-similarities
$k(\x,\x)$ of the training points need to be stored if a normalization is required for a non-stationary kernel function.

The challenge of an out-of-sample extension can be solved in different ways,
here we suggest two strategies:

(1) In the direct approach, one needs to calculate the kernel evaluations of $k(\x',\x)$, with respect to some $\x$, e.g. the support vectors, needed in the prediction model. This is done directly on the original, unapproximated kernel function. If the kernel function has to be normalized an additional step, as shown in Remark 1, is required. 
Note that by evaluating self similarities $k(\x,\x)$ the parameter $\lambda_{shift}$ needs to be added. This approach is particular useful if the MEKA approximation is very accurate and the evaluation of the kernel function is cheap.

(2) The indirect approach maps the new point in the approximated kernel representation as follows.
From the MEKA algorithm part 2 we have stored the landmark matrices of each cluster $\Q^{i}$. Additionally, we need to store the cluster-wise matrices (from the SVD) used to generate $\Q^{i}$, which is rather cheap. Now we calculate the similarities of $k(\x',\x_l)$ for each block $i$, using the original kernel function, where $l \in \mathcal{I}_i$ are the landmark indices of block $i$. These small landmark vectors are used to generate an extended $\Q^{i}$. From the enlarged $\Q^{i}$ a block-matrix $\Q$ is constructed and can be used in the same way as in the MEKA approach. Additional modifications regarding the shift correction and normalization can be applied as shown before.

\section{CLASSIFICATION EXPERIMENT INFORMATION} \label{sup:sec:5}
Typical hyperparameters which are obtained in the classification experiments from Section 4.3 are listed in Table \ref{tab:hyperparameter}. Note that the overall rank approximation is $c$ times $k$, since the same rank per cluster strategy was used. The parameters are tuned with a grid-search and 5-fold cross-validation. The classification results were collected via a 10-fold cross-validation and the hyperparameter tuning was executed in each fold. Hence, Table \ref{tab:hyperparameter} provides just an excerpt of the utilized parameters. 
\begin{table}[h]
    \label{tab:hyperparameter}
    \caption{Typical hyperparameters for classification experiments.\\} 
    \begin{center}
    \begin{tabular}{c|c|c|c|c|c|c|c|c|c|c}
         \textbf{Dataset}&\textbf{ID}  & $k$ & $c$ & $\gamma$ &$\rho$& $p;a$ & $C_{rbf}$ & $C_{elm}$&$C_{poly}$ & $C_{tl1}$\\
         \hline
         &&&&&&&&&&\\
         spambase&1 & 128    & 3  & 0.1 & 39.9 & 10;2 & 1000&1000& 1000&1\\
         artificial 1&2 & 64    &  3 &  1 & 1.4 & 8;2 &   100 &1 & 1000&10\\
         cpusmall&3 & 16 & 3 & 10 & 8.4 & 10;2 &10&10&1000&1\\
         gesture&4 &  16   &  3 &  15  & 22.4 & 10;2 &1000 &1000&100&1\\
         artificial 2&5 &  16  &  3   &  0.1 & 10.5 & 2;3 &  100 & 1&1&1\\
         pendigit& 6 & 16 & 3 & 1 & 11.2 & 8;3 & 10 & 100&100&100\\
    \end{tabular}
    \end{center}
\end{table}

In Table \ref{tab:approxError} the relative approximation error of the matrices used in the classification experiments from Section 4.3 are presented. Note that the approximation error of the corrected matrix depends heavily on the magnitude of the shift parameter, as shown in the error bound of Section 4.2. Although the reconstruction error of the corrected matrix worsens in some cases, at the same time, the classification accuracy does not suffer from this effect.
\begin{table}[h]
\caption{Relative approximation error (mean and standard deviation) of MEKA approximation (M1) and Lanczos shift corrected matrix (M2).\label{tab:approxError}}
\begin{center}
\begin{tabular}{c|c|c|c|c|c|c|c|c|c}
\multicolumn{2}{c}{}&\multicolumn{2}{|c|}{\textbf{RBF}}&\multicolumn{2}{c|}{\textbf{Elm}}&\multicolumn{2}{c|}{\textbf{Poly}}&\multicolumn{2}{c}{\textbf{TL1}}\\
\textbf{Dataset}&\textbf{ID}&\textbf{M1}&\textbf{M2}&\textbf{M1}&\textbf{M2}&\textbf{M1}&\textbf{M2}&\textbf{M1}&\textbf{M2}\\
\hline
&&&&&&&&&\\
\multirow{2}{*}{spambase}&\multirow{2}{*}{1}&     0.01&0.35&0.17&5.30&0.01&0.49&0.00&0.10\\
&&(0.01)&(0.57)&(0.03)&(2.23)&(0.01)&(0.51)&(0.00)&(0.06)\\[0.15cm]
\multirow{2}{*}{artificial 1}&\multirow{2}{*}{2}&0.00&0.00&0.00&0.02&0.00&0.00&0.01&0.18\\
&&(0.00)&(0.00)&(0.00)&(0.02)&(0.00)&(0.00)&(0.00)&(0.13)\\[0.15cm]
\multirow{2}{*}{cpusmall}&\multirow{2}{*}{3}&0.11&2.92&0.06&1.66&0.00&0.00&0.00&0.04\\
&&(0.04)&(1.93)&(0.02)&(0.70)&(0.00)&(0.00)&(0.00)&(0.03)\\[0.15cm]
\multirow{2}{*}{gesture}&\multirow{2}{*}{4}&0.22&23.41&0.06&1.61&0.47&18.09&0.01&0.20\\
&&(0.07)&(10.86)&(0.02)&(0.40)&(0.04)&(8.98)&(0.00)&(0.03)\\[0.15cm]
\multirow{2}{*}{artificial 2}&\multirow{2}{*}{5}&0.00&0.02&0.07&2.70&0.00&0.00&0.00&0.02\\
&&(0.00)&(0.01)&(0.01)&(0.58)&(0.00)&(0.00)&(0.00)&(0.01)\\[0.15cm]
\multirow{2}{*}{pendigit}&\multirow{2}{*}{6}&0.21&5.82&0.09&2.71&0.00&0.03&0.00&0.06\\
&&(0.05)&(3.98)&(0.01)&(0.30)&(0.00)&(0.02)&(0.00)&(0.01)\\[0.15cm]
\end{tabular}
\end{center}
\end{table}
\newpage
\section{KERNEL PROPERTIES}  \label{sup:sec:6}
Table \ref{tab:kernelProps} presents a brief overview of the different kernel functions used in this paper. We do not claim completeness of the listed properties and literature references but show information to support the relevance of an extended MEKA approach. The kernel properties of the Gaussian rbf and polynomial kernel are collected from \cite{shawe2004kernel} and in case of the other kernels the information are obtained from the first publishing paper. 
\begin{landscape}
\begin{table}
    \centering
    \caption{Overview of utilized kernels and their properties provided with an excerpt of their applications from the literature and with preprocessing steps applied in this paper.\\\label{tab:kernelProps} }    
    \begin{tabular}{c|c|c|l}
    
        \textbf{Kernel}&\textbf{Properties}&\textbf{Publications}&\textbf{Preprocessing}\\
        \hline\hline
        &&&\\
        Gaussian rbf&\parbox{5.5cm}{$\gamma \ge 0$,\\ $k_{rbf}(\x,\y) \ge 0$,\\$k_{rbf}(\x,\x) = 1$,\\infinite dimensional feature space,\\unitary-invariant,\\shift-invariant}&\parbox{6cm}{introduced in \cite{scholkopf1997comparing} for the SVM;\\ among other things it is used for kernel based methods on EEG signals \citep{BAJOULVAND201762}, opinion mining and sentiment analysis \citep{gopi2020classification}, ground penetrating radar analysis \citep{tbarki2016rbf}}&$[0,1]$-Normalization\\
        &&&\\\hline
        Polynomial&\parbox{5.5cm}{$p > 0, q \ge 0$,\\$\binom{d+p}{p}$ dimensional feature space,\\unitary-invariant,\\non-stationary}&\parbox{6cm}{\vspace{0.3cm}first used by \cite{poggio1975optimal};\\among other things it is used for termite detection \citep{achirul2018comparison}, person re-identification \citep{chen2015similarity}, speaker verification \citep{yaman2013using}}&L2-Normalization\\
        &&&\\\hline
        Extreme Learning&\parbox{5.5cm}{parameter-insensitive,\\rbf alternative,\\differentiable,\\non-stationary}&\parbox{6cm}{\vspace{0.3cm}introduced in \cite{frenay2011parameter};\\among other things it is used for clustering by fuzzy neural gas \citep{geweniger2013clustering}, alumina concentration estimation \citep{zhang2017alumina}, predicting wear loss \citep{ulas2020new}}&\parbox{3.5cm}{($\sigma,\mu$)-Normalization,\\L2-Normalization}\\
        &&&\\\hline
        Truncated Manhattan&\parbox{5.5cm}{\vspace{0.3cm}$0 \le \rho \le d$,\\two-level deep piecewise linear,\\compactly supported,\\indefinite,\\$\rho = 0.7d$ stable performance,\\shift-invariant}&\parbox{6cm}{introduced in \cite{tl1Suykens};\\among other things it is used in LS-SVM and PCA \citep{HUANG2017162}, piecewise linear kernel support vector clustering \citep{SHANG2017464}}&$[0,1]$-Normalization\\
    \end{tabular}
\end{table}
\end{landscape}

\bibliography{supplement_literature}